\documentclass{article}

\usepackage[preprint]{neurips_2022}

\usepackage[utf8]{inputenc} % allow utf-8 input
\usepackage[T1]{fontenc}    % use 8-bit T1 fonts
\usepackage{hyperref}       % hyperlinks
\usepackage{url}            % simple URL typesetting
\usepackage{booktabs}       % professional-quality tables
\usepackage{amsfonts}       % blackboard math symbols
\usepackage{nicefrac}       % compact symbols for 1/2, etc.
\usepackage{microtype}      % microtypography
\usepackage[table,svgnames,dvipsnames]{xcolor}         % colors
\usepackage{pythonhighlight}
\usepackage{amsmath}
\usepackage{longtable}      % multipage tables
\usepackage{caption}
\usepackage{tikz}
\usepackage{nicefrac}       % compact symbols for 1/2, etc.
\usepackage{microtype}      % microtypography
\usepackage[capitalise]{cleveref}
\usepackage{multicol}
\usepackage{subfigure}
\usepackage{txfonts}

\usepackage{placeins}

\usepackage{longtable}

\usepackage{multirow}

\usepackage{array}
\newcolumntype{x}[1]{>{\centering\arraybackslash\hspace{0pt}}p{#1}}

\usepackage{color,soul}

\usetikzlibrary{calc,trees,positioning,arrows.meta,chains,shapes.geometric,%
decorations.pathreplacing,decorations.pathmorphing,shapes,%
matrix,shapes.symbols,plotmarks,decorations.markings,shadows}
\usetikzlibrary{arrows.meta}
\usetikzlibrary{calc, positioning, arrows}

\usepackage[framemethod=tikz]{mdframed}  % hot finding boxes

\DeclareRobustCommand{\performance}[1]{\textcolor{JungleGreen}{#1}}
\DeclareRobustCommand{\data}[1]{\textcolor{DarkOrchid}{#1}}
\DeclareRobustCommand{\hardware}[1]{\textcolor{Mahogany}{#1}}

\mdfdefinestyle{side}{
  rightline=false,
  innerleftmargin=10,
  innerrightmargin=0,
  innertopmargin=0,
  innerbottommargin=0,
  outerlinewidth=3pt,
  topline=false,
  leftline=true,
  bottomline=false,
  skipabove=\topsep,
  skipbelow=\topsep,
}

\title{The Falcon Series of Open Language Models}

\author{
\textbf{\underline{The Falcon LLM Team}\footnotemark} \vspace{0.3cm}\\
\small \textbf{Ebtesam Almazrouei \hspace{0.3cm} Hamza Alobeidli \hspace{0.3cm} Abdulaziz Alshamsi \hspace{0.3cm} Alessandro Cappelli} \vspace{0.2cm}\\
\small \textbf{Ruxandra Cojocaru 
\hspace{0.3cm} Mérouane Debbah
\hspace{0.3cm} Etienne Goffinet
\hspace{0.3cm} Daniel Hesslow \hspace{0.3cm} Julien Launay} \vspace{0.2cm}\\
\small \textbf{Quentin Malartic
\hspace{0.3cm} Daniele Mazzotta
\hspace{0.3cm} Badreddine Noune \hspace{0.3cm} Baptiste Pannier \hspace{0.3cm} Guilherme Penedo}\vspace{0.4cm}\\
Technology Innovation Institute, Abu Dhabi
}
\begin{document}

\maketitle
\vspace{-.35in}

\begin{center}
\url{https://huggingface.co/tiiuae/}
\end{center}

\definecolor{adaptation}{HTML}{FCCDE5}
\definecolor{objective}{HTML}{FFC2BA}
\definecolor{architecture}{HTML}{FaF9B3}
\definecolor{evaluation}{HTML}{B9DEFF}
\definecolor{neutral}{HTML}{CFCFCF}
\definecolor{fine-tuning}{HTML}{D7F3E7}
\definecolor{fine-tuning-darker}{HTML}{c4eddc}

\maketitle

\begin{abstract}
We introduce the Falcon series: 7B, 40B, and 180B parameters causal decoder-only models trained on a diverse high-quality corpora predominantly assembled from web data. The largest model, Falcon-180B, has been trained on over 3.5 trillion tokens of text--the largest openly documented pretraining run. Falcon-180B significantly outperforms models such as PaLM or Chinchilla, and improves upon concurrently developed models such as LLaMA 2 or Inflection-1. It nears the performance of PaLM-2-Large at a reduced pretraining and inference cost, making it, to our knowledge, one of the three best language models in the world along with GPT-4 and PaLM-2-Large. We report detailed evaluations, as well as a deep dive into the methods and custom tooling employed to pretrain Falcon. Notably, we report on our custom distributed training codebase, allowing us to efficiently pretrain these models on up to 4,096 A100s on cloud AWS infrastructure with limited interconnect. We release a 600B tokens extract of our web dataset, as well as the Falcon-7/40/180B models under a permissive license to foster open-science and accelerate the development of an open ecosystem of large language models. 
\footnotetext{Authors listed alphabetically, contributions detailed in \cref{sec:contributions}. Correspondence to falconllm@tii.ae}
\end{abstract}

\begin{figure}[h]
% UPDATE FIGURES: LEFT ZERO-SHOT OVERALL, RIGHT ONE-SHOT AGAINST PALM-2 SERIES
\centering
\includegraphics[width=.5\textwidth]{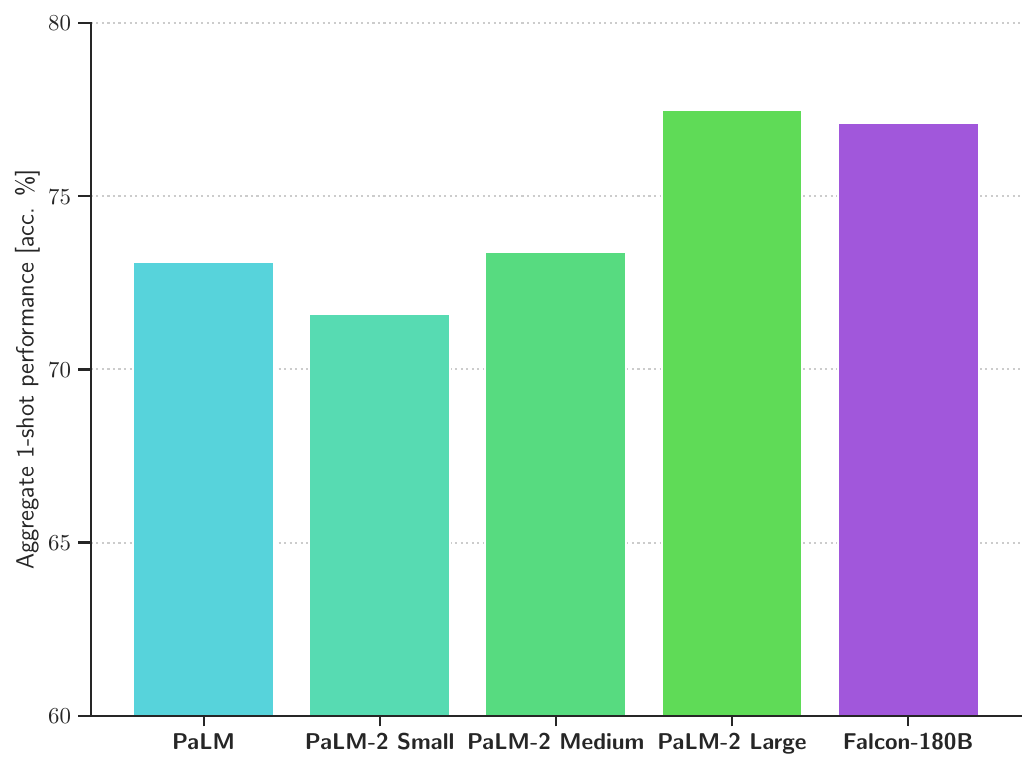}
\caption{\textbf{The Falcon series of models achieves competitive performance, with Falcon-180B nearly matching the 1-shot performance of PaLM-2 Large.} 1-shot performance of PaLM \citep{palm}, PaLM-2 \citep{palm-2}, and Falcon-180B, on a set of tasks from~\citet{brown2020gpt3}. These evaluation results are only a small snapshot of our evaluations, see \cref{sec:evaluation} for details and comparisons with GPT-3.5/4, LLaMA-1/2, and Inflection-1.}
\label{fig:main_lead}
\end{figure}

\newpage

\section{Introduction}

The on-going Cambrian explosion of language models has been primarily fueled by the unique \emph{scalability} of popular Transformer-based recipes. This scalability manifests over multiple axes:
\begin{itemize}
    \item{\textbf{\performance{Performance scalability}} (and predictability).} Increase in pretraining compute budgets systematically yield improvements in language modeling capabilities, in a consistent and predictable way \citep{kaplan2020scaling}. Falcon-180B is the first publicly documented GPT-3-sized model to follow the updated scaling law recommendations of \citet{hoffmann2022training}, with a total pretraining length of 3,500 billion tokens, without any upsampling.
    \item{\textbf{\data{Data scalability.}}} To scale-up pretraining efficiently, and to decouple pretraining and inference compute, increasingly large models should be trained for longer, on larger corpora. To sustain the Falcon series, we developed RefinedWeb \citep{refinedweb}, a 5 trillion tokens high-quality filtered and deduplicated web dataset--the largest publicly documented.
    \item{\textbf{\hardware{Hardware scalability.}}} Transformer models \citep{vaswani2017attention} are naturally suited for modern GEMM optimized hardware, allowing their training and inference to be efficiently distributed over a large number of accelerators \citep{megatron-2, pope2023efficiently}. With Falcon-180B, we demonstrate scaling-up pretraining to 4,096 A100 40GB with only 50 Gbps interconnect per accelerator on cost-efficient AWS cloud infrastructure.
\end{itemize}

Building upon these fundamentals, increasingly large language models give rise to so-called emergent capabilities \citep{wei2022emergent}. These capabilities can be further tailored to human preference, to build instruction-following or chatty models \citep{ouyang2022training}. All together these methods have lead to the widespread deployment of large language models in customer-facing applications, such as ChatGPT (GPT-3.5/4, \citet{gpt-4}), Claude, Bard (PaLM-2, \citet{palm-2}), or Pi (Inflection-1,~\citet{inflection-1}). In this paper, we primarily report on the pretraining alone of the Falcon series of models, and leave further downstream finetuning and alignment to future works.

The Falcon series is made of three causal decoder-only models trained on up to 4,096 A100. We assembled a pretraining dataset of 3,500 billion tokens, predominantly sourced from our work on RefinedWeb \citep{refinedweb}--a massive filtered and deduplicated web dataset. The architecture of the models is based on PaLM \citep{palm}, although we independently validated each decision, ultimately resulting in some minor tweaks--see \cref{sec:ablations} for details. The Falcon series leverages extensive custom tooling (i.e., pretraining codebase, data pipeline), of which development started in August 2022, with training of the models kicked-off in December 2022. In-depth evaluation shows that the Falcon series is competitive across scale, and that Falcon-180B nears the performance of PaLM-2 Large, positioning it as the best open model and in the top-3 of the best language models.

\paragraph{Contributions.} With this paper and the Falcon series, we make the following contributions:
\vspace{-0.1in}
\begin{itemize}
    \item \textbf{Public documentation of the pretraining of a large-scale model.} Recent state-of-the-art models have been scarcely documented, hindering further research and progress in the field. At variance with these works, we extensively document the pretraining of the Falcon series.
    \item \textbf{Open data and models.} To accelerate research, and to enable community-driven improvements of large language models, we openly release Falcon-7/40/180B, and a 600 billion tokens extract of the RefinedWeb dataset: \url{https://huggingface.co/tiiuae/}.
\end{itemize}

\begin{table}[b]
\vspace{-0.4in}
\centering
\caption{\textbf{The Falcon series of models covers a wide range of \performance{capabilities} and \hardware{inference requirements}, enabled by \data{large-scale web data}.} Falcon-7B can efficiently run on consumer hardware (e.g., Apple M2), while Falcon-180B typically requires dedicated inference infrastructure (e.g., $8 \times\text{A100 80GB}$ ). We report steady zero-shot performance gains across the entire Falcon series.}
\vspace{0.1in}
\begin{tabular}{@{}lccc@{}}
\toprule
                      & \textbf{Falcon-7B}   & \textbf{Falcon-40B}  & \textbf{Falcon-180B} \\ \midrule
\textbf{Pretraining} {[}tokens{]}   & 1,500B               & 1,000B               & 3,500B               \\
\textbf{Compute} {[}PF-days{]}      & 730                  & 2,800                & 43,500               \\
\textbf{Training} {[}A100s{]} & 384                  & 384                  & 4,096                \\
\textbf{Availability}               & Apache 2.0               & Apache 2.0              & Responsible use license                    \\
\textbf{Agg. performance} (\cref{sec:eai})     & 60.8 & 67.1 & 70.3 \\
\textbf{Closest model}             & <GPT-3                 & Chinchilla           & PaLM-2 Large         \\ \bottomrule
\end{tabular}
\end{table}

\newpage

\tableofcontents

\newpage

\section{State-of-the-art: from language modeling to frontier models}

We provide in this section an overview of general trends and works adjacent to the Falcon series. For an in-depth literature review of individual technical components, see the relevant sections. 

\textbf{Language modeling.} Beyond corpus/task-specific approaches, the first large-scale vector-based word embeddings methods \citep{mikolov2013efficient, pennington2014glove} pioneered unsupervised learning from massive unstructured text corpora. The integration of deep recurrent neural architectures enabled models to deal with polysemy and to integrate contextual information \citep{peters-etal-2018-deep}; up to the emergence of the transfer learning paradigm, leveraging universal models specialized to downstream tasks through finetuning \citep{howard2018universal}. Despite the existence of many of the first principles currently used, early scaling attempts \citep{jozefowicz2016exploring} only had mixed success, partly due to the fastidiousness and poor scalability on common hardware of recurrent approaches.

\textbf{Transfomer models.} The introduction of the attention-based Transformer architecture \citet{vaswani2017attention} sparked an explosion in the number of recipes to produce efficient, generalist models: from embedding and classification focused encoder-only BERTs \citep{kenton2019bert}, to causal decoder-only GPTs \citep{radford2018gpt}. Specifically, GPT-2 \citep{radford2019language} was the first series of models to popularize emergent few-shot generalization abilities, allowing a model to understand and perform arbitrary tasks simply from in-context instructions and demonstrations. 

\textbf{Large language models.} The aforementioned works laid out the key components to current models; the last ingredient, scaling, was demonstrated by GPT-3 \citep{brown2020gpt3}, and consecrated by the outlining of scaling laws 
\citep{kaplan2020scaling}. As increasingly large amounts of compute are spent to pretrain models, commensurate gains are made in language modeling performance. The tantalizing prospect of a systematic and direct path to more capable language models lead to a "scaling frenzy": first with reproductions of GPT-3 with Jurassic-1 \citep{J1WhitePaper} or PanGu-Alpha \citep{zeng2021pangu}, and with open efforts such as GPT-J \citep{gpt-j}, OPT \citep{zhang2022opt}, or BLOOM \citep{bloom}; second, with works pushing further the limits of scaling with Gopher \citep{gopher}, MT-NLG \citep{smith2022using}, or PaLM \citep{palm}. Increased development and adoption of large language models also lead to improvements of pretraining methods. Notably, \citet{hoffmann2022training} demonstrated with Chinchilla that optimal scaling should actually jointly increase model size and pretraining dataset. For deployment in the real-world, it may even be desirable to train far past so-called optimality, to decouple training and inference compute and reduce serving costs. This is illustrated with the LLaMA models \citep{llama, llama2}, with 7B/13B/30B/70B parameters models trained on up to 2 trillion tokens.

\textbf{Frontier models.} Concurrently to this work, so-called frontier models have emerged, under the shifting definition of "large-scale machine-learning models that exceed the capabilities currently present in the most advanced existing models, and can perform a wide variety of tasks". Although a moving goal, we attribute recent works on GPT-4 \citep{gpt-4} and PaLM-2 \citep{palm-2} as early contributions to this category. These stand out by their significantly increased compute budget, and improved capabilities. See \cref{sec:unofficial} for details on our approach to undocumented models.

\begin{figure}[h]
\centering
\includegraphics[width=.9\textwidth]{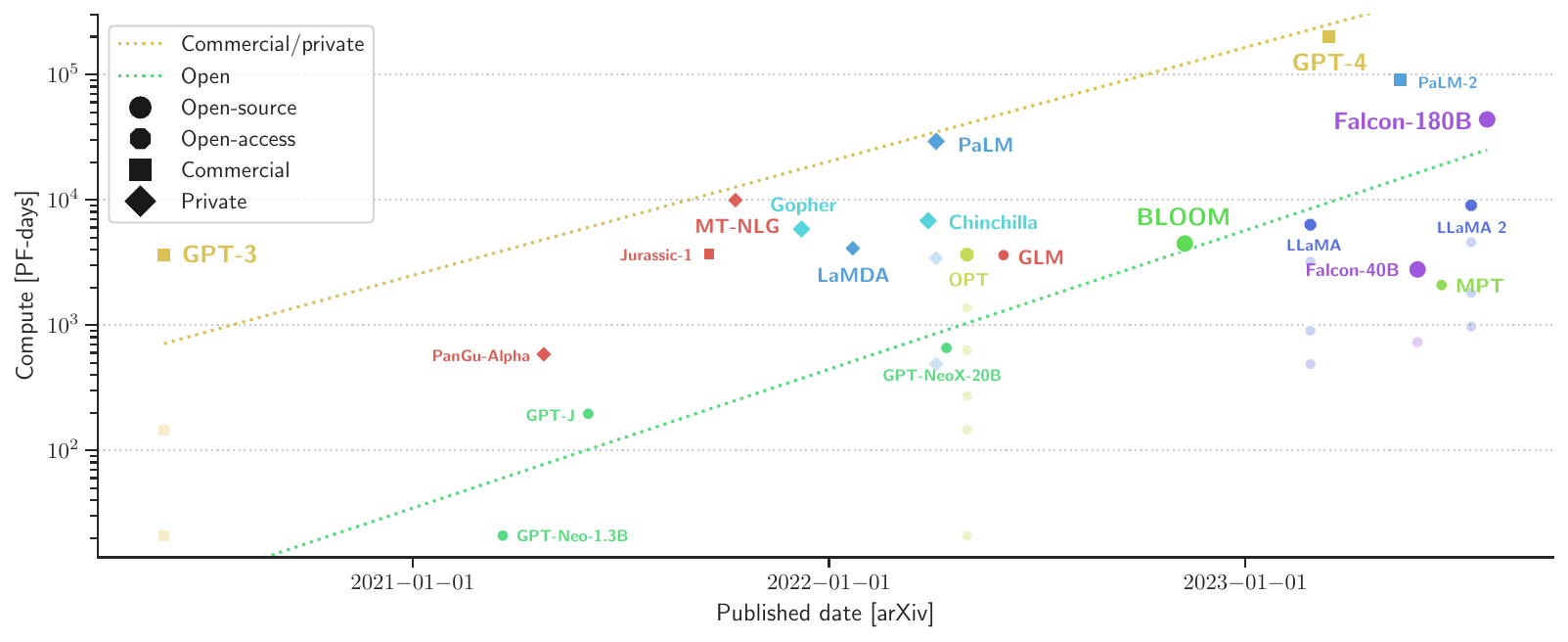}
\caption{\hardware{\textbf{Although open models lag behind closed models in pretraining compute (by $\sim$18 months), the gap is not widening.}} After a 1 year lag, GPT-3 sparked a "scaling frenzy", with the emergence of numerous LLMs. Models in the >10,000 PF-days range remain rare for open-source.}
\label{fig:model-flops}
\end{figure}

\newpage

\section{Design philosophy}
\label{sec:design}

Inspired by the bitter lesson \citep{sutton2019bitter}, we believe that scalable methods that best leverage compute are ultimately the most effective. Accordingly, in designing the Falcon series of models, we focused primarily on \textbf{scalability}, across three axes: \performance{performance}, \data{data}, and \hardware{hardware}. 

\textbf{\performance{Performance scalability.}} The sustained increase in the scale of large language models (\cref{fig:model-flops}) has been primarily motivated by so-called scaling laws: with increased pretraining compute comes commensurate improvements in language modeling capabilities~\citep{hestness2017deep, kaplan2020scaling}. This systematic path to model improvement has proved far more reliable than waiting for the occasional research epiphany to manifest a paradigm-shifting method. But upstream and downstream performance are not simply motivators of scaling, they are also a powerful driver. Indeed, quantifying the impact of modeling interventions (e.g., architecture tweaks, data sourcing, hyperparameters selection) is critical to sustaining the feedback loop provided by small-scale ablations. However, the question of \emph{what} to quantify is not trivial: upstream performance alone can be at odds with downstream tasks \citep{Tay2021ScaleEI}, and even downstream metrics may not be aligned with human preferences \citep{stiennon2020learning}. This is made even more challenging by the fundamental contrast between pretraining objective (i.e., predict the next word on a predominantly web-based corpora) and common downstream use (i.e., follow users' instructions in a helpful, harmless, and honest way) \citep{bai2022training}. As we focus on the pretraining stage of the Falcon models, we choose to center our evaluations on measuring \textbf{zero/few-shot generalization on large aggregates of natural language tasks} with the EleutherAI Harness \citep{eval-harness}--similar to the setup of \citet{le2022language, wang2022language}. We build aggregates enabling comparisons with other state-of-the-art models, but note the difficulty in executing principled comparisons: practices diverge widely across papers in terms of task selection, prompting, and even mode of evaluation--standardized benchmarks remain only scarcely adopted. We discuss in \cref{sec:limitations} limitations of our evaluation setup.

\begin{figure}[t]
    \centering
    \includegraphics[width=\linewidth]{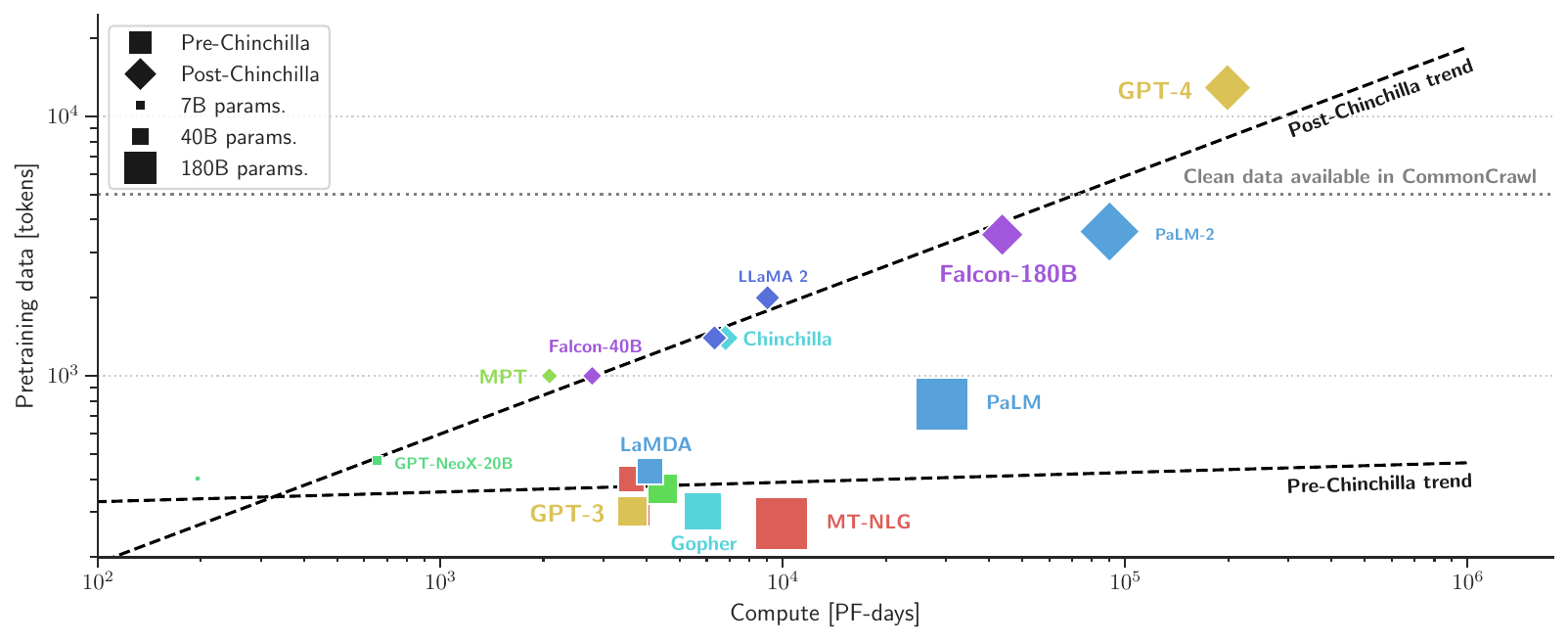}
    \caption{\textbf{\data{Two eras of pretraining practices, from predominant model scaling $\blacksquare$ to joint model and data scaling $\Diamondblack$}}. Before \cite{hoffmann2022training}, models ($\blacksquare$) predominantly scaled their parameter count at a fixed dataset size (around 300 billion tokens), in line with the recommendations of \cite{kaplan2020scaling}. Afterward ($\Diamondblack$), models started scaling both model size and dataset size jointly, sharply increasing the need for scalable data pipelines. Estimate of clean data available in CommonCrawl from our work in \citet{refinedweb}, considering English only--would double if allowing multilinguality.}
    \label{fig:hoffmann-era}
\end{figure}

\textbf{\data{Data scalability.}} An increase in pretraining compute budget can be either spent towards a larger model, or towards longer training. While \citet{kaplan2020scaling} had first found optimal scaling to be mostly model size driven, \citet{hoffmann2022training} revised this finding and found that joint scaling is preferable--see \cref{fig:hoffmann-era} to see impact. Furthermore, growing model size also increases inference burden; where as increasing the pretraining dataset size is decoupled from inference costs. Recent open models have been trained on datasets of up to two trillions tokens \citep{llama2}; because naive repetition risks degrading the model \citep{hernandez2022scaling, data_constrained}, this has led to concerns about the sustainability of data scaling \citep{villalobos2022will}. These concerns are exacerbated by the widely belief that curated corpora are necessary to build state-of-the-art models, requiring manual addition of individual sources such as arXiv papers, books, and more \citep{brown2020gpt3, gao2020pile}. For the Falcon series of models, we choose to focus on \textbf{scaling high-quality web data through stringent filtering and deduplication}, enabling us to collect an English web dataset of 5,000 billion tokens and to not repeat any data during training. We extensively report on this work in \citet{refinedweb}, but also provide some key elements in this paper.

\textbf{\hardware{Hardware scalability.}} Large-scale training requires thousands of hardware accelerators to work efficiently in unison; making the best use of these accelerators requires in turn principled distributed training methods \citep{megatron-1}. Methods that are able to best run efficiently and leverage large-scale compute are often the ones that gain the most traction in the community \citep{hooker2021hardware}, as best evidenced by the Transformer architecture itself \citep{vaswani2017attention}. Furthermore, it is difficult to find architectural improvements that significantly improve the task performance of models, compared to the impact of data for instance \citep{le2022language}. Accordingly, \textbf{we focus architectural decisions not on improving task performance, but on improving hardware scalability and throughput}. We are also concerned with inference scalability \citep{pope2023efficiently}, leading us to the adoption of tweaks such as a (revised) multiquery attention scheme \citep{multiquery}.

Finally, beyond scalability, we are also concerced with \textbf{cost-efficiency} and relying on \textbf{proven approaches}. We reimplement a 3D parallelism strategy \citep{megatron-2} combined with optimizer sharding~\citep{zero-msft}, enabling us to run on a more cost-efficient cloud AWS infrastructure with limited interconnect. We also put a strong focus on memory-saving methods, enabling us to run on cheaper 40GB A100. Reimplementing our data preprocessing and pretraining pipelines from scratch allow us to extensively verify them, rather than relying on an unknown external codebase. We also do not explore radical departures from the classic designs of large language models, such as state-space models \citep{fu2022hungry}, as these have usually not been proven at scale.

\section{Experiments and motivations for data, architecture, and hyperparameters}
\label{sec:ablations}
\vspace{-0.1in}

We first focus on a series of small-scale experiments with models in the 1B-3B parameters range to validate recommended practices, as well as to identify interesting tweaks. We also conducted extensive experiments to validate our web data pipeline, reported in \citet{refinedweb}. With each subject of interest, we also outline common practices and our motivations for exploring this direction.
\vspace{-0.2in}

\subsection{Setup for small-scale experiments}
\label{sec:ablations-setup}

\textbf{Small models.} For these ablations, we seek to be able to rapidly iterate with limited compute costs. This leads us to train 1/3 billion parameters models, for 30/60 billion parameters respectively. We choose these short training lengths to be illustrative of the optimality regime from \citet{hoffmann2022training} under which our larger models are likely to be trained. We base our reference architecture and hyperparameters on the one described for GPT-3 \citep{brown2020gpt3}, with the caveat of using ALiBi positionnal embeddings as our baseline \citep{press2021alibi}. With reasonable resources (32-64 A100), these ablation models can be trained overnight or in a few days, enabling rapid iterations. We note that a caveat of using smaller models is that they may not be illustrative of some of the behaviours of larger models: for instance, \citet{dettmers2022gpt3} found that outlier features emerge at the 6B scale, impacting quantization; concerns around data duplication and memorization have also shown to disprotionately affect larger models \citep{carlini2022quantifying, hernandez2022scaling}

\textbf{Dedicated aggregates.} Although small models enable rapid iterations, they also have limited zero-shot capabilities; naive bulk evaluation would result in most tasks being close to the random baseline, and in significant noise in the evaluations. Using models trained on subsets of The Pile, we identified tasks that showed both reasonable performance at small-scale, and limited variability across runs. We independently quantified variance and average performance on over 50 tasks across 5 runs with a different seed (we use this $\sigma$ for architecture experiments) and across 10 runs with different data subsets and shuffling (for data experiments). Based on this analysis, we source 11 tasks from the evaluation setup of \citet{brown2020gpt3,le2022language,srivastava2023beyond} for our ablations (\texttt{zs-main/data/web}). Note that differences between these three subsets are mostly due to differing practices across time and between teams. \texttt{zs-comp} is a subset of \texttt{main} for comparisons with other models, based on commonly reported tasks. We also report perplexities on The Pile \citep{gao2020pile} (\texttt{ppl-pile}) for architectures (for data experiments, we found perplexities on The Pile to mostly illustrate differences in formatting rather than content), and on a restricted subset of 3 NLP tasks with low variance (\texttt{zs-small}). For our small-scale evaluations, we use both the EleutherAI harness \citep{eval-harness} and BigBench~\citep{srivastava2023beyond}; note that our inference and evaluation codebase differ significantly between this setup and the final results we report in \cref{sec:evaluation}, so results are not directly comparable. We present an outline of the aggregates in \cref{tab:ablations_aggregates}.

\begin{table*}[b]
    \centering
    \caption{\textbf{To evaluate small models used in ablations (1/3B models trained on 30/60B tokens), we build four aggregates across 11 tasks on which to measure zero-shot performance and perplexity.} We trained 15 reference models on subsets of The Pile and with random seeds to identify tasks with performance better than random and low variablity at this scale. All evaluations leverage the EAI Harness \citep{eval-harness} except date which is taken from BigBench~\citep{srivastava2023beyond}. \texttt{main} was built for architecture and hyperparameters ablations; \texttt{data} for data mixtures experiments; \texttt{web} for small-scale ablations on web data; and \texttt{core} for its low variance on a reduced number of tasks. This setup only covers natural language abilities. For \texttt{main}, \texttt{data}, and \texttt{web}, differences are mostly due to individual preferences at the time of the experiments.}
    \vspace{0.1in}
    \label{tab:ablations_aggregates}
    \begin{scriptsize}
    \begin{tabular}{llccccccc}
    \toprule
        \textbf{Tasks} & \textbf{Type} & \textbf{Random} & \texttt{main} & \texttt{comp} & \texttt{data} & \texttt{web} & \texttt{core} & \texttt{pile} \\
        \midrule
         LAMBADA \citep{paperno2016lambada} & Reading Comprehension & 0.0 & \checkmark & &  & & \checkmark & \\
         RACE \citep{lai2017large} & Reading Comprehension & 25.0 & \checkmark & & & & &\\ 
         HellaSwag \citep{zellers2019hellaswag} & Common Sense & 25.0 & \checkmark &\checkmark & \checkmark  & \checkmark & \checkmark &\\
         Winogrande \citep{sakaguchi2019winogrande} & Common Sense & 50.0 & & \checkmark  & \checkmark & &\\ 
         PIQA \citep{bisk2020piqa} & Common Sense & 50.0 & \checkmark &\checkmark & \checkmark & \checkmark& \checkmark &\\ 
         BoolQ \citep{clark2019boolq} & Common Sense & 50.0 & \checkmark & & \checkmark & & \\
         COPA \citep{gordon2012copa} & Common Sense & 50.0 &\checkmark & & \checkmark &  & \\ 
         Date \citep{srivastava2023beyond} & Common Sense & 25.0 & \checkmark & & & & \\ 
         ARC \citep{clark2018arc} & Question Answering & 25.0 & \checkmark &\checkmark & \checkmark & \checkmark& &\\ 
         OpenBookQA \citep{mihaylov2press2021train018openbookqa} & Question Answering & 25.0 & & \checkmark \\
         SciQ \citep{welbl2017sciq} & Question Answering & 25.0 & \checkmark & & \checkmark & \checkmark & &  \\ 
         The Pile \citep{gao2020pile} & Language Modeling &  &&  & & & &  \checkmark \\ 
        \bottomrule
    \end{tabular}
    \vspace{-0.1in}
    \end{scriptsize}
\end{table*}

\subsection{Data: web vs curated, code and multilinguality impact on English performance}
\label{sec:ablations-data}

\subsubsection{Web data alone can outperform curated corpora}
\label{sec:webdata_exp}

\smallskip

\begin{mdframed}[style=side]
Our motivations for predominantly training on web data, details of the processing pipeline, and extensive evaluations are detailed in our dedicated \data{\textbf{\textsc{RefinedWeb}}} paper \citep{refinedweb}. In this section, we only highlight key ablations guiding our decision to focus our efforts on web data.
\end{mdframed}
\vspace{-0.1in}

\textbf{Background.} Since its origins with simpler and shallower statistical language models \citep{shannon1951prediction, mikolov2013efficient}, natural language processing has long leveraged unstructured massive text corpora. If these corpora were at first built "sentence-wise" \citep{chelba2013one}, the emergence of more advanced architectures enabled models to best use long context information present in unified documents \citep{devlin2018bert, radford2018gpt}. Starting with single-domain sources, such as Wikipedia or BookCorpus \citep{zhu2015aligning}, datasets scaled along with models, and massive web-scrapes gained prevalence \citep{ortiz2019oscar, raffel2019t5}. However, it is widely believed that web data alone is insufficient to build performant models \citep{brown2020gpt3, gao2020pile}. Accordingly, large language models train on mixed corpora, combining both large-scale web data, and curated so-called "high-quality" individual sources (e.g., books, technical papers, social media conversations)-- see \cref{tab:sota_mixes} for an overview of common pretraining mixes.

However, as we outlined in the \data{\textbf{data scalability}} discussion in \cref{sec:design}, sourcing the trillions of tokens required for pretraining a modern language model may be challenging. This leads us to challenge the idea that curated data is fundamentally better than web data. Notably, building upon the work of \citet{lee2022deduplicating}, we study how stringent deduplication and extensive filtering inspired by~\citet{gopher} may enable web data alone to train performant models.

\begin{table}[b]
\centering
\caption{\data{\textbf{Following the recommendations of \citet{hoffmann2022training}, pretraining datasets have increased in size, causing an increase in the prevalence of web data.}} Web data sources are massive web scrapes such as C4 \citep{raffel2019t5}, sourced from CommonCrawl. Curated web data undergoes a targeted domain filtering: this includes CC-News \citep{Hamborg2017} for instance. We consider sources such as arXiv, Wikipedia, or PubMed as technical curated data, and sources such as Reddit, HackerNews, or StackOverflow as conversational. We report not the size of the overall dataset, but the amount of tokens used for pretraining. For LaMDA \citep{thoppilan2022lamda}, numbers are roughly estimated as only rough counts are provided in the paper. }
\vspace{0.1in}
    \label{tab:sota_mixes}
\begin{small}
\begin{tabular}{lccccccccc}
\toprule
                    & \textbf{Web} & \textbf{Curated web} & \textbf{Curated} &               &           &               &               &  & \textbf{Total} \\
                    &              &                      &                  & Books         & Tech. & Code          & Conv. &  Epochs                           &                \\
\midrule
\textbf{GPT-3}     & 60 \%        & 22 \%                & 18 \%            & 16 \%         & 2 \%      & 0 \%          & 0 \%          & 2.4                         & 300B            \\
\textbf{The Pile}   & 18 \%        & 10 \%                & 72 \%            & 15 \%         & 40 \%     & 7 \%          & 10 \%         & 1.8                        & 340B            \\
\textbf{MT-NLG}     & 38 \%      & 29 \%              & 33 \%          & 16 \%       & 9 \%    & 2 \%        & 6 \%        & 2                        & 270B            \\
\textbf{Gopher}     & 58 \%        & 10 \%                & 32 \%            & 27 \%         & 2 \%      & 3 \%          & 0 \%          & $\sim$1                        & 300B            \\
\textbf{LaMDA}      & 25 \%        & 0 \%        & 75 \%            & 0 \% & 25 \%     &  0 \% & 50 \%         &            & 340B           \\
\textbf{PaLM}       & 27 \%        & 1 \%                 & 72 \%            & 13 \%         & 4 \%      & 5 \%          & 50 \%         &                             & 780B            \\ \midrule
\textbf{Chinchilla} & 55 \%        & 10 \%                & 35 \%            & 30 \%         & 1 \%      & 4 \%          & 0 \%          & 1.1                        & 1400B           \\
\textbf{LLaMA}      & 82 \%        & 0 \%                 & 18 \%            & 5 \%        & 7 \%      & 4 \%        & 2 \%          & 1.7                        & 1400B           \\
\textbf{Falcon}     & 84 \%        & 0 \%                 & 16 \%            & 6 \%          & 2 \%      & 3 \%          & 5 \%          & 1                           & 3500B   \\       
\bottomrule
\end{tabular}
\end{small}
\end{table}

\smallskip

\begin{mdframed}[style=side]
\textbf{Question.} Can web data alone (with filtering and deduplication) be used to train models outperforming models trained on curated data, as measured by natural language zero-shot performance?
\end{mdframed}
\vspace{-0.1in}

\textbf{Methods.} We train 1/3B parameters models on 27/60B tokens, on a number of datasets of interest and on intermediary artefacts of our data pipeline. For state-of-the-art web datasets, we consider two versions of OSCAR \citep{ortiz2019oscar, oscar_new} and C4 \citep{raffel2019t5}. For curated datasets, we consider The Pile \citep{gao2020pile}, the most popular pre-aggregated dataset--many models have also elected to base their pretraining data on specific components of The Pile \citep{smith2022using, zhang2022opt}. RW-Raw corresponds to the output of our pipeline with the least amount of filtering, immediately after text extraction--but still with English language identification applied as well as URL blocklist for known adult content; RW-Filtered applies a first round of heuristics, similar to the ones used by \citet{gopher}; finally, \data{\textsc{RefinedWeb}} corresponds to our final web dataset, with deduplication applied in two stages. We evaluate all models on the \texttt{zs-web} aggregate--see \cref{tab:ablations_aggregates} for details of its composition.

\begin{table*}[b]
    \centering
    \caption{\textbf{\data{Curation is not a silver bullet for zero-shot generalization: small-scale models trained on \data{\textsc{RefinedWeb}} outperform models trained on web data (C4, OSCAR), and on curated corpora (The Pile).}}  Zero-shot accuracy on \texttt{zs-web} aggregate ($\sigma=0.69$, \emph{likely} $\pm 0.69\%$, \emph{very likely} $\pm 1.38\%$ differences in scores). All models trained with identical architectures and hyperparameters, for the same amount of tokens. We find that OSCAR-22.01 underperforms other datasets signficantly, perhaps because deduplication is only optional. C4 is a strong baseline, with OSCAR-21.09 lagging slightly behind, but we find that RefinedWeb outperforms both web datasets and the curated dataset, The Pile--performance gap with C4 is insufficient to be conclusive, but C4 would be too small for our models. Both filtering and deduplication contribute significantly to improving zero-shot performance.}
    \vspace{0.1in}
    \label{tab:ablations-web}
    \begin{scriptsize}
    \begin{tabular}{cccccccc}
    \toprule
    & \multicolumn{3}{l}{\textsc{\textbf{Massive web datasets}}} & \multicolumn{1}{l}{\textsc{\textbf{Curated}}} & \multicolumn{3}{l}{\textsc{\textbf{Ours}}} \\ \midrule
       & OSCAR-21.09 & OSCAR-22.01 & C4 & The Pile & RW-Raw & RW-Filtered & \textbf{\data{\textsc{RefinedWeb}}}  \\\midrule
        \textbf{1B@27GT} & 55.0\% & 52.7\% & 55.7\% & 53.4\% & 52.7\% & 54.3\% & \textbf{56.2\%} \\
        \textbf{3B@60GT} & 59.1\% & 55.9\% & 59.6\% & 57.9\% & 57.4\% & 58.2\% & \textbf{59.8\%} \\
       \bottomrule
    \end{tabular}
    \end{scriptsize}
    \vspace{-0.15in}
\end{table*}

\textbf{Results.} Results for this round of experiments are presented in \cref{tab:ablations-web}. In line with expectations, we find that raw web data (RW-Raw) performs poorly; similarly OSCAR-22.01 offers the worst performance of all datasets. This is likely because its creators have opted to distribute it by default without any deduplication applied. Conversely, OSCAR-21.09 and C4 are both strong baselines. We notably find that The Pile very likely does not deliver better performance than web data. 

Subsequent stages in our pipeline significantly uplift dataset quality and the performance of models trained on it. Filtering alone enables us to close the gap with The Pile, while the addition of stringent deduplication allows for \data{\textsc{RefinedWeb}} to be the best dataset among the ones we benchmarked. 

We note two limitations of these experiments. First, the models are small, and trained on a limited amount of data. However, it is likely that gains from deduplication actually increase with model scale, as larger models are more sensitive to duplicates \citep{hernandez2022scaling} and more likely to memorize individual samples \citep{carlini2022quantifying}. Second, our evaluation focuses on natural language tasks. It is unlikely that models trained on web data alone would compare favorably to models trained on The Pile on code tasks for instance, as The Pile explicitely include code, while massive web scrapes are likely to be mostly devoid of it except for some incidental occurrences.

\begin{mdframed}
\textbf{Finding.} Challenging beliefs on data quality, filtered and deduplicated web data \emph{alone} allows models to match the natural language tasks performance of models trained on curated data.
\end{mdframed}

\subsubsection{Against a strong web baseline, curated data can even be detrimental}
\label{sec:curated_exp}

\textbf{Background.} In \cref{sec:webdata_exp} and \cref{tab:sota_mixes}, we have noted that large language models use pretraining datasets combining both massive web crawl data and individual curated sources. Such sources were first employed to build domain-specific models \citep{beltagy2019scibert}; they have also been proposed to broaden the expressiveness of models, for instance for conversational modalities \citep{adiwardana2020towards, thoppilan2022lamda}. Some of these sources can also exist at the intersection of strongly curated data and crawls: \citet{roots} has for instance proposed to seed the first links in a crawl using human-selected URLs. However, these tailored sources raise challenges for practioners. First, individual corpora are much less scalable than massive web crawls, as they require scattered work instead of a centralized pipeline. Second, the providers of some of these sources have begun taking steps to forbid LLMs from being trained on their data \cite{Paresh_2023}, and adequate licensing may be costly to obtain. Based on our findings from the previous section, we may wonder what happens when we combine curated data with a strong web baseline like RefinedWeb.

\begin{mdframed}[style=side]
\textbf{Question.} When added in substitution of a strong web baseline, is curated data from individual corpora still beneficial to the natural language zero-shot performance of a model?
\end{mdframed}

\begin{table}[t]
\begin{center}
\caption{\textbf{We split curated data in three broad categories: conversations, books, and technical.} Individual components inspired by \citet{gao2020pile}, but and processed through our own data pipeline.}
\label{tab:data_categories}
\vspace{0.2in}
\begin{tabular}{lp{10cm}}
\toprule
\textbf{Conversations} & Reddit \citep{baumgartner2020pushshift}, HackerNews, OpenSubtitles \citep{tiedemann-2016-finding}, Ubuntu IRC, Youtube Subtitles, StackOverflow \\
\textbf{Books} & Project Gutenberg \cite{rae2019compressive} \\
\textbf{Technical} & ArXiv, PubMed Central, PubMed Abstracts, USPTO Patents \\
\bottomrule
\end{tabular}
\end{center}
\end{table}

\textbf{Methods.} We train small 1B models on 30B tokens, with the pretraining data split between web data and a specific curated category. We sample training on 1, 10, 25, 50, 75, and 100\% of the targeted category. We only consider a one-dimensional approach, and mix web data with a single category of curated data. We split our categories in books, conversations, and technical data as outlined in \cref{tab:data_categories}. For the individual corpora making these categories, we draw inspiration from The Pile \citep{gao2020pile} which we enhance with data from Reddit \citep{baumgartner2020pushshift} for the conversational category. Our web data is taken from RefinedWeb \citep{refinedweb} and we process curated sources through a similar pipeline, applying filtering and deduplication to make for a fair comparison. We evaluate performance on the \texttt{zs-data} aggregate--see \cref{tab:ablations_aggregates} for detailed make-up.

\begin{figure}[h]
\centering
\includegraphics[width=.6\textwidth]{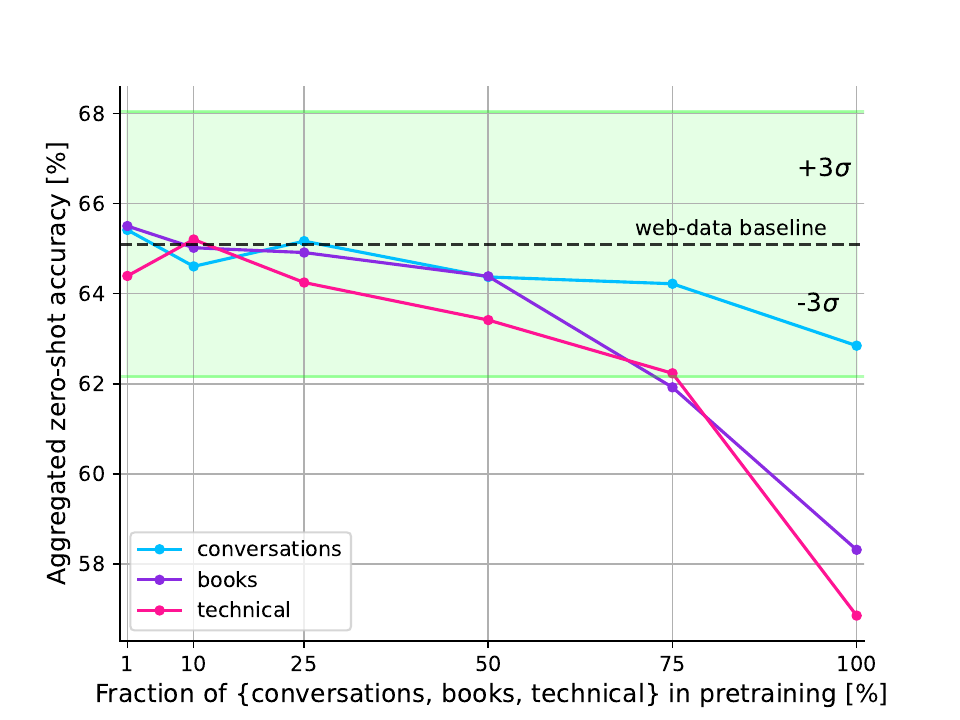}
\caption{\textbf{\data{When compared to a strong web data baseline, high-quality curated data does not improve zero-shot performance.}} Over-reliance on a single type of curated data is even detrimental to performance. The make-up of the conversations, books, and tech data corpora is outlined in \cref{tab:data_categories}. We report the zero-shot performance using different data-mixtures and compare with a baseline model trained on RefinedWeb only, our deduplicated and filtered web-only dataset \cite{refinedweb}. The shaded green-area indicates the $\pm3\sigma$ confidence interval based on 10 experiments across data splits.}
\label{fig:mixture-ablation}
\end{figure}

\textbf{Results.} We plot the zero-shot accuracy as we increase the fraction of curated data in \cref{fig:mixture-ablation}. When combined with a strong web baseline, we find that the addition of curated data never significantly uplifts performance. In fact, excess of curated data even worsens performance: for books and technical, past 50\%, we start observing meaningful degradation of accuracy. We believe this is likely caused by "mode collapse" compared to the high diversity of web data. 

Interestingly, conversations perform decently throughout, with the smallest degradation at 100\%. We posit this could either be due to our conversation category being the most diverse of the three, or to conversations being closer to the distribution of tasks. Indeed, this category likely includes people interacting, answering questions, and giving instructions to one another--a style which is less prevalent in books or technical-driven content like papers and patents.

Once again, this ablation is limited by the scope of the tasks considered. It is likely that highly technical tasks may benefit from domain specific data, and that web data would be a rather poor baseline for code tasks. Furthermore, our zero-shot evaluations are all done under a short context length: books could for instance be beneficial in helping the model learn long-range correlations.

\begin{mdframed}
\textbf{Finding.} When added in substitution of a strong web baseline, curated categories of data do not systematically result in an improvement in natural language zero-shot performance.
\end{mdframed}

\newpage
\subsubsection{Limited code and multilingual data do not strongly degrade English performance}
\label{sec:code_ml}

\smallskip

\begin{mdframed}[style=side]
\textbf{Question.} Can limited amounts (5-10\%) of code and multilingual data be substituted in pretraining data added without compromising the English performance of the model?
\end{mdframed}

\textbf{Multilinguality.} Some of the first large-scale deployment of RNNs and Transformers were for machine translation \citep{wu2016google}; with the attention mechanism originally introduced for such use cases \citep{bahdanau2014neural}. Accordingly, it's no surprise that multilingual language models rapidly flourished along their monolingual counterparts \citep{xue2021mt5}. However, multilingual \textit{generative} large language models have remained more elusive. Indeed, both \citet{XGLM} and~\citet{what_lang_model} have reported that massive multilinguality comes at the expense of English performance, resulting in multilingual models that underperform their monolingual counterparts \citep{bloom}. This so-called curse of multilinguality \citep{conneau2020unsupervised} has lead practitioners to be weary of naively adding other languages in bulk -- even PaLM \citep{palm}, which explicitly targets multilingual capabilities, only trains on 20\% non-English data. Furthermore, multilingual data is scarcely available \citep{costa2022no}: nearly 60\% of the documents in CommonCrawl are English, and the distribution of top languages is skewed towards European ones. Notably, Mandarin Chinese is only the 6th top language in CommonCrawl, despite being 2nd worldwide, and Hindi does not show-up even in the top-20 despite being 3rd worldwide~\citep{ethnologue}.

\begin{table}[b]
\centering
\vspace{-0.3in}
\caption{\textbf{We split languages in three categories: English-only, Restricted (adding the 3 most spoken languages in Europe), and European.} We only consider languages with a Latin alphabet, and sufficient presence in CommonCrawl to collect at least 10 billion tokens.}
\vspace{0.1in}
\label{tab:ml-groups}
\begin{tabular}{lp{11cm}}
\toprule
\textbf{Set}        & \textbf{Languages}                                                                                                                             \\
\midrule
\textbf{English}    & English                                                                                                                                        \\
\textbf{Restricted} & English, German, Spanish, French                                                                                                               \\
\textbf{European}   & English, German, Spanish, French, Italian, Dutch, Polish, Portuguese, Czech, Swedish, Romanian, Danish, Norwegian, Catalan, Slovene, Bulgarian \\
\bottomrule
\end{tabular}
\end{table}

We choose to experiment with a restricted multilingual setup: we consider languages with a Latin alphabet, and focus on those for which we can collect a non-trivial amount out of CommonCrawl (over 10 billion tokens) using our data pipeline. Our splits for experiments are in \cref{tab:ml-groups}. We train models with a fixed 10\% multilingual data (weighing individual languages according to their prevalence in CommonCrawl), and evaluate on English tasks. Each data split uses a dedicated tokenizer.

We present results in \cref{tab:ml-code-ablations}. We find that the performance degradation from $10\%$ multilinguality is very limited, and that the addition of other European languages over German, Spanish, and French do not drive additional degradation. We saw most of the reduction in performance on HellaSwag, while other tasks are not strongly impacted. We note that these experiments apply in a very restricted setting, and may not be illustrative of other multilingual setups (e.g., languages without a latin alphabet, using other languages to compensate for running out of English tokens, etc.)

\textbf{Code.} Large language models have demonstrated strong coding abilities, either through finetuning after general pretraining \citep{chen2021evaluating, palm, roziere2023code}, or through dedicated pretraining recipes \citep{li2023starcoder}. Furthermore, at variance with multilingual data, code data is plentiful \citep{kocetkov2022stack}, with trillions of tokens available from crawling public repositories. Code tasks are a predominant application of LLM-powered assistants \citep{luo2023wizardcoder}, and so allowing for such use cases with Falcon is important. However, we do not want to risk compromising language capabilities: we thus validate here the common practice of adding limited code data to pretraining, in line with other models (see \cref{tab:sota_mixes} for common fractions in pretraining).

We select the top-30 programming languages from GitHub, and substitute 5\% of our pretraining data for code. Note that we apply deduplication to the code dataset, but adjust our heuristics to avoid removing too much of the data. We consider only performance on English tasks. Results are outlined in \cref{tab:ml-code-ablations}. We find the addition of code data to have a similar effect to multilinguality (if only weaker, perhaps because of the smaller proportion), with only little to no task degradation.

\begin{mdframed}
\textbf{Finding.} Small fractions of code and multilingual data (5-10\%), in line with common recipes for large language models, do not broadly impact zero-shot performance on English tasks.
\end{mdframed}

We note that the small scale of our ablations is here a stronger limitation, likely leading us to a more conservative choice on multilinguality and code. It has been argued that larger models, thanks to their increased capacity, can better deal with multilinguality \citep{shaham2022causes}. Code data has also been shown for larger models to boost commonsense abilities \citep{madaan2022language}. More broadly, a similar effect has been observed for multilingual models \citep{aghajanyan2023scaling}.

\begin{table}[t]
\centering
\vspace{-0.3in}
\caption{\textbf{\data{Including in the pretraining data a small fraction of code or multilingual data broadly does not degrade performance significantly, except on specific tasks.}} Overall, the degradation observed is barely measurable on our larger aggregate, and mostly driven by HellaSwag on the \texttt{zs-small} one. \underline{Underlined} values have crossed the \emph{likely} 1-$\sigma$ degradation threshold.}
\vspace{0.1in}
\label{tab:ml-code-ablations}
\begin{tabular}{lcc}
\toprule
\textbf{Pretraining data} & \multicolumn{2}{l}{\textbf{Zero-shot accuracy}} \\ 
& \texttt{zs-main} $\uparrow$ & \texttt{zs-small} $\uparrow$\\
\emph{Likely} threshold (1-$\sigma$) & $\pm 1.0$ & $\pm 0.5$\\
\midrule
English-only & \textbf{53.7} & \textbf{49.2} \\
10\% Restricted  & 53.4 & \underline{48.3} \\
10\% European  & 53.6 & \underline{48.2} \\
\midrule
5\% Code & 53.6 & \underline{48.5} \\ 
\bottomrule
\end{tabular}
\end{table}

\subsection{Architecture and pretraining: validating popular recipes, and inference optimizations}
\label{sec:arch_ablations}

\subsubsection{Extending multiquery into multigroup for tensor parallel training and inference}
\label{sec:multiquery_abl}

\textbf{Background.} Unanimously, large language models first adopted the multihead attention scheme described in \citet{vaswani2017attention}. Each token produces $n_{\text{head}}$ triplets of (query, keys, and values), and the result of each head is then summed to produce the final output of the attention module. However, this scheme can be altered. \citet{multiquery} found that one can share the same keys and values between all attention heads with only a small degradation in performance. In this so-called multiquery attention, the number of heads for the queries remains $n_{q} = n_{\text{head}}$ but there is only one head for the keys and values, $n_{kv} = 1$ . This significantly reduces inference memory consumption: during autoregressive generation, the keys and values are cached to accelerate generation--with multiquery, the K,V-cache size is divided by $n_{\text{head}}$ compared to vanilla attention, resulting in a 10-100x reduction in memory consumption for common models--see \cref{tab:multiquery}. Multiquery improves the scalability of inference for large models \citep{pope2023efficiently}. \citet{palm} has recently popularized this architectural modification, which has notably been adopted by LLaMA-2 \citep{llama2}.

\textbf{Scaling.} Interestingly, multiquery is disproportionately effective for larger models. With $N$ the total number of parameters, $d_\text{model}$ the model size, $n_\text{layer}$ the number of layers, and assuming a fixed $d_\text{head} = d_\text{model} / n_{\text{head}}$--which is typically the case when using FlashAttention \citep{flash_attn}. For efficient scaling, it is recommended that $n_\text{layer} \sim \mathcal{O}(\log(N))$ \citep{levine2020limits}; since we can approximate $N \simeq n_\text{layer} (d_\text{model})^2$ \citep{kaplan2020scaling}, it follows that the size of the K,V-cache with multihead attention scales in $\mathcal{O}(\sqrt{N}\log(n))$. Conversely, for multiquery the K,V-cache only stores a fixed $2 d_\text{head}$ per layer, which does not increase with width; this results in overall more efficient scaling, in $\mathcal{O}(\log(N))$ instead.

\begin{table*}[t]
\begin{center}
\caption{\textbf{\hardware{Multiquery/group schemes significantly reduces the size of the K,V-cache for inference.}} Assuming TP=1 for Falcon-7B and TP=8 for Falcon-40/180B, sequence length 2,048.} 
\label{tab:multiquery}
\begin{tabular}{lcccccc}
\toprule
\textbf{Attention scheme} &  $n_{q}$ & $n_{kv}$ & \multicolumn{4}{l}{\textbf{K,V-cache for a 2,048 sequence}}\\
& & &  & 7B & 40B & 180B \\
\midrule
Vanilla & $n_{head}$ & $n_{head}$ & $\mathcal{O}(\sqrt{N}\log(n))$ & 1GB & 4GB & 10GB \\
Multiquery \citep{multiquery} & $n_{head}$ & 1 & $\mathcal{O}(\log(N))$ & 20MB & 30MB & 40MB \\
Multigroup \citep{grouped-query-attn} & $n_{head}$ & \texttt{TP} & $\mathcal{O}(\log(N))$ & N/A & 250MB & 335MB \\
\bottomrule
\end{tabular}
\end{center}
\end{table*}

\textbf{Multigroup.} One caveat of multiquery attention is that it is difficult to efficiently parallelize when relying on tensor parallelism, as is common with GPU-based infrastructure \citep{megatron-1}. Either each GPU keeps a copy of the shared key/value, recomputing them individually and then sharing gradients to keep them in sync, or they are computed on a single GPU and then communicated as necessary. We propose to introduce separate key/value pairs for each tensor parallel rank, simplifying the required communications. As in \citet{multiquery}, we keep $n_{q} = n_{\text{head}}$, but now have $n_{kv}=\texttt{TP}$. This scheme doesn't change the scaling of the K,V-cache, as it only applies a fixed TP factor. Concurrently to the development of the Falcon series, \citet{grouped-query-attn} also proposed this modification; we refer to this variant of attention as grouped query attention or multigroup. We note that the communication reduction applies not just during inference, but also during training.

\textbf{Results.} We train 1B/3B models on 30/60B tokens from The Pile \citep{gao2020pile}, with multiquery and varying degrees of multigroup attention. Importantly, we do not control for the reduction in parameters caused by the loss of additional keys and values--some degradation is thus expected. Results are presented in \cref{tab:kv}. Both multiquery and multigroup do not result in any large reduction in zero-shot performance, even without compensating for the reduction in trainable parameters.

\begin{mdframed}
\textbf{Recipe decision.} To improve performance scalability for the largest models, the Falcon series implement multigroup with KV=TP for all models (respectively 1/8/8 for Falcon-7/40/180B).
\end{mdframed}

\begin{table}[b]
\centering
\caption{\textbf{\hardware{Even without controlling for the reduction in parameters, multiquery only comes at a limited zero-shot performance cost.}} Impact on perplexity is more directly measurable, whereas impact on zero-shot performance is less consistent. Multigroup with KV=8 consistently performs close to the vanilla baseline. \underline{Underlined} values have crossed the \emph{very likely} 2-$\sigma$ degradation threshold.}
\label{tab:kv}
\vspace{0.2in}
\centerline{
\begin{tabular}{cccccc}
\toprule
\textbf{Model size} & \textbf{KV} & \multicolumn{3}{l}{\textbf{Performance}} \\
& & \texttt{zs-main} $\uparrow$ & \texttt{zs-small} $\uparrow$ & \texttt{ppl-pile} $\downarrow$\\
\multicolumn{2}{l}{\emph{Very likely} threshold (2-$\sigma$)} & $\pm 2.2$ & $\pm 0.8$ & $\pm 0.005$\\
\midrule
1B & 1 & 48.5 & 42.6 & \underline{0.908} \\
 & 2 & 48.2 & \underline{42.1} & 0.899 \\
 & 4 & 48.9 & 42.4 & \underline{0.908}\\
 & 8 & 48.6 & 42.8 & \underline{0.903} \\
& Vanilla & \textbf{49.2} & \textbf{43.1} & \textbf{0.895} \\ \midrule
3B & 1 & \underline{52.7} & \underline{48.6} & \underline{0.825}\\
 & 8 & \textbf{54.6} & \textbf{50.1} & \underline{0.819} \\
 & Vanilla & 54.4 & 49.8 & \textbf{0.807} \\
\bottomrule
\end{tabular}}
\end{table}

\subsubsection{Rotary positionnal embeddings may only offer a limited edge over ALiBi}

\textbf{Background.} By default, attention does not provide positional information to the model: it only sees the sequence as a bag-of-word. Accordingly, the original Transformer architecture adopted absolute sinusoidal embeddings to encode positional information \citet{vaswani2017attention}. However, absolute embeddings have since declined in popularity, the community shifting to relative embeddings instead. While this shift is well motivated empirically \citep{relative_pos, what_lang_model}, practionners have yet to crystallize on a single relative positional embedding: BLOOM and MPT \citep{bloom, MPT30B} use ALiBi \citep{press2021alibi}, while GPT-J, PaLM, and LLaMA \citep{gpt-j, palm, llama, llama2} use Rotary Positional Embeddings (RoPE) \citep{su2021roformer}. RoPE are often cited as delivering better upstream performance in the works above, while ALiBi benefits from built-in extrapolation abilities. Another recently introduced alternative are Universal Relative Positonal Embeddings, URPE \citep{urpe}, which address shortcomings in the expressivity of typical relative positional embeddings. As a curiosity, we note that the autoregressive mask of a causal model also provides some positionnal information to the model \citep{what_lang_model, haviv2022transformer}, enabling training without any positional embeddings to be comparable to absolute sinusoidal ones for zero-shot performance.

Concurrently to this work, recipes have emerged to enable zero-shot or finetuned length extrapolation with RoPE \citep{chen2023extending}, briding the gap with ALiBi for extrapolation.

\begin{table}[t]
\centering
\caption{\textbf{\hardware{Although at small-scale URPE and RoPE may likely be better than ALiBi, that advantage isn't as clear at increased size.}} We find ALiBi to be likely worst than rotary on two of our three aggregates for 1B models, but to be closer to the performance of rotary at the 3B scale. \underline{Underlined} values have crossed the \emph{likely} 1-$\sigma$ degradation threshold over RoPE.}
\label{tab:rotary}
\vspace{0.2in}
\centerline{
\begin{tabular}{cccccc}
\toprule
\textbf{Model size} & \textbf{Pos. Emb.} & \multicolumn{3}{l}{\textbf{Performance}} \\
& & \texttt{zs-main} $\uparrow$ & \texttt{zs-small} $\uparrow$ & \texttt{ppl-pile} $\downarrow$\\
\multicolumn{2}{l}{\emph{Likely} threshold (1-$\sigma$)} & $\pm 1.1$ & $\pm 0.4$ & $\pm 0.002$\\
\midrule
1B & ALiBi & 49.2 & \underline{43.1} & \underline{0.895} \\
 & URPE & 49.6 & \underline{43.1} & 0.885 \\
 & RoPE & \textbf{50.0} & \textbf{44.2} & \textbf{0.883}\\ \midrule
3B & ALiBi & \textbf{54.4} & \underline{49.8} & \underline{0.807}\\
 & RoPE & \textbf{54.4} & \textbf{50.5} & \textbf{0.799} \\
\bottomrule
\end{tabular}}
\vspace{-0.2in}
\end{table}

\textbf{Results.} We train 1/3B models on 30/60B tokens on The Pile \citep{gao2020pile}. We report results in \cref{tab:rotary}. We find no evidence for URPE outperforming RoPE--since it would require significant modifications to the fused-attention kernels to deliver acceptable performance, we do not pursue it further. At the 1B scale, we find a likely advantage to using RoPE over ALiBi; however, that advantage diminishes at the 3B scale, and is insufficient to conclude clearly. One remaining advantage of ALiBi is its compute overhead: it is significantly cheaper to compute than RoPE; however, with custom Triton kernels (\cref{other_kernels}) we are able to mitigate that overhead. 

\begin{mdframed}
\textbf{Recipe decision.} In-line with other popular large-scale models, we adopt rotary positionnal embeddings, and use custom kernels to mitigate the overhead.
\end{mdframed}

\subsubsection{The extra memory cost of GLU may not be worth it for cost-efficient training}

\textbf{Background.} Activations based on gated linear units \citep{shazeer2020glu} are widely believed to outperform traditional activation functions such as GeLU \citep{what_lang_model}. They have seen adoption in models such as PaLM and LLaMA \citep{palm, llama, llama2}. 

\textbf{Scaling.} Scaling-wise, GLU activations have been preferred as well, as they increase the size of the MLP (doubling its first layer), shifting more compute towards simple matrix multiplications. However, this does come at a cost: the memory required to store the intermediary activations in the MLP is higher. Remember that, typically, the inputs to the activation function are saved for the backward (as recomputing the function itself is negligible). For gated units, this input is now twice large. Overall, SwiGLU doubles intermediary activations, and increases by 50\% the number of parameters in the MLP: the activation memory per parameter for the MLP is thus increased by 33\%.

\textbf{Results.} In \cref{tab:tweaks}, training 1B models on 30B tokens of The Pile, we find no clear benefits from adopting SwiGLU for zero-shot performance--and the improvement on perplexity is just at the threshold of being unlikely to characterize an actual improvement due to variance in the evaluation setup. We note that this could be due to the fact SwiGLU may require dedicated hyperparameters tuning--as for all architecture ablations, we simply adopted the ones of GPT-3 \citep{brown2020gpt3}. Furthermore, we saw little additional throughput gains from SwiGLU (as we already use parallel attention/MLP layers). Since we train the Falcon series on 40GB A100 to optimize costs, we were concerned early on about memory consumption--accordingly, SwiGLU, with its increased memory intensity, and similar performance to GeLU, would likely be a net negative for us. 

\begin{mdframed}
\textbf{Recipe decision.} Out of concern for the memory footprint of our trainings on A100-40GB, and because of no clear uplift in zero-shot, we choose not to adopt SwiGLU.
\end{mdframed}

\begin{table}[t]
\centering
\caption{\textbf{\hardware{Small architectural tweaks like GLU, z-loss, and removing biases are unlikely to improve zero-shot performance.}} However, in some scenarios, these have been proposed to improve scalability and/or stability, which may warrant their adoption. \underline{Underlined} values have crossed above the \emph{unlikely} 0.4-$\sigma$ improvement threshold over our baseline.}
\label{tab:tweaks}
\vspace{0.2in}
\centerline{
\begin{tabular}{cccccc}
\toprule
\textbf{Model size} & \textbf{Pos. Emb.} & \multicolumn{3}{l}{\textbf{Performance}} \\
& & \texttt{zs-main} $\uparrow$ & \texttt{zs-small} $\uparrow$ & \texttt{ppl-pile} $\downarrow$\\
\multicolumn{2}{l}{\emph{Unlikely} threshold (0.4-$\sigma$)} & $\pm 0.4$ & $\pm 0.2$ & $\pm 0.001$\\
\midrule
1B & & \textbf{49.2} & 43.1 & 0.895 \\
 & SwiGLU & \textbf{49.2} & 43.1 & \textbf{\underline{0.891}} \\
 & z-loss & 49.0 & \textbf{\underline{43.6}} & 0.895 \\\midrule
 3B &  & \textbf{54.5} & \textbf{49.8} & \textbf{0.807} \\
 & No biases & 54.4 & \textbf{49.8} & \textbf{0.807} \\
\bottomrule
\end{tabular}}
\end{table}

\subsubsection{Small tweaks help scalability: parallel layers and no biases in linear layers}

\begin{figure}[t]
    \centering
    \begin{tikzpicture}[
dot/.style = {circle, fill, minimum size=#1,
                inner sep=0pt, outer sep=0pt},
               scale=0.30, every node/.style={scale=0.30}
]]
\draw[fill=architecture,rounded corners] (0, 0) rectangle (4, 4);
\node[] at (2.0, 2.0) {\Huge Attention};
\draw[fill=objective,rounded corners] (0.0, 8.0) rectangle (4.0, 12.0);
\node[] at (2.0, 10.0) {\Huge MLP};
\draw[-{Latex[width=1.5mm]},] (-2.0, -3.0) to [bend left = 0] (-2.0, 15.0);
\draw[-{Latex[width=1.5mm]},rounded corners] (-2.0, -2.0) -| (2.0, 0.0);
\draw[-{Latex[width=1.5mm]},rounded corners] (2.0, 4.0) |- (-2.0, 5.7);
\draw[-{Latex[width=1.5mm]},rounded corners] (-2.0, 6.3) -| (2.0, 8.0);
\draw[-{Latex[width=1.5mm]},rounded corners] (2.0, 12.0) |- (-2.0, 14.0);
\node[right] at (-2.0, 5.2) {\huge+};
\node[right] at (-2.0, 13.5) {\huge+};
\draw[fill=architecture,rounded corners] (10.0, 4.0) rectangle (14.0, 8.0);
\node[] at (12.0, 6.0) {\Huge Attention};
\draw[fill=objective,rounded corners] (16.0, 4.0) rectangle (20.0, 8.0);
\node[] at (18.0, 6.0) {\Huge MLP};
\draw[-{Latex[width=1.5mm]},] (8.0, -3.0) to [bend left = 0] (8.0, 15.0);
\draw[-{Latex[width=1.5mm]},rounded corners] (8.0, 2.0) -| (12.0, 4.0);
\draw[-{Latex[width=1.5mm]},rounded corners] (12.0, 8.0) |- (8.0, 9.7);
\draw[-{Latex[width=1.5mm]},rounded corners] (8.0, 1.7) -| (18.0, 4.0);
\draw[-{Latex[width=1.5mm]},rounded corners] (18.0, 8.0) |- (8.0, 10.0);
\node[right] at (8.0, 9.2) {\huge+};

\end{tikzpicture}
    \caption{\hardware{\textbf{Parallelizing the attention and MLP blocks allows us to remove one sync point during tensor parallel training.}} This was first proposed by \citet{gpt-j} for GPT-J.}
    \label{fig:parllel_attn}
    \vspace{-0.2in}
\end{figure}
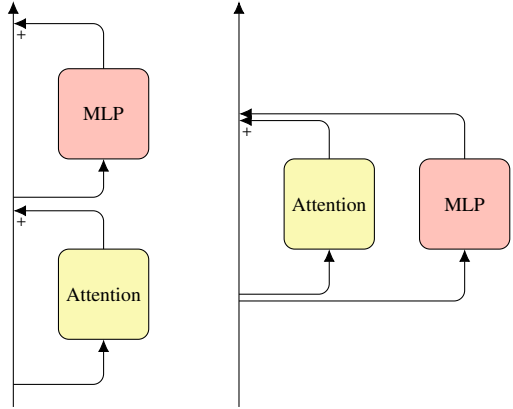

\textbf{Parallel attention and MLP blocks.} \citet{gpt-j} first introduced parallel attention and MLP layers while training GPT-J. This augmentation is important to reduce the communication costs associated with tensor parallelism: this simple modification cuts the number of \texttt{all\_reduce} necessary from two to one per layer. We found no measurable degradation in zero-shot performance or perplexity, in line with \citet{palm}, and adopt this practice. See \cref{fig:parllel_attn} for an illustration.

\textbf{No biases.} \citet{palm} found that removing the biases in the linear layers and layer norms improves stability. We validate that removing the biases in the linear layer does not result in worse performance (see \cref{tab:tweaks}): neither in terms of language modeling loss nor in terms of the final zero-shot performance. Accordingly, we remove biases from the linear layers in the Falcon series.

\begin{mdframed}
\textbf{Recipe decision.} We adopt parallel attention and MLP, and remove biases from linear layers.
\end{mdframed}

\subsubsection{Validating best practices for hyperparameters: z-loss, weight decay, LR search}
\label{sec:hyperparam_abl}

\textbf{Z-loss.} First introduced in the \texttt{mesh-tensorflow} codebase\footnote{As a fun exercise for the reader, we encourage trying to dig up that citation, starting from PaLM \citep{palm}, which popularized this practice outside of Google by outlining it in its main text.} \citet{meshtf}, z-loss aims at increasing the stability of training, by encouraging the logits to stay close to zero. It can be implemented as an auxiliary loss: z\_loss=$10^{-4} \text{ log}^2(\Sigma_i e^{z_i})$, where $z_i$ is the output logits of the model. Note that z-loss does not have a significant impact on task performance at small scale (\cref{tab:tweaks}).

\begin{mdframed}
\textbf{Recipe decision.} We adopt z-loss, as it is claimed to improve large-scale training stability and does not impact zero-shot performance in our ablations.
\end{mdframed}

\newpage
\textbf{Weight decay.} We attempted to reproduce the weight decay schedule from \citet{palm}, but failed to obtain an improvement--we suspect this is due to differences in initialization. We found that weight decay has a disproportionate effect on datasets which have not been deduplicated/are of lower quality (\cref{tab:decay}). We use AdamW for weight decay \citep{loshchilov2018decoupled}.

\begin{mdframed}
\textbf{Recipe decision.} We use a fixed weight decay of 0.1 with AdamW for all Falcon models.
\end{mdframed}

\begin{table}[b]
\centering
\vspace{-0.2in}
\caption{\textbf{\data{Weight decay is likely to improve performance, especially so for datasets such as The Pile, which may not have been adequately deduplicated.}} Surprisingly, the effect of weight decay is disproportionately strong depending on the underlying dataset.}
\label{tab:decay}
\vspace{0.1in}
\centerline{
\begin{tabular}{ccccccccc}
\toprule
\textbf{Dataset} & \textbf{Weight decay} & \multicolumn{3}{l}{\textbf{Performance}} \\
& & \texttt{zs-main} $\uparrow$ & \texttt{zs-small} $\uparrow$ & \texttt{ppl-pile} $\downarrow$\\
\multicolumn{2}{l}{\emph{Likely} threshold (1-$\sigma$)} & $\pm 1.1$ & $\pm 0.4$ & $\pm 0.002$\\
\midrule
RefinedWeb & 0. & \textbf{52.1} & 47.9 & 1.07 \\
 & 1. & 52.0 & \underline{\textbf{48.4}} & \underline{\textbf{1.06}} \\ \midrule
The Pile & 0. & 50.3 & 43.7 & 0.877 \\
 & 1. & \underline{\textbf{51.7}} & \underline{\textbf{45.0}} & \underline{\textbf{0.868}} \\
\bottomrule
\end{tabular}}
\end{table}

\textbf{Optimal learning rate.} Practices for setting the learning rate of a run differs, from naive grid search to more principled approaches \citep{yang2022tensor, dinan2023effective}. We further discuss in \cref{sec:donotwork} our failures to implement and reproduce some of the later; in this short section, we focus on the naive approach and our validation of it. Broadly speaking, setting too high of a learning rate risks to cause divergence of the run and instabilities during training; too low, on the opposite, will leave some upstream and downstream performance on the table, leading to inefficient training. 

We propose to search through possible learning rate in the following away: (1) we select 4-6 roughly logarithmically-spaced candidate learning rates, anchoring around the ones used in GPT-3 \citep{brown2020gpt3}, and favoring higher learning rates; (2) we run through a long 500 million tokens warmup for all candidate learning rates; (3) we pick the learning rate which has achieved the lowest loss at this point, and discard any learning rate which has already caused spikes. 

We test this method at small scale, picking 7 learning rates for a 1B model, and then comparing the ranking obtained after warmup against the actual final ranking achieved. Results are presented in \cref{tab:lr_search}. We find that this naive method succeeds at finding the best learning rate from the rankings at the end of warm-up. More broadly, ranks at the end of warm-up and at the end of training are relatively stable, with only the 2nd and 3rd best switching places.

\begin{mdframed}
\textbf{Recipe decision.} From candidates LRs, we pick the one with the lowest loss after warm-up.
\end{mdframed}

\begin{table}[t]
\caption{\textbf{\hardware{Loss rankings at the end of learning rate warm-up broadly reflects rankings at the end of training, enabling us to search for optimal learning rates efficiently}} This simple heuristic is easy to use, and consume only a fraction of resources for a larger run.}
\vspace{0.1in}
\label{tab:lr_search}
\centering
\centerline{
\begin{tabular}{ccccccc}
\toprule
\multicolumn{2}{c}{\textbf{Learning rate}} & \multicolumn{2}{c}{\textbf{End LR warm-up [0.5GT]}} & \multicolumn{2}{c}{\textbf{End of run [27GT]}} & \textbf{Run stability}\\ 
Factor & LR & Improv. $\downarrow$ & Rank & Improv. $\downarrow$ & Rank\\
\bottomrule
 x1 & $2\times10^{-4}$&  $\pm 0.0\%$ & (4) & $\pm 0.0\%$ & (4) & No spikes \\
 x2 & $4\times10^{-4}$&  \underline{$-2.0\%$} & (2) & $-1.7\%$ & (3) & No spikes\\
 x5 & $1\times10^{-3}$&  $\mathbf{-2.6\%}$ & (1) & $\mathbf{-2.5\%}$ & (1) & No spikes\\
 x10 & $2\times10^{-3}$& $-1.4\%$ & (3) & \underline{$-1.9\%$} & (2) & One small spike\\
 x20 &  $4\times10^{-3}$& $+1.3\%$ & (5) & $+1.9\% $& (5) & Multiple spikes\\
 x50 & $1\times10^{-2}$& $+7.6\%$ & (6) & $+6.8\%$ & (6) & Multiple large spikes\\
 x100 & $2\times10^{-2}$& \multicolumn{4}{c}{Diverging after 0.2GT} \\
 \bottomrule
\end{tabular}
}
\end{table}

\subsection{Further experimentation required: ideas that did not make the cut}
\label{sec:donotwork}

In this section, we briefly mention some practices and ideas we experimented with, but for which we were unable to reproduce results or obtain a meaningful outcome. We note that this is not be viewed as an indictment of these practices: many have been adopted in popular models successfully. 

\textbf{Alternative training objectives.} The largest of language models have been typically trained with a causal decoder-only architecture and objective \citep{brown2020gpt3, gopher, palm}. \citet{wang2022language} found that such models exhibit better zero-shot abilities than masked encoder-decoders such as T5 \citep{raffel2019t5}; however, they also found that after multitask finetuning~\citep{Sanh2021MultitaskPT}, masked encoder-decoder performed better, highlighting that different regimes and use cases may favor different architectures and objective. Throughout, a non-causal decoder-only (so-called prefix language model) performed competitively as a close second. With UL2, \citet{tay2022ul2} found that these paradigms could be unified, by training on a mixture of objectives instead. We experimented with UL2, but were unable to obtain an uplift in zero-shot performance, even after adapting tasks to the various paradigms when relevant. Due to time and ressource constraints, we did not end-up pushing our experiments further, as \citet{tay2022transcending} showed that a posteriori adaptation to the UL2 objective was not only possible but efficient.
% We also experimented with forgetful causal masking \citep{liu2022forgetful}, and observed similar outcomes.

For code models, so-called fill-in-the-middle (FIM) training \citep{bavarian2022efficient} has been popular, as it addresses a common use cases for such models. FIM is claimed to come at little to no expense of autoregressive modeling capabilities; we were broadly able to confirm these results, showcasing likely degradation only for a handful of tasks for intermediary infilling rates (0.25-0.5). Low and high infilling rates had the lowest effect on zero-shot performance. Nevertheless, due to lack of wide adoption at the time, we choose to skip FIM, and to instead consider it as an adaptation step for a Falcon-Coder model--this was concurrently demonstrated for Code-LLaMA \citep{roziere2023code}. 

\textbf{Principled hyperparameters.} We experimented with $\mu$-parametrization \citep{yang2022tensor} as a way to scale our hyperparameters in a principled way from smaller to larger runs. Unfortunately, we were unable to demonstrate an improvement over our naive strategy of hyperparameters estimation.

\textbf{Alternative optimizers.} Encouraged by its strong reported results on BERT and T5, we experimented with Amos \citep{tian2022amos}, an alternative to Adam with adaptive learning rate and weight decay. We were unable to obtain an improvement for causal decoder-only models, even at small scale. 

\textbf{Conversation mining.} In developing RefinedWeb, we experimented with the idea of mining specific types of data from the web. Training dedicated classifiers based on BERT models often resulted in over-fitting, so we instead used simple heuristics (e.g., identifying arguments by the density of transition words, conversations by finding turns of users). We were able to obtain significant zero-shot performance uplift, however this data was very scarce. Since training of the models was started before we had processed all of CommonCrawl, we could not effectively surface all of that data and use it in priority over standard web data, leading us to leave that idea for future models.

\subsection{Wrapping-it up: validating overall dataset and architecture recipes}
\vspace{-0.1in}

For convenience, we refer readers to Section ? for a full overview of the Falcon recipe. We will now validate the performance of the recipe at larger scale, with a longer training run and comparisons with models from the state-of-the-art. We independently verify: (1) the pretraining dataset, in particular its web component; (2) the architecture, by training a model adopting it on The Pile. 

\textbf{Dataset validation.} We train 1B and 7B parameters models for 27B tokens and 350B tokens, to reproduce common practices from previous models. We use our baseline architecture, which is based on GPT-3 \citep{brown2020gpt3} with ALiBi \citep{press2021alibi}. We train on The Pile \citep{gao2020pile}, RefinedWeb (our web dataset, \citet{refinedweb}), and the Falcon data mixture which combines RefinedWeb and curated sources without any upsampling. This last mixture is designed for Falcon-180B, and targets a total of 3,500B tokens. See \cref{sec:implem_data} for details.

In \cref{tab:abl_validation}, we find our data mixture significantly uplifts performance. The large majority of these gains are also achieved by RefinedWeb alone, without the curated data. This highlights our previous finding that web data alone, when adequately filtered and deduplicated, can train performant models. We find our 1/7B model compares favorably with other models from the state-of-the-art, however we note our setup put it at a small advantage by training for slightly longer.

\textbf{Architecture validation.} We follow the set-up of our dataset validation, and train the architecture validation 1B models on The Pile for 27B and 350B tokens. Note that we do not include in this experiment hyperparameters tweaks: all models use weight decay, and the same learning rate from~\citet{brown2020gpt3}--this experiment concerns only the architecture of the models itself. 

We find our architecture comes with a small performance degradation (\cref{tab:abl_validation}), which we mostly attribute to the reduction in parameters caused by multiquery. We suspect that growing up the multiquery model would likely close that gap. Nevertheless, it's interesting to note that while data improvements have very significant effects, architecture improvements are mostly focused on improving training and inference scalability; they do not result in an uplift in task performance.

\begin{table}[h!]
\centering
\caption{\textbf{\data{Our data recipe, predominantly based on our work on RefinedWeb \citep{refinedweb}, significantly improves upon The Pile and other models from the state-of-the-art.} \hardware{Because multiquery makes models smaller (and we do not control for that effect), our architecture comes with a small zero-shot performance degradation.}} We suspect controlling for parameter count would bring our architecture to be on-par or better than the baseline. Nevertheless, improvements to the architecture mostly deliver improvements in hardware scalability for inference and training, while improvements to the data recipe significantly uplift the downstream performance of models. \underline{Underline} for relevant changes, \textbf{bold} for improvement over baseline, \emph{italics} for degradation over baseline. $^\dagger$ flags independent evaluations with the EleutherAI Harness \citep{eval-harness}, and $^\text{a}$ indicates our architecture run is better, while $^\text{d}$ shows our data run is better.}
\label{tab:abl_validation}
\vspace{0.1in}
\centering
\begin{small}
\begin{tabular}{ccccccccccc}
\toprule
\textbf{Scale} & \textbf{Dataset} & \textbf{Architecture} & \multicolumn{4}{l}{\textbf{Performance}} \\
& & & \texttt{zs-main} $\uparrow$ & \texttt{zs-comp} $\uparrow$ & \texttt{zs-small} $\uparrow$ & \texttt{ppl-pile} $\downarrow$\\
\midrule
\multirow{5}{*}{1B@27GT} & The Pile & Baseline & 51.7 & 40.3 & 45.0 & 0.868 \\ \cmidrule{2-7} 
 & \underline{Falcon} & Baseline  & \textbf{53.5} & 42.3  & \textbf{48.8} &  \\
 & \underline{RefinedWeb} & Baseline & 53.2 & \textbf{43.4} & 48.4 &  \\ \cmidrule{2-7}
 & The Pile & \underline{Falcon} & \textit{51.1} & \textit{40.0} & \textbf{45.1} & \textit{0.870} \\ \midrule \midrule
 \multirow{4}{*}{1B@350GT} & The Pile & Baseline & 57.8 & 47.1 & 54.0 & 0.763 \\ \cmidrule{2-7} 
  & \underline{RefinedWeb} & Baseline & \textbf{59.8} & \textbf{50.1} & \textbf{55.7} & \\ \cmidrule{2-7} 
   & The Pile & \underline{Falcon} & \textit{56.6} & \textit{46.1} & \textit{52.7} & \textit{0.775} \\ \midrule
   1B@300GT & OpenAI & \texttt{babbage}$^\dagger$ & & 47.8$^\text{d}$ & & \\
   1B@380GT & The Pile & GPT-Neo$^\dagger$ & & 44.3$^\text{a,d}$ & & \\
   1B@300GT & The Pile & BS-A\&S$^\dagger$ & & 46.1$^\text{d}$ & & \\
   1B@300GT & The Pile & Pythia$^\dagger$ & & 45.2$^\text{a,d}$ & & \\
\midrule \midrule
7B@350GT & \underline{RefinedWeb} & Baseline & & \textbf{55.3} & & \\
6B@300GT & OpenAI & \texttt{curie}$^\dagger$ & & 53.7$^\text{d}$ & & \\
6B@400GT & The Pile & GPT-J$^\dagger$ & & 53.5$^\text{d}$ & & \\
\bottomrule
\end{tabular}
\vspace{-0.2in}
\end{small}

\end{table}

\section{Implementation}
\label{sec:implementation}

Based on our findings from the previous ablations \cref{sec:ablations}, and further tests and best practices from the literature, we now describe the codebases and methods used to train the Falcon series of models.  

\subsection{The Falcon dataset: predominantly web, with added curated and conversational data}
\label{sec:implem_data}

Based on estimates of our compute budget of 30,000-50,000 PF-days, we target a pretraining dataset size in the range of \textbf{3,000-5,000 billion tokens}--we use \citet{hoffmann2022training} as an upper boundary for model size and lower boundary for pretraining length. This is more than 2x the size of the dataset for Chinchilla, and 10x the one for GPT-3; although this range of size is recently becoming more common in concurrent models like LLaMA-2 or OLMo \citep{llama2, dolma}. Out of concerns for memorization and degradation caused by repeating data \citep{carlini2022quantifying, hernandez2022scaling} we choose to \textbf{not upsample any sources}. 

\textbf{High-level overview.} In \cref{sec:webdata_exp}, we have shown that sufficiently filtered and deduplicated web data can deliver performant models: this leads to focus on scaling-up web data to achieve the scale necessary. Because improvements to data quality translate to significant improvements to downstream performance, and because data processing is tremendously cheaper than model training, we do not concern ourselves too much with optimizing for costs with data processing--it is likely that a 90\% cheaper recipe with less than a 10\% relative performance degradation could be found. 

We still include a small amount of curated data, inspired by The Pile \citep{gao2020pile} with the addition of conversations from Reddit \citep{baumgartner2020pushshift}, as it is unlikely to degrade performance if adequately processed (\cref{sec:curated_exp}) and as it may broaden the downstream applicability of the model. However, these sources are bound to remain a minority, given that we do not allow any upsampling--they end-up accounting for 13\% of our final dataset. Regarding code and multilinguality, we take a conservative approach: based on our results in \cref{sec:code_ml}, we include 8\% multilingual data and 3\% code. These lower fractions than the ones experimented with are due to stock constraint; specifically for code, further improvements to our pipeline enabled us to significantly scale availability, but this was after the models had started training, so we did not revise the mix.

The final Falcon mixture is presented in \cref{tab:falcon-mixture}. We designed the mixture based on a 3,500B tokens pretraining dataset, not allowing any upsampling of the curated sources. Despite differing training lengths, the same mixture (in \%) is used for Falcon-7B, 40B, and 180B. 

\begin{table}[b]
\caption{\data{\textbf{The final Falcon mixture is predominantly web-based (nearly 85\%), but includes other curated corpora (without any upsampling) to broaden the expressiveness of the model.}} Individual curated corpora are inspired from The Pile \citep{gao2020pile}, but rebuilt from scratch to ensure high-quality and compatibility with our data pipeline. Code stock is a rough estimate based on an updated pipeline; code data used in Falcon was sourced from permissively licensed GitHub repositories. Mixture was designed to avoid upsampling; note that total stocks for RefinedWeb were not known at the beginning of training, as processing was still in-progress. Quantities in tokens.}
\vspace{0.2in}
\centering
\label{tab:falcon-mixture}
\begin{tabular}{lp{5cm}ccc}
\toprule
\textbf{Corpora}    &  & & \multicolumn{2}{l}{\textbf{Pretraining}}      \\
Name                & Source          & Stock & Fraction             & Used \\ \midrule
\textbf{RefinedWeb-English} & Filtered and deduplicated CommonCrawl, see \citet{refinedweb}                 & $\sim$5,000B       & 76\%                      & 2,700B      \\
\textbf{RefinedWeb-Euro} & Filtered and deduplicated multilingual (Europe-focused) CommonCrawl, see \citet{refinedweb}                  & $\sim$2,000B      & 8\%                     & 400B     \\
\textbf{Books}   & Project Gutenberg                &  215B      &  6\%                    & 214B     \\
 \textbf{Conversations}                   & Reddit, StackOverflow, HackerNews, IRC, YouTube Subtitles                & 170B       & 5\%                     & 168B     \\
   \textbf{Code}                   & GitHub              & $\sim$1,000B       & 3\%                     & 115B     \\
  \textbf{Technical}                   & arXiv, PubMed, USPTO, Wikipedia              & 60B       & 2\%                     & 57B     \\ \bottomrule
\end{tabular}
\end{table}

\subsubsection{The Macrodata Refinement pipeline and the RefinedWeb dataset}

\smallskip

\begin{mdframed}[style=side]
Our web data processing pipeline is extensively described in the dedicated paper \data{\textbf{\textsc{RefinedWeb}}} paper \citep{refinedweb}. In this section, we only highlight key components and decisions.
\end{mdframed}
\vspace{-0.1in}

\begin{figure}[b]
\centering
\includegraphics[width=0.7\linewidth]{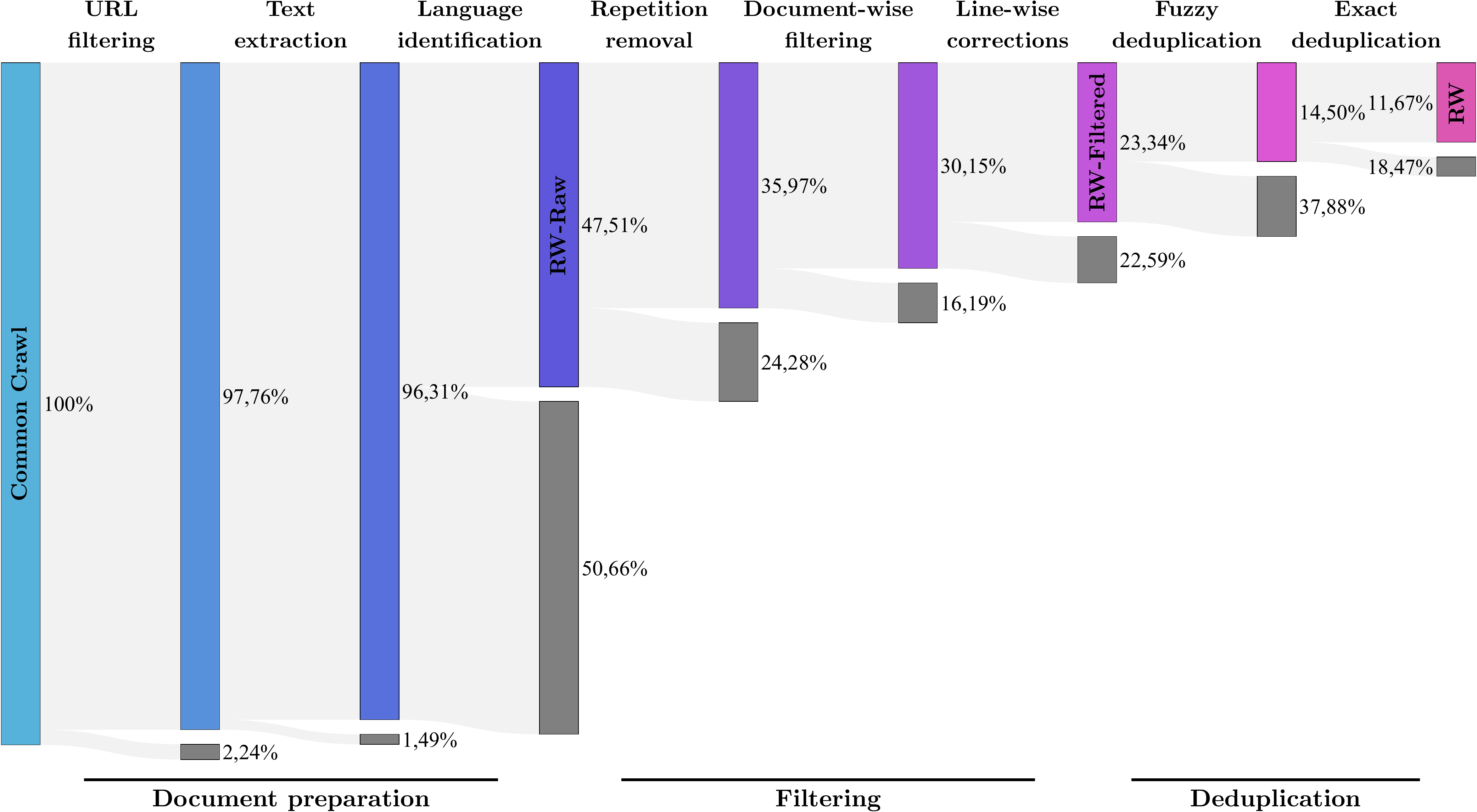}
\caption{\data{\textbf{Subsequent stages of Macrodata Refinement remove nearly 90\% of the documents originally in CommonCrawl.}} Notably, filtering and deduplication each result in a halving of the data available: around 50\% of documents are discarded for not being English, 24\% of remaining for being of insufficient quality, and 12\% for being duplicates. We report removal rate (\textcolor{gray}{grey}) with respect to each previous stage, and kept rate (\data{shade}) overall. Figure from \citet{refinedweb}.}
\label{fig:mdr_pipeline}
\end{figure}

To scale-up pretraining data, two approaches are possible:
\begin{itemize}
    \item \textbf{Repeat data.} This is the easiest option, and was the norm in computer vision originally. However, most large language models have been far more conservative, usually only upsampling specific corpora for 2-6 times (see \cref{tab:sota_mixes}). This is largely due to concerns with memorization \citep{carlini2022quantifying} and with deduplicates disproportionately degrading the quality of models \citep{lee2022deduplicating,hernandez2022scaling}. Recently, \citet{data_constrained} has argued that while up to 4 epochs on the same data may be acceptable, further repetition will cause degradation--this leads us to eschew this strategy.
    \item \textbf{Scale-up web data processing.} While scaling curated sources is cumbersome and requires extensive manual work, web data is a massive, plentiful source. Improvements to web data have high leverage, as they impact a large amount of tokens at once: public crawls such as CommonCrawl may contain in excess of 50-100 trillion tokens, such that even a 90\% rejection rate would result in a trillion-scale dataset. However, raw web data is also of extremely poor quality \citep{trinh2018simple, kreutzer2022quality}, containing large amounts of undesirable adult content and machine generated spam. We choose to focus our work on improving the quality of web data, through large-scale filtering and deduplication.
\end{itemize}

These approaches are orthogonal, but not antagonist; scaling to frontier models, with pretraining datastes of 10-100 trillion tokens, will likely require repeating massive web datasets for a few epochs. 

\textbf{Philosophy.} RefinedWeb differentiates itself from previous web datasets in the following ways:

\begin{itemize}
    \item \textbf{Extreme-scale.} Our Macrodata Refinement pipeline focuses on scalability: we used up to 20,000 CPU cores to produce RefinedWeb. With nearly five trillion deduplicated tokens, the RefinedWeb dataset is the largest documented pretraining dataset, supporting the training of larger models than previously though possible without relying on multiple epochs.
    \item \textbf{Stringent deduplication and filtering.} Inspired by \citet{lee2022deduplicating}, RefinedWeb is fully deduplicated. Fuzzy deduplication with MinHash is used to first massively shrink the dataset, and then extract substring deduplication is applied. Filtering heuristics are also first used to reduce text extraction artefacts, and to remove machine-generated content. We find in \cref{sec:webdata_exp} that this allows web data to match curated corpora.
    \item \textbf{Neutral filtering.} With the exception of language identification, the Macrodata Refinement pipeline does not rely on ML-based filtering strategies. Indeed, such filters can easily introduce or amplify biases into the data \cite{dodge2021documenting, welbl2021challenges}.
\end{itemize}

\textbf{Overview.} The Macrodata Refinement pipeline is split into three subsequent stages (see also \cref{fig:mdr_pipeline} for detailed removal rates): (1) document preparation; (2) filtering; (3) deduplication. 

For the document preparation, before undertaking any compute-heavy processing, we first filter documents based on the URL alone, using a blocklist of adult sites and scoring URLs based on their name. We found that the preprocessed .WET files offered by CommonCrawl still contain undesirable content (e.g., navigation menus) so we instead process raw .WARC files (HTML response) with \texttt{trafilatura} to extract natural text.  Finally, we use the fastText classifier from CCNet \citep{wenzek2020ccnet} to identify the top language of documents. For English, about 48\% of documents remain.

In the filtering stage, we apply a number of heuristics to remove repeated text (which may be an artefact of crawling/text extraction), and documents which are outliers in terms of length, symbol-to-word ratio, etc. These heuristics are inspired by \citet{gopher}. We also introduce so-called line-wise corrections, which remove lingering artefacts such as likes counter or navigation buttons. The size of the suitable data is against halved, resulting in about 23\% of CommonCrawl being kept.

Finally, we apply large-scale deduplication in two steps: first, we remove approximate duplicates at the document-level with MinHash \citep{broder1997resemblance}, before removing exact substring matches with a suffix array \citep{manber1993suffix}. With the deduplication settings of \citet{lee2022deduplicating}, this results in a final halving of the usable data, down to only about 12\% of the total data in CommonCrawl.

\subsubsection{The Microdata curated corpora and conversational masking}
\label{sec:micro_data}

In \cref{sec:curated_exp}, we found that adding curated data from conversations, books, or technical sources did not further improve performance on top of a strong web baseline like RefinedWeb. However, we believe this sort of data, along with code, can broaden the expressiveness of the model, and its applicability to different kinds of downstream tasks not captured by our evaluation setup. Accordingly, we also add a small fraction of curated data, with individual sources inspired from \citep{gao2020pile}. We also add data from Reddit \citep{baumgartner2020pushshift}, and introduce a new attention masking strategy for formatting tree-like conversations efficiently. On top of the Microdata curated corpora, we apply the Macrodata Refinement pipeline with adjusted filter settings (e.g., tuning document length thresholds for books) and deduplicate individual corpora.  

\textbf{Components from The Pile.} We reuse individual components of The Pile \citep{gao2020pile}, except where there are significant quality or licensing concerns. In all cases, we reimplement them from scratch, to ensure the formats match our data pipeline. We introduced a number of special tokens to reproduce structured information in curated corpora and better control generation: \texttt{>>TITLE<<}, \texttt{>>ABSTRACT<<}, \texttt{>>INTRODUCTION<<}, and \texttt{>>COMMENT<<}. When \LaTeX~files are available, we convert them to markdown, similar to \citet{lewkowycz2022solving}. Heuristics from Macrodata Refinement where hand-tuned for each corpora, manually analysing rejected and accepted samples; for books, we also introduce new rules to remove irrelevant content such as indexes, disclaimers, or tables of content.

\textbf{Conversational data.} Increasingly, large language models are deployed in "chatty" use cases, with back and forth interactions between users and models \citep{adiwardana2020towards, zheng2023judging}. Altough models are adapted downstream to this use case, we put a focus on enhancing pretraining with conversational data; we notably add data from Reddit \citep{baumgartner2020pushshift}.

\textbf{Conversation trees and attention masking.} One issue with using data from online forums such as Reddit or HackerNews is that this data is formatted in trees, with diverging turns of conversation between users. Past models have sampled trajectories from these trees \citep{thoppilan2022lamda, palm}, but this either means that data has to be repeated, or some trajectories left out. Instead, we find that we can use the attention mask to encode a tree-like structure, allowing later comments to only attend to comments from their trajectory, ignoring "side" comments. Conversations are serialized into sequences in depth-first order, and then comments/turns are masked if they are not relevant to the current one (i.e., in a different branch or deeper). For positionnal embeddings, we use the depth of the tree, which naturally serializes the conversation. We illustrate with pseudocode for converting depth first positions to an attention mask in \cref{sec:attn_mask_code}. We also employ this strategy to mask documents from one another, instead of relying on the \texttt{<EOD>} token alone. Note that cursory experiments did not find a benefit for zero-shot performance.

\begin{figure}[t]
    \centering
    \input{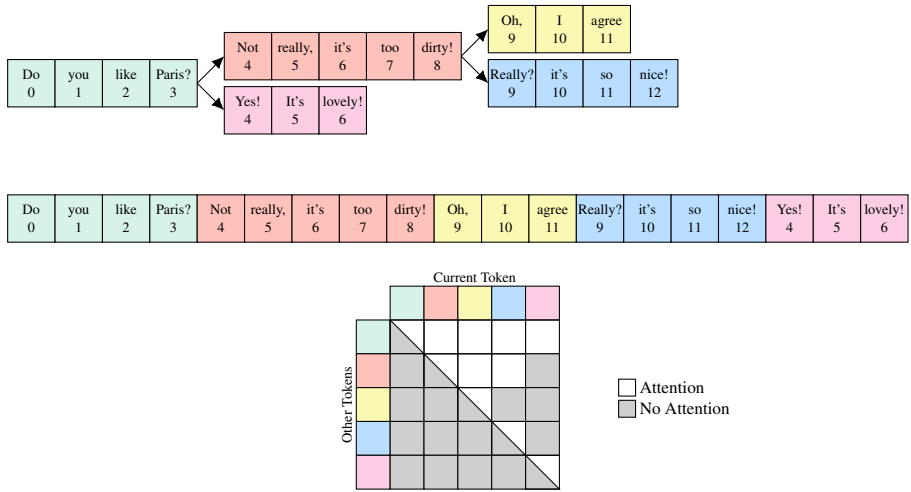}
    \caption{\textbf{\data{Tree-like attention masking enables us to realize all conversation trajectories without sampling or repeating data.}} The tree is serialized depth-first, where the position of the token is the depth in the tree. This allows us to train on all conversations in the tree at the same time, without repeating the early turns of the conversation. For instance, the passage \textit{Yes! It's lovely}, in \textcolor{adaptation}{pink}, has visibility on \textit{Do you like Paris?} in \textcolor{fine-tuning-darker}{green} and causal attention on itself, but cannot see other passages.}
    \label{fig:attn_mask}
    \vspace{-0.1in}
\end{figure}

\subsection{The Falcon architecture and recipe for efficient inference and (stable) training}

Our goals with the Falcon architecture are to maximize training and inference efficiency, while minimizing impact to downstream performance and risks for the models. In \cref{sec:arch_ablations}, we outlined a number of decisions we made based on the ablations:
\vspace{-0.1in}
\begin{itemize}
    \item \textbf{Architecture.} We use multigroup attention (\cref{sec:multiquery_abl}) to improve the scalability of inference, an extension of multiquery \citep{multiquery}; we use rotary embeddings \citep{su2021roformer}; we do not use GLU \citep{shazeer2020glu} because of the increased memory footprint, and use vanilla GeLU instead; we use parallel attention and MLP blocks \citep{gpt-j} and remove biases from linear layers \citep{palm}. 
    \item \textbf{Hyperparameters.} We use z-loss to help with stability \citep{meshtf}; we use a fixed 0.1 weight decay; we perform a learning rate logarithmic grid search during warm-up and pick the learning rate with the lowest loss at the end of warm-up.
\end{itemize}

\subsubsection{Architectural nitpicks: separate layer norms, tied embeddings, and scaling-up}

\textbf{Layer norms.} When using parallel attention, it is possible to either have separate layer norms for the MLP and the attention block (closer to the vanilla Transformer architecture), or to use a unified layer norm for both. Since the gradient computation with regards to the input of the layer norm is linear, it is possible to maintain the favourable communication volume of parallel attention and MLP while using two layer norms. Furthermore, after training, it is possible to merge the two layer norms back into one, by multiplying in the weights and biases of the layer norm into the subsequent linear layer. This leads us to stay close to the vanilla architecture, and to use two separate layer norms; however, we note this introduces unnecessary additional complexity for downstream conversion to other popular formats. For Falcon-7B (later trained), we switched to a single layer norm. 

\textbf{Tied embeddings.} Tying embeddings is a ubiquitous practice for Transformer models: the embedding weights converting the tokens $x$ into $z^0=xW$ are the same that converts the embedding back into the predicted logits $p=z^n W^T$. Although it is still used by most recent LLMs (GPT-3 \citet{brown2020gpt3}, PaLM \citet{palm}, LLaMA \citet{llama}), the original motivations \citep{press2016tied, inan2016tying} may not entirely be relevant. Notably, weight sharing was at the time used to reduce the size of models; but for models with over 100 billion parameters, embeddings are not a significant fraction of the parameters. Furthermore, weight sharing poses challenges in distributed training, as it requires additionnal communications. The semantic argument remains, but cursory experiments showed no strong impact at the 1B parameters scale. Still, in the interest of not adding additional risks to the training of the Falcon series, we keep our embeddings tied.

\textbf{Vocabulary size.} The size of the vocabulary in language models can differ widely from character level models with 256 entries \citep{xue2022byt5} to massively multilingual models with millions \citep{liang2023xlmvocab}. For generative models, practices have collapsed around two modes: vocabularies in the 30-60k range for monolingual models \citep{brown2020gpt3, gopher, llama}, or in the +100k range for models more inclined towards multilinguality \citep{bloom, palm}. Although larger vocabularies may have better fertility, and hence yield faster inference per byte of text, they also come with some caveats: from a scalability perspective, they can lead to unbalanced pipeline stages, and may require more storage space; also, it is unclear whether models optimally use them \citep{J1WhitePaper}. We train our tokenizer on a vocabulary size of 65,024, and store vocabulary information in a 16bit unsigned integer--this leaves about 500 extra values to use for downstream adaptations (e.g., paradigm tokens for UL2 \citet{tay2022transcending}).

\textbf{Model scaling.} We outline the shape and hyperparameters of the Falcon models in \cref{tab:hyperparams}. When increasing compute budget, resources can be either spent towards a larger model (increased parameter count) or towards longer training (increased token count). \citet{hoffmann2022training} recommends a joint (i.e., equal) increase for optimal scaling. However, this finding should be nuanced in two ways: (1) increasing parameter count also increases downstream inference costs, which can be significant if a model is widely deployed; (2) on the other hand, if data-constrained, increasing the size may be a way to trade compute for increased downstream performance. For Falcon, we choose to use \citet{hoffmann2022training} as a lower bound for pretraining length and as an upper bound for model size.  

Beyond model size, there is also the question of how to shape these parameters: e.g., should they be allocated to make a deeper or shallower model, to widen attention heads or to increase their count. \citet{what_lang_model} conducted a short review, highlighting that model depth is typically only scaled logarithmically with total parameter count \citep{levine2020limits}. Practices around attention head size however varied. We broadly follow the logarithmic depth scaling recommendation, and uses a fixed attention head size of 64 to optimize for performance with FlashAttention \citep{flash_attn}. We also fix our number of queries in multigroup to be equal to the number of tensor parallel degrees used during training, and train for a fixed sequence length of 2,048. We note that this context length can be efficiently increased with an a posteriori adaptation \citep{chen2023extending}.

\begin{table}[t]
\caption{\textbf{Summary of the shape, hyperparameters, and distribution strategy of the Falcon models.} Falcon-7B was trained after Falcon-40/180B, with an experimental increased batch size.}
\vspace{0.1in}
\centering
\label{tab:hyperparams}
\begin{tabular}{lx{2cm}x{2cm}x{2cm}}
\toprule
                     & \textbf{Falcon-7B} & \textbf{Falcon-40B}   & \textbf{Falcon-180B}  \\ \midrule
\textbf{Data}      & 1,500B             & 1,000B                & 3,500B                \\ \midrule
\textbf{Shape}       &                    &                       &                       \\
$n_\text{layer}$     & 32                 & 60                    & 80                    \\
$d_\text{model}$     & 4,544              & 8,192                 & 14,848                \\
$d_\text{head}$      & \multicolumn{3}{c}{64}                                             \\
$n_\text{q}$         & 71                 & 128                   & 232                   \\
$n_\text{kv}$        & 1                  & 8                     & 8                     \\
$d_\text{vocab}$     & \multicolumn{3}{c}{65,024}                                         \\
$n_\text{tokens}$        & \multicolumn{3}{c}{2,048}                                          \\\midrule
\textbf{Pretraining} &                    &                       &                       \\
Learning rate        & $6 \times 10^{-4}$ & $1.85 \times 10^{-4}$ & $1.25 \times 10^{-4}$ \\
\phantom{LR} Decay                & \multicolumn{3}{c}{Cosine, divides by 10}                                    \\
\phantom{LR} Ramp-up              & 4B                 & 4B                    & 4B                    \\
Batch-size           & 2,304              & 1,152                 & 2,048                 \\
\phantom{LR} Warm-up              & 30B                & 100B                  & 100B                  \\
Weight decay         & \multicolumn{3}{c}{0.1}                                            \\
Gradient clipping    & 1.                 & 0.6                   & 0.4                   \\
Z-loss               & \multicolumn{3}{c}{$1 \times 10^{-4}$}                             \\\midrule
\textbf{Parallelism} &                    &                       &                       \\
TP                   & 1                  & 8                     & 8                     \\
PP                   & 2                  & 4                     & 8                     \\ 
DP                   & 192                & 12                    & 64                   \\ \bottomrule
\end{tabular}
\vspace{-0.1in}
\end{table}

\subsubsection{Large language model alchemy: hyperparameters for pretraining}

We report hyperparameters used during pretraining in \cref{tab:hyperparams}, and highlight some decisions below.

\textbf{Learning rate search.} The procedure we describe in \cref{sec:hyperparam_abl} results in a learning rate of $6 \times 10^{-4}$, $1.85 \times 10^{-4}$, and $1.15 \times 10^{-4}$ for Falcon-7, 40, and 180B respectively. This is significantly higher than learning rates reported by previous models, but we found our training runs to be (mostly) stable. In hindsight we believe our search procedure may result in higher learning rates than optimal--an acceptable tradeoff, since recovering from spikes is relatively easy (\cref{spikes}).

\textbf{Learning rate ramp-up.} We perform a long ramp-up over 4 billion tokens for all models. 

\textbf{Batch size warm-up.} Practices around batch-size warm-up have surprisingly diverged widely depending on the underlying hardware: models trained on GPUs, such as GPT-3 \citep{brown2020gpt3}, BLOOM \citep{bloom}, or MT-NLG \citep{smith2022using}, have typically performed a fined-grained warm-up for 10-20 billion tokens; meanwhile, models trained on TPU, such as Gopher \citep{gopher}, Chinchilla \citep{hoffmann2022training}, or PaLM \citep{palm} often double the batch size mid-training or take larger batch size steps at 25/50\% through training. Some recent models have also elected to skip batch size warm-up entirely \citep{zhang2022opt, llama, llama2}. At small scale, we found longer warm-ups to never hurt downstream performance, and in fact to generally deliver better models; this leads us to adopt a long warm-up strategy, over 100 billion tokens--note that we are able to scale the data parallelism degree and overall size of the cluster during training, which means this comes at limited throughput cost. For Falcon-7B, wall-clock time constraints lead us to opt for an accelerated schedule over only 30 billion tokens.

\textbf{Gradient clipping.} We set the treshold for gradient clipping to $0.6$ for Falcon-40B. For Falcon-180B, we initially started with $0.6$ but later reduced it to $0.4$ to improve training stability. In both cases, we tuned the gradient clipping threshold to only affects outlier events and not the vast majority of the training steps. For Falcon-7B, we set it at 1. We find that training at this scale is anyway not really prone to instabilities, even with the large batch size we adopted.

\textbf{Optimizer.} We use AdamW \citep{adamW}: it is the most commonly used optimizer, and has been proven time and time again to perform well, both in general and for large language model training in particular. To increase performance during training we use the fused optimizer kernel from Megatron \cite{megatron-1}. However, we note that fused kernels for optimizers are less important in large scale training, especially when when optimizer sharding is used.

\textbf{No dropout.} As large language models are typically trained for a single epoch of relatively unique data, they typically do not use dropout \citep{dropout}. Shortage of data and recent papers suggesting a few epochs may be tolerable \citep{token_crisis} will surely challenge this practice as future model require tens of trillions of tokens, but for the Falcon series we did not use dropout.

\subsection{Large-scale distributed training on cloud infrastructure with Gigatron}

Rather than training on expensive dedicated HPC resources, we elected to train the Falcon series on cloud infrastructure to improve cost-efficiency. The Falcon series was trained on clusters of \texttt{p4d} on AWS--with up to 4,096 A100s for Falcon-180B. Key metrics for our training infrastructure include:
\begin{itemize}
    \item \textbf{Nodes with $8 \times \text{A100 40GB}$.} We found that configurations with 4x A100 40/80GB, popular in some datacenters with power or granularity constraints, resulted in lower throughputs because of the reduction in available degrees of tensor parallelism (4 instead of 8). However, we found the 40GB version of the A100 to offer increased availability and cost-efficiency.
    \item \textbf{50Gbps interconnect per GPU.} State-of-the-art infrastructure will come with 200Gbps interconnect per A100, powered by low-latency InfiniBand; however, these configurations can be prohibitively expensive. We found that for models up to size of Falcon-180B, bandwidth had only a limited impact on overall throughput (mostly linked to the size of the \texttt{all\_reduce} across data parallel degrees). Conversely, the higher latency of EFA on \texttt{p4d} was the main bottleneck for pipeline communications, especially for small scale models.
    \item \textbf{No distributed filesystem.} We stream data directly from S3, instead of relying on a dedicated filesystem. Distributed filesystems are expensive and difficult to maintain, and the small data I/O volumes incurred by large language model training do not justify their use.
\end{itemize}

Our infrastructure accordingly exists as an in-between between a true HPC system and a flexible cloud environment--notably, we found that the GPUs still had to share a single spine in the datacenter, as multispine configurations were unreliable.  Altough it is more cost-efficient, it also requires us to be mindful of its limitations. We found that the popular (and simple) recipe of training with fully sharded data parallelism \citep{zero-msft} did not scale well to this infrastructure. Instead, we required the finer control of 3D parallelism \citep{megatron-2} to achieve optimal performance. 

Because of limitations in open-source frameworks at the time, we elected to build our own proprietary distributed training framework. \textsc{\textbf{\hardware{Gigatron}}} is based on \texttt{pytorch}, and at its core implements a 3D distributed parallelism strategy \citep{megatron-1, megatron-2} combined with ZeRO optimizer sharding \citep{zero-msft} to reduce memory consumption and improve scalability.

\subsubsection{Combining 3D parallelism for fine-grained control, and ZeRO for scalability}

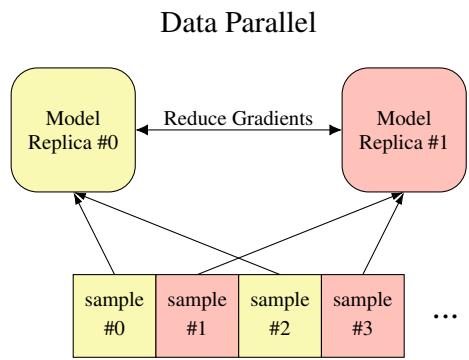
\begin{figure}[b]
    \centering
    \vspace{-0.2in}
    \begin{tikzpicture}[
dot/.style = {circle, fill, minimum size=#1,
                inner sep=0pt, outer sep=0pt},
               scale=0.55, every node/.style={scale=0.55}
]]
\node[above=0.4] at (5.5, 3.0) {\huge Data Parallel};
\draw[fill=architecture,rounded corners=10] (0.0, 0.0) rectangle (3.0, 3.0);
\node[above] at (1.5, 1.5) {\Large Model};
\node[below] at (1.5, 1.5) {\Large Replica \#0};
\draw[fill=objective,rounded corners=10] (8.0, 0.0) rectangle (11.0, 3.0);
\node[above] at (9.5, 1.5) {\Large Model};
\node[below] at (9.5, 1.5) {\Large Replica \#1};
\draw[{Latex[width=1.5mm]}-{Latex[width=1.5mm]},] (3.0, 1.5) to [bend left = 0] (8.0, 1.5);
\node[above] at (5.5, 1.5) {\Large Reduce Gradients};
\draw[fill=architecture,] (1.5, -4.0) rectangle (3.5, -2.0);
\node[above] at (2.5, -3.0) {\Large sample};
\node[below] at (2.5, -3.0) {\Large  \#0};
\draw[-{Latex[width=1mm]},] (2.5, -2.0) to [bend left = 0] (1.5, 0.0);
\draw[fill=objective,] (3.5, -4.0) rectangle (5.5, -2.0);
\node[above] at (4.5, -3.0) {\Large sample};
\node[below] at (4.5, -3.0) {\Large  \#1};
\draw[-{Latex[width=1mm]},] (4.5, -2.0) to [bend left = 0] (9.5, 0.0);
\draw[fill=architecture,] (5.5, -4.0) rectangle (7.5, -2.0);
\node[above] at (6.5, -3.0) {\Large sample};
\node[below] at (6.5, -3.0) {\Large  \#2};
\draw[-{Latex[width=1mm]},] (6.5, -2.0) to [bend left = 0] (1.5, 0.0);
\draw[fill=objective,] (7.5, -4.0) rectangle (9.5, -2.0);
\node[above] at (8.5, -3.0) {\Large sample};
\node[below] at (8.5, -3.0) {\Large  \#3};
\draw[-{Latex[width=1mm]},] (8.5, -2.0) to [bend left = 0] (9.5, 0.0);
\node[] at (10.5, -3.0) {\Huge ...};
\end{tikzpicture}
    \caption{\textbf{\hardware{Data parallelism creates model replicas on each device and process different samples in parallel.}} Model replicas are placed on different devices and compute gradients on different data samples in parallel. The gradients are then reduced before the optimization step is carried out.}
    \label{fig:dp}
\end{figure}

\textbf{Data Parallelism (DP).} Data parallelism (\cref{fig:dp}) is by far the most commonly used form of parallelism. All machine learning frameworks now provide ways to easily parallelize model training across data samples: \texttt{pytorch} with Distributed Data Parallel (\texttt{DDP}), \texttt{jax} with \texttt{pmap}, and \texttt{tensorflow} with \texttt{MirroredStrategy} \citep{pytorch, jax, tensorflow}. Data parallelism is appealing thanks to its simplicity: the distributed machinery can be hidden away inside the framework easily, without any interactions with the user-defined architecture. In its simplest implementation, data parallelism only requires two modifications to the training procedure: (1) each device needs to operate on unique data samples; (2) the gradients needs to be reduced across devices to keep the weights in sync. We note that this \texttt{all\_reduce} will grow with the batch size, eventually making data parallelism bandwidth-bound. Furthermore, data parallelism is however no cure-all: since the model is not sharded, memory footprint per device is constant, even as we add more devices. Accordingly, data parallelism alone is constrained to models which fit on a single device.

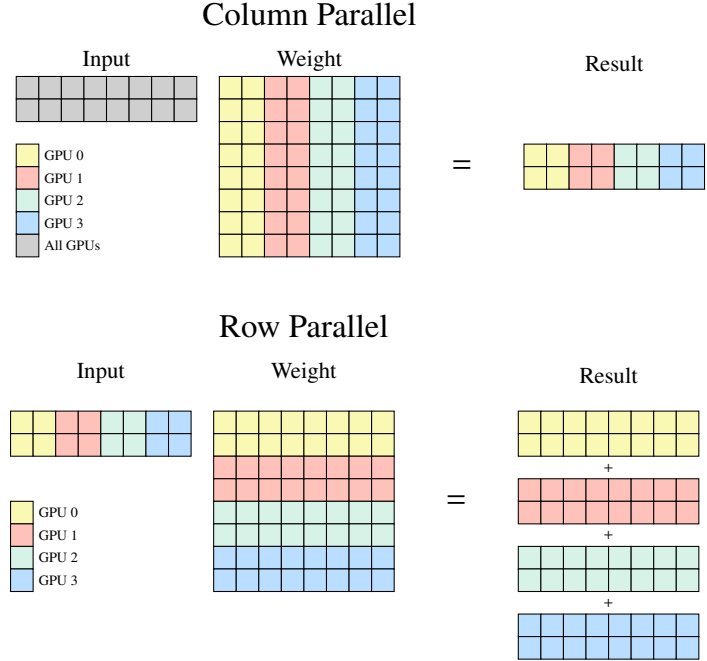
\begin{figure}
    \centering
    \begin{tikzpicture}[
dot/.style = {circle, fill, minimum size=#1,
                inner sep=0pt, outer sep=0pt},
               scale=0.6, every node/.style={scale=0.6}
]]
\node[above] at (2.0, 4.0) {\Large Input};
\node[above] at (6.5, 4.0) {\Large Weight};
\node[above] at (13.25, 4.0) {\Large Result};
\node[above] at (6.5, 5.0) {\huge Column Parallel};
\draw[fill=architecture,] (4.5, 3.5) rectangle (5.0, 4.0);
\draw[fill=architecture,] (4.5, 3.0) rectangle (5.0, 3.5);
\draw[fill=architecture,] (5.0, 3.5) rectangle (5.5, 4.0);
\draw[fill=architecture,] (5.0, 3.0) rectangle (5.5, 3.5);
\draw[fill=architecture,] (4.5, 2.5) rectangle (5.0, 3.0);
\draw[fill=architecture,] (4.5, 2.0) rectangle (5.0, 2.5);
\draw[fill=architecture,] (5.0, 2.5) rectangle (5.5, 3.0);
\draw[fill=architecture,] (5.0, 2.0) rectangle (5.5, 2.5);
\draw[fill=architecture,] (4.5, 1.5) rectangle (5.0, 2.0);
\draw[fill=architecture,] (4.5, 1.0) rectangle (5.0, 1.5);
\draw[fill=architecture,] (5.0, 1.5) rectangle (5.5, 2.0);
\draw[fill=architecture,] (5.0, 1.0) rectangle (5.5, 1.5);
\draw[fill=architecture,] (4.5, 0.5) rectangle (5.0, 1.0);
\draw[fill=architecture,] (4.5, 0.0) rectangle (5.0, 0.5);
\draw[fill=architecture,] (5.0, 0.5) rectangle (5.5, 1.0);
\draw[fill=architecture,] (5.0, 0.0) rectangle (5.5, 0.5);
\draw[fill=objective,] (5.5, 3.5) rectangle (6.0, 4.0);
\draw[fill=objective,] (5.5, 3.0) rectangle (6.0, 3.5);
\draw[fill=objective,] (6.0, 3.5) rectangle (6.5, 4.0);
\draw[fill=objective,] (6.0, 3.0) rectangle (6.5, 3.5);
\draw[fill=objective,] (5.5, 2.5) rectangle (6.0, 3.0);
\draw[fill=objective,] (5.5, 2.0) rectangle (6.0, 2.5);
\draw[fill=objective,] (6.0, 2.5) rectangle (6.5, 3.0);
\draw[fill=objective,] (6.0, 2.0) rectangle (6.5, 2.5);
\draw[fill=objective,] (5.5, 1.5) rectangle (6.0, 2.0);
\draw[fill=objective,] (5.5, 1.0) rectangle (6.0, 1.5);
\draw[fill=objective,] (6.0, 1.5) rectangle (6.5, 2.0);
\draw[fill=objective,] (6.0, 1.0) rectangle (6.5, 1.5);
\draw[fill=objective,] (5.5, 0.5) rectangle (6.0, 1.0);
\draw[fill=objective,] (5.5, 0.0) rectangle (6.0, 0.5);
\draw[fill=objective,] (6.0, 0.5) rectangle (6.5, 1.0);
\draw[fill=objective,] (6.0, 0.0) rectangle (6.5, 0.5);
\draw[fill=fine-tuning,] (6.5, 3.5) rectangle (7.0, 4.0);
\draw[fill=fine-tuning,] (6.5, 3.0) rectangle (7.0, 3.5);
\draw[fill=fine-tuning,] (7.0, 3.5) rectangle (7.5, 4.0);
\draw[fill=fine-tuning,] (7.0, 3.0) rectangle (7.5, 3.5);
\draw[fill=fine-tuning,] (6.5, 2.5) rectangle (7.0, 3.0);
\draw[fill=fine-tuning,] (6.5, 2.0) rectangle (7.0, 2.5);
\draw[fill=fine-tuning,] (7.0, 2.5) rectangle (7.5, 3.0);
\draw[fill=fine-tuning,] (7.0, 2.0) rectangle (7.5, 2.5);
\draw[fill=fine-tuning,] (6.5, 1.5) rectangle (7.0, 2.0);
\draw[fill=fine-tuning,] (6.5, 1.0) rectangle (7.0, 1.5);
\draw[fill=fine-tuning,] (7.0, 1.5) rectangle (7.5, 2.0);
\draw[fill=fine-tuning,] (7.0, 1.0) rectangle (7.5, 1.5);
\draw[fill=fine-tuning,] (6.5, 0.5) rectangle (7.0, 1.0);
\draw[fill=fine-tuning,] (6.5, 0.0) rectangle (7.0, 0.5);
\draw[fill=fine-tuning,] (7.0, 0.5) rectangle (7.5, 1.0);
\draw[fill=fine-tuning,] (7.0, 0.0) rectangle (7.5, 0.5);
\draw[fill=evaluation,] (7.5, 3.5) rectangle (8.0, 4.0);
\draw[fill=evaluation,] (7.5, 3.0) rectangle (8.0, 3.5);
\draw[fill=evaluation,] (8.0, 3.5) rectangle (8.5, 4.0);
\draw[fill=evaluation,] (8.0, 3.0) rectangle (8.5, 3.5);
\draw[fill=evaluation,] (7.5, 2.5) rectangle (8.0, 3.0);
\draw[fill=evaluation,] (7.5, 2.0) rectangle (8.0, 2.5);
\draw[fill=evaluation,] (8.0, 2.5) rectangle (8.5, 3.0);
\draw[fill=evaluation,] (8.0, 2.0) rectangle (8.5, 2.5);
\draw[fill=evaluation,] (7.5, 1.5) rectangle (8.0, 2.0);
\draw[fill=evaluation,] (7.5, 1.0) rectangle (8.0, 1.5);
\draw[fill=evaluation,] (8.0, 1.5) rectangle (8.5, 2.0);
\draw[fill=evaluation,] (8.0, 1.0) rectangle (8.5, 1.5);
\draw[fill=evaluation,] (7.5, 0.5) rectangle (8.0, 1.0);
\draw[fill=evaluation,] (7.5, 0.0) rectangle (8.0, 0.5);
\draw[fill=evaluation,] (8.0, 0.5) rectangle (8.5, 1.0);
\draw[fill=evaluation,] (8.0, 0.0) rectangle (8.5, 0.5);
\draw[fill=neutral,] (0.0, 3.5) rectangle (0.5, 4.0);
\draw[fill=neutral,] (0.0, 3.0) rectangle (0.5, 3.5);
\draw[fill=neutral,] (0.5, 3.5) rectangle (1.0, 4.0);
\draw[fill=neutral,] (0.5, 3.0) rectangle (1.0, 3.5);
\draw[fill=neutral,] (1.0, 3.5) rectangle (1.5, 4.0);
\draw[fill=neutral,] (1.0, 3.0) rectangle (1.5, 3.5);
\draw[fill=neutral,] (1.5, 3.5) rectangle (2.0, 4.0);
\draw[fill=neutral,] (1.5, 3.0) rectangle (2.0, 3.5);
\draw[fill=neutral,] (2.0, 3.5) rectangle (2.5, 4.0);
\draw[fill=neutral,] (2.0, 3.0) rectangle (2.5, 3.5);
\draw[fill=neutral,] (2.5, 3.5) rectangle (3.0, 4.0);
\draw[fill=neutral,] (2.5, 3.0) rectangle (3.0, 3.5);
\draw[fill=neutral,] (3.0, 3.5) rectangle (3.5, 4.0);
\draw[fill=neutral,] (3.0, 3.0) rectangle (3.5, 3.5);
\draw[fill=neutral,] (3.5, 3.5) rectangle (4.0, 4.0);
\draw[fill=neutral,] (3.5, 3.0) rectangle (4.0, 3.5);
\draw[fill=architecture,] (11.25, 2.0) rectangle (11.75, 2.5);
\draw[fill=architecture,] (11.25, 1.5) rectangle (11.75, 2.0);
\draw[fill=architecture,] (11.75, 2.0) rectangle (12.25, 2.5);
\draw[fill=architecture,] (11.75, 1.5) rectangle (12.25, 2.0);
\draw[fill=objective,] (12.25, 2.0) rectangle (12.75, 2.5);
\draw[fill=objective,] (12.25, 1.5) rectangle (12.75, 2.0);
\draw[fill=objective,] (12.75, 2.0) rectangle (13.25, 2.5);
\draw[fill=objective,] (12.75, 1.5) rectangle (13.25, 2.0);
\draw[fill=fine-tuning,] (13.25, 2.0) rectangle (13.75, 2.5);
\draw[fill=fine-tuning,] (13.25, 1.5) rectangle (13.75, 2.0);
\draw[fill=fine-tuning,] (13.75, 2.0) rectangle (14.25, 2.5);
\draw[fill=fine-tuning,] (13.75, 1.5) rectangle (14.25, 2.0);
\draw[fill=evaluation,] (14.25, 2.0) rectangle (14.75, 2.5);
\draw[fill=evaluation,] (14.25, 1.5) rectangle (14.75, 2.0);
\draw[fill=evaluation,] (14.75, 2.0) rectangle (15.25, 2.5);
\draw[fill=evaluation,] (14.75, 1.5) rectangle (15.25, 2.0);
\node[] at (9.875, 2.0) {\huge =};
\draw[fill=architecture] (0.0, 2.0) rectangle (0.5, 2.5);
\node[right] at (0.5, 2.25) {\small GPU 0};
\draw[fill=objective] (0.0, 1.5) rectangle (0.5, 2.0);
\node[right] at (0.5, 1.75) {\small GPU 1};
\draw[fill=fine-tuning] (0.0, 1.0) rectangle (0.5, 1.5);
\node[right] at (0.5, 1.25) {\small GPU 2};
\draw[fill=evaluation] (0.0, 0.5) rectangle (0.5, 1.0);
\node[right] at (0.5, 0.75) {\small GPU 3};
\draw[fill=neutral] (0.0, 0.0) rectangle (0.5, 0.5);
\node[right] at (0.5, 0.25) {\small All GPUs};
\node[] () [below = 1.5em]  {};
\end{tikzpicture}
    \begin{tikzpicture}[
dot/.style = {circle, fill, minimum size=#1,
                inner sep=0pt, outer sep=0pt},
               scale=0.6, every node/.style={scale=0.6}
]]
\node[above=0.3] at (2.0, 4.0) {\Large Input};
\node[above=0.3] at (6.5, 4.0) {\Large Weight};
\node[above=0.3] at (13.25, 4.0) {\Large Result};
\node[above=0.3] at (6.5, 5.0) {\huge Row Parallel};
\draw[fill=architecture,] (0.0, 3.5) rectangle (0.5, 4.0);
\draw[fill=architecture,] (0.0, 3.0) rectangle (0.5, 3.5);
\draw[fill=architecture,] (0.5, 3.5) rectangle (1.0, 4.0);
\draw[fill=architecture,] (0.5, 3.0) rectangle (1.0, 3.5);
\draw[fill=objective,] (1.0, 3.5) rectangle (1.5, 4.0);
\draw[fill=objective,] (1.0, 3.0) rectangle (1.5, 3.5);
\draw[fill=objective,] (1.5, 3.5) rectangle (2.0, 4.0);
\draw[fill=objective,] (1.5, 3.0) rectangle (2.0, 3.5);
\draw[fill=fine-tuning,] (2.0, 3.5) rectangle (2.5, 4.0);
\draw[fill=fine-tuning,] (2.0, 3.0) rectangle (2.5, 3.5);
\draw[fill=fine-tuning,] (2.5, 3.5) rectangle (3.0, 4.0);
\draw[fill=fine-tuning,] (2.5, 3.0) rectangle (3.0, 3.5);
\draw[fill=evaluation,] (3.0, 3.5) rectangle (3.5, 4.0);
\draw[fill=evaluation,] (3.0, 3.0) rectangle (3.5, 3.5);
\draw[fill=evaluation,] (3.5, 3.5) rectangle (4.0, 4.0);
\draw[fill=evaluation,] (3.5, 3.0) rectangle (4.0, 3.5);
\draw[fill=architecture,] (4.5, 3.5) rectangle (5.0, 4.0);
\draw[fill=architecture,] (4.5, 3.0) rectangle (5.0, 3.5);
\draw[fill=architecture,] (5.0, 3.5) rectangle (5.5, 4.0);
\draw[fill=architecture,] (5.0, 3.0) rectangle (5.5, 3.5);
\draw[fill=architecture,] (5.5, 3.5) rectangle (6.0, 4.0);
\draw[fill=architecture,] (5.5, 3.0) rectangle (6.0, 3.5);
\draw[fill=architecture,] (6.0, 3.5) rectangle (6.5, 4.0);
\draw[fill=architecture,] (6.0, 3.0) rectangle (6.5, 3.5);
\draw[fill=architecture,] (6.5, 3.5) rectangle (7.0, 4.0);
\draw[fill=architecture,] (6.5, 3.0) rectangle (7.0, 3.5);
\draw[fill=architecture,] (7.0, 3.5) rectangle (7.5, 4.0);
\draw[fill=architecture,] (7.0, 3.0) rectangle (7.5, 3.5);
\draw[fill=architecture,] (7.5, 3.5) rectangle (8.0, 4.0);
\draw[fill=architecture,] (7.5, 3.0) rectangle (8.0, 3.5);
\draw[fill=architecture,] (8.0, 3.5) rectangle (8.5, 4.0);
\draw[fill=architecture,] (8.0, 3.0) rectangle (8.5, 3.5);
\draw[fill=objective,] (4.5, 2.5) rectangle (5.0, 3.0);
\draw[fill=objective,] (4.5, 2.0) rectangle (5.0, 2.5);
\draw[fill=objective,] (5.0, 2.5) rectangle (5.5, 3.0);
\draw[fill=objective,] (5.0, 2.0) rectangle (5.5, 2.5);
\draw[fill=objective,] (5.5, 2.5) rectangle (6.0, 3.0);
\draw[fill=objective,] (5.5, 2.0) rectangle (6.0, 2.5);
\draw[fill=objective,] (6.0, 2.5) rectangle (6.5, 3.0);
\draw[fill=objective,] (6.0, 2.0) rectangle (6.5, 2.5);
\draw[fill=objective,] (6.5, 2.5) rectangle (7.0, 3.0);
\draw[fill=objective,] (6.5, 2.0) rectangle (7.0, 2.5);
\draw[fill=objective,] (7.0, 2.5) rectangle (7.5, 3.0);
\draw[fill=objective,] (7.0, 2.0) rectangle (7.5, 2.5);
\draw[fill=objective,] (7.5, 2.5) rectangle (8.0, 3.0);
\draw[fill=objective,] (7.5, 2.0) rectangle (8.0, 2.5);
\draw[fill=objective,] (8.0, 2.5) rectangle (8.5, 3.0);
\draw[fill=objective,] (8.0, 2.0) rectangle (8.5, 2.5);
\draw[fill=fine-tuning,] (4.5, 1.5) rectangle (5.0, 2.0);
\draw[fill=fine-tuning,] (4.5, 1.0) rectangle (5.0, 1.5);
\draw[fill=fine-tuning,] (5.0, 1.5) rectangle (5.5, 2.0);
\draw[fill=fine-tuning,] (5.0, 1.0) rectangle (5.5, 1.5);
\draw[fill=fine-tuning,] (5.5, 1.5) rectangle (6.0, 2.0);
\draw[fill=fine-tuning,] (5.5, 1.0) rectangle (6.0, 1.5);
\draw[fill=fine-tuning,] (6.0, 1.5) rectangle (6.5, 2.0);
\draw[fill=fine-tuning,] (6.0, 1.0) rectangle (6.5, 1.5);
\draw[fill=fine-tuning,] (6.5, 1.5) rectangle (7.0, 2.0);
\draw[fill=fine-tuning,] (6.5, 1.0) rectangle (7.0, 1.5);
\draw[fill=fine-tuning,] (7.0, 1.5) rectangle (7.5, 2.0);
\draw[fill=fine-tuning,] (7.0, 1.0) rectangle (7.5, 1.5);
\draw[fill=fine-tuning,] (7.5, 1.5) rectangle (8.0, 2.0);
\draw[fill=fine-tuning,] (7.5, 1.0) rectangle (8.0, 1.5);
\draw[fill=fine-tuning,] (8.0, 1.5) rectangle (8.5, 2.0);
\draw[fill=fine-tuning,] (8.0, 1.0) rectangle (8.5, 1.5);
\draw[fill=evaluation,] (4.5, 0.5) rectangle (5.0, 1.0);
\draw[fill=evaluation,] (4.5, 0.0) rectangle (5.0, 0.5);
\draw[fill=evaluation,] (5.0, 0.5) rectangle (5.5, 1.0);
\draw[fill=evaluation,] (5.0, 0.0) rectangle (5.5, 0.5);
\draw[fill=evaluation,] (5.5, 0.5) rectangle (6.0, 1.0);
\draw[fill=evaluation,] (5.5, 0.0) rectangle (6.0, 0.5);
\draw[fill=evaluation,] (6.0, 0.5) rectangle (6.5, 1.0);
\draw[fill=evaluation,] (6.0, 0.0) rectangle (6.5, 0.5);
\draw[fill=evaluation,] (6.5, 0.5) rectangle (7.0, 1.0);
\draw[fill=evaluation,] (6.5, 0.0) rectangle (7.0, 0.5);
\draw[fill=evaluation,] (7.0, 0.5) rectangle (7.5, 1.0);
\draw[fill=evaluation,] (7.0, 0.0) rectangle (7.5, 0.5);
\draw[fill=evaluation,] (7.5, 0.5) rectangle (8.0, 1.0);
\draw[fill=evaluation,] (7.5, 0.0) rectangle (8.0, 0.5);
\draw[fill=evaluation,] (8.0, 0.5) rectangle (8.5, 1.0);
\draw[fill=evaluation,] (8.0, 0.0) rectangle (8.5, 0.5);
\draw[fill=architecture,] (11.25, 3.5) rectangle (11.75, 4.0);
\draw[fill=architecture,] (11.25, 3.0) rectangle (11.75, 3.5);
\draw[fill=architecture,] (11.75, 3.5) rectangle (12.25, 4.0);
\draw[fill=architecture,] (11.75, 3.0) rectangle (12.25, 3.5);
\draw[fill=architecture,] (12.25, 3.5) rectangle (12.75, 4.0);
\draw[fill=architecture,] (12.25, 3.0) rectangle (12.75, 3.5);
\draw[fill=architecture,] (12.75, 3.5) rectangle (13.25, 4.0);
\draw[fill=architecture,] (12.75, 3.0) rectangle (13.25, 3.5);
\draw[fill=architecture,] (13.25, 3.5) rectangle (13.75, 4.0);
\draw[fill=architecture,] (13.25, 3.0) rectangle (13.75, 3.5);
\draw[fill=architecture,] (13.75, 3.5) rectangle (14.25, 4.0);
\draw[fill=architecture,] (13.75, 3.0) rectangle (14.25, 3.5);
\draw[fill=architecture,] (14.25, 3.5) rectangle (14.75, 4.0);
\draw[fill=architecture,] (14.25, 3.0) rectangle (14.75, 3.5);
\draw[fill=architecture,] (14.75, 3.5) rectangle (15.25, 4.0);
\draw[fill=architecture,] (14.75, 3.0) rectangle (15.25, 3.5);
\draw[fill=objective,] (11.25, 2.0) rectangle (11.75, 2.5);
\draw[fill=objective,] (11.25, 1.5) rectangle (11.75, 2.0);
\draw[fill=objective,] (11.75, 2.0) rectangle (12.25, 2.5);
\draw[fill=objective,] (11.75, 1.5) rectangle (12.25, 2.0);
\draw[fill=objective,] (12.25, 2.0) rectangle (12.75, 2.5);
\draw[fill=objective,] (12.25, 1.5) rectangle (12.75, 2.0);
\draw[fill=objective,] (12.75, 2.0) rectangle (13.25, 2.5);
\draw[fill=objective,] (12.75, 1.5) rectangle (13.25, 2.0);
\draw[fill=objective,] (13.25, 2.0) rectangle (13.75, 2.5);
\draw[fill=objective,] (13.25, 1.5) rectangle (13.75, 2.0);
\draw[fill=objective,] (13.75, 2.0) rectangle (14.25, 2.5);
\draw[fill=objective,] (13.75, 1.5) rectangle (14.25, 2.0);
\draw[fill=objective,] (14.25, 2.0) rectangle (14.75, 2.5);
\draw[fill=objective,] (14.25, 1.5) rectangle (14.75, 2.0);
\draw[fill=objective,] (14.75, 2.0) rectangle (15.25, 2.5);
\draw[fill=objective,] (14.75, 1.5) rectangle (15.25, 2.0);
\draw[fill=fine-tuning,] (11.25, 0.5) rectangle (11.75, 1.0);
\draw[fill=fine-tuning,] (11.25, 0.0) rectangle (11.75, 0.5);
\draw[fill=fine-tuning,] (11.75, 0.5) rectangle (12.25, 1.0);
\draw[fill=fine-tuning,] (11.75, 0.0) rectangle (12.25, 0.5);
\draw[fill=fine-tuning,] (12.25, 0.5) rectangle (12.75, 1.0);
\draw[fill=fine-tuning,] (12.25, 0.0) rectangle (12.75, 0.5);
\draw[fill=fine-tuning,] (12.75, 0.5) rectangle (13.25, 1.0);
\draw[fill=fine-tuning,] (12.75, 0.0) rectangle (13.25, 0.5);
\draw[fill=fine-tuning,] (13.25, 0.5) rectangle (13.75, 1.0);
\draw[fill=fine-tuning,] (13.25, 0.0) rectangle (13.75, 0.5);
\draw[fill=fine-tuning,] (13.75, 0.5) rectangle (14.25, 1.0);
\draw[fill=fine-tuning,] (13.75, 0.0) rectangle (14.25, 0.5);
\draw[fill=fine-tuning,] (14.25, 0.5) rectangle (14.75, 1.0);
\draw[fill=fine-tuning,] (14.25, 0.0) rectangle (14.75, 0.5);
\draw[fill=fine-tuning,] (14.75, 0.5) rectangle (15.25, 1.0);
\draw[fill=fine-tuning,] (14.75, 0.0) rectangle (15.25, 0.5);
\draw[fill=evaluation,] (11.25, -1.0) rectangle (11.75, -0.5);
\draw[fill=evaluation,] (11.25, -1.5) rectangle (11.75, -1.0);
\draw[fill=evaluation,] (11.75, -1.0) rectangle (12.25, -0.5);
\draw[fill=evaluation,] (11.75, -1.5) rectangle (12.25, -1.0);
\draw[fill=evaluation,] (12.25, -1.0) rectangle (12.75, -0.5);
\draw[fill=evaluation,] (12.25, -1.5) rectangle (12.75, -1.0);
\draw[fill=evaluation,] (12.75, -1.0) rectangle (13.25, -0.5);
\draw[fill=evaluation,] (12.75, -1.5) rectangle (13.25, -1.0);
\draw[fill=evaluation,] (13.25, -1.0) rectangle (13.75, -0.5);
\draw[fill=evaluation,] (13.25, -1.5) rectangle (13.75, -1.0);
\draw[fill=evaluation,] (13.75, -1.0) rectangle (14.25, -0.5);
\draw[fill=evaluation,] (13.75, -1.5) rectangle (14.25, -1.0);
\draw[fill=evaluation,] (14.25, -1.0) rectangle (14.75, -0.5);
\draw[fill=evaluation,] (14.25, -1.5) rectangle (14.75, -1.0);
\draw[fill=evaluation,] (14.75, -1.0) rectangle (15.25, -0.5);
\draw[fill=evaluation,] (14.75, -1.5) rectangle (15.25, -1.0);
\node[] at (13.25, 2.75) {+};
\node[] at (13.25, 1.25) {+};
\node[] at (13.25, -0.25) {+};
\node[] at (9.875, 2.0) {\huge =};
\draw[fill=architecture] (0.0, 1.5) rectangle (0.5, 2.0);
\node[right] at (0.5, 1.75) {\small GPU 0};
\draw[fill=objective] (0.0, 1.0) rectangle (0.5, 1.5);
\node[right] at (0.5, 1.25) {\small GPU 1};
\draw[fill=fine-tuning] (0.0, 0.5) rectangle (0.5, 1.0);
\node[right] at (0.5, 0.75) {\small GPU 2};
\draw[fill=evaluation] (0.0, 0.0) rectangle (0.5, 0.5);
\node[right] at (0.5, 0.25) {\small GPU 3};

\end{tikzpicture}
    \caption{\textbf{\hardware{By alternating column and row parallelism, no communications are required between two subsequent matrix multiplications, enabling tensor parallelism to efficiently split attention and MLP blocks across GPUs.}} For column parallel matrix multiplication, the input is replicated (\texttt{all\_reduce} in the backward) across all GPUs, which can then independently perform their operation. The result is already split across GPUs to execute the next matrix multiplication in a row parallel fashion. Finally, the output of each GPU is summed through an \texttt{all\_reduce}. Intermediary results between the two matrix multiplications are never communicated.}
    \label{fig:tp}
\end{figure}

\textbf{Tensor Parallelism (TP).} To share the model across devices, we need to turn to model parallelism. Introduced by \citet{megatron-1} in its most popular form, tensor parallelism splits linear layers in the attention block and MLP in a principled way to reduce communication volume. This can be viewed as an instance of width-wise model parallelism. Specifically, for the simple case of a two-layer MLP, by making the first layer column parallel and the second row parallel, only a single all reduce is necessary to produce the final result--no communication of the intermediary result is required. To propagate the gradient backward, the matrices are transposed, and hence the row parallel layer turn into a column parallel layer and vice-versa: this maintains the the efficient communication pattern. We illustrate column and row parallelism in \cref{fig:tp}. A similar approach can be taken to split attention blocks, splitting the heads across GPUs--wherein the K,Q,V calculations are column parallel and the final projection row parallel, with all computations in between independent of one another between GPUs. Accordingly, entire blocks in Transformers can be efficiently parallelized this way. Tensor parallel, however, requires both high-bandwith and low-latency interconnect to be effective: accordingly, on current GPU infrastructure, it is constrained within a single node, and cannot efficiently be used across nodes. Models will thus typically be trained with a tensor parallel degree up to 8; this may still be insufficient to create small enough shards of the model.

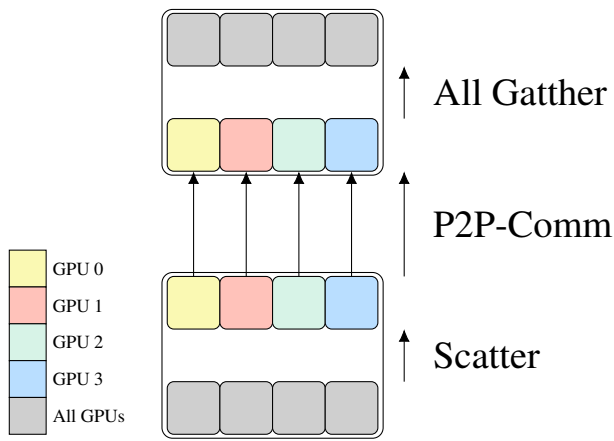
\begin{figure}
    \centering
    \begin{tikzpicture}[
dot/.style = {circle, fill, minimum size=#1,
                inner sep=0pt, outer sep=0pt},
               scale=0.7, every node/.style={scale=0.7}
]]
\draw[fill=neutral,rounded corners=2.5] (1.0, 0.0) rectangle (2.0, 1.0);
\draw[fill=architecture,rounded corners=2.5] (1.0, 2.0) rectangle (2.0, 3.0);
\draw[fill=architecture,rounded corners=2.5] (1.0, 5.0) rectangle (2.0, 6.0);
\draw[fill=neutral,rounded corners=2.5] (1.0, 7.0) rectangle (2.0, 8.0);
\draw[-{Latex[width=1.5mm]},] (1.5, 3.0) to [bend left = 0] (1.5, 5.0);
\draw[fill=neutral,rounded corners=2.5] (2.0, 0.0) rectangle (3.0, 1.0);
\draw[fill=objective,rounded corners=2.5] (2.0, 2.0) rectangle (3.0, 3.0);
\draw[fill=objective,rounded corners=2.5] (2.0, 5.0) rectangle (3.0, 6.0);
\draw[fill=neutral,rounded corners=2.5] (2.0, 7.0) rectangle (3.0, 8.0);
\draw[-{Latex[width=1.5mm]},] (2.5, 3.0) to [bend left = 0] (2.5, 5.0);
\draw[fill=neutral,rounded corners=2.5] (3.0, 0.0) rectangle (4.0, 1.0);
\draw[fill=fine-tuning,rounded corners=2.5] (3.0, 2.0) rectangle (4.0, 3.0);
\draw[fill=fine-tuning,rounded corners=2.5] (3.0, 5.0) rectangle (4.0, 6.0);
\draw[fill=neutral,rounded corners=2.5] (3.0, 7.0) rectangle (4.0, 8.0);
\draw[-{Latex[width=1.5mm]},] (3.5, 3.0) to [bend left = 0] (3.5, 5.0);
\draw[fill=neutral,rounded corners=2.5] (4.0, 0.0) rectangle (5.0, 1.0);
\draw[fill=evaluation,rounded corners=2.5] (4.0, 2.0) rectangle (5.0, 3.0);
\draw[fill=evaluation,rounded corners=2.5] (4.0, 5.0) rectangle (5.0, 6.0);
\draw[fill=neutral,rounded corners=2.5] (4.0, 7.0) rectangle (5.0, 8.0);
\draw[-{Latex[width=1.5mm]},] (4.5, 3.0) to [bend left = 0] (4.5, 5.0);
\draw[-{Latex[width=1.5mm]},] (5.5, 1.0) to [bend left = 0] (5.5, 2.0);
\draw[-{Latex[width=1.5mm]},] (5.5, 3.0) to [bend left = 0] (5.5, 5.0);
\draw[-{Latex[width=1.5mm]},] (5.5, 6.0) to [bend left = 0] (5.5, 7.0);
\node[right=0.3] at (5.5, 1.5) {\huge Scatter};
\node[right=0.3] at (5.5, 4.0) {\huge P2P-Comm};
\node[right=0.3] at (5.5, 6.5) {\huge All Gatther};
\draw[rounded corners] (0.8999999999999999, -0.07500000000000018) rectangle (5.1, 3.075);
\draw[rounded corners] (0.8999999999999999, 4.925) rectangle (5.1, 8.075);
\draw[fill=architecture] (-2.0, 2.8) rectangle (-1.2999999999999998, 3.5);
\node[right] at (-1.2999999999999998, 3.15) {GPU 0};
\draw[fill=objective] (-2.0, 2.0999999999999996) rectangle (-1.2999999999999998, 2.8);
\node[right] at (-1.2999999999999998, 2.4499999999999997) {GPU 1};
\draw[fill=fine-tuning] (-2.0, 1.4) rectangle (-1.2999999999999998, 2.0999999999999996);
\node[right] at (-1.2999999999999998, 1.7499999999999998) {GPU 2};
\draw[fill=evaluation] (-2.0, 0.7) rectangle (-1.2999999999999998, 1.4);
\node[right] at (-1.2999999999999998, 1.0499999999999998) {GPU 3};
\draw[fill=neutral] (-2.0, 0.0) rectangle (-1.2999999999999998, 0.7);
\node[right] at (-1.2999999999999998, 0.35) {All GPUs};

\end{tikzpicture}
    \caption{\textbf{\hardware{Instead of each rank redundantly sending the full tensor, scatter/gather optimization has each rank sends a shard of the tensor, and then leverages the fast intranode interconnect to \texttt{gather} it back.}} This patterns is more efficient as internode communications are typically much slower than intranode ones; note that the scatter is effectively a no-op, as each rank already has the full tensor. Internode rail optimization typically enables GPUs of the same rank in different nodes to communicate peer-to-peer with minimal intermediaries, further accelerating this pattern.}
    \label{fig:scatter_gather}
\end{figure}

\textbf{Pipeline parallelism.} Model parallelism can also be performed depth-wise, by grouping layers into subsequent stages to be executed on different accelerators. However, naive pipeline parallelism would either be inefficient or result in an immediate roadblock: indeed, stages have to split batches to concurrently process multiple forwards and backwards. We adopt the PipeDream-Flush schedule of \citet{pipedream}, now commonly referred to as 1F1B. We found more involved schedules, such as Interleaved-1F1B \citep{megatron-2}, to be beneficial when the number of microbatches per model replica is small, typically below 64. After the batch size rampup is completed, we train with a larger number of microbatches; as such we do not observe any speedup from interleaving. In order to reduce the communication volumes, we consistently make use of so called \texttt{scatter-gather} optimizations, where the activations sent to the next stage are first sharded over the tensor parallel degree and then gathered again once received on the next stage (see \cref{fig:scatter_gather} for an illustration). This optimizes for underlying interconnect topology (intranode communications over NVLink where the \texttt{gather}is executed are significantly faster), even more so if internode links have been rail optimized. This optimization reduces communication volume for pipeline parallelism by a factor of the tensor parallel world size, which in our case is 8. 

\textbf{Optimizer sharding.} Model weights and gradients are materialized in a contiguous \texttt{bfloat16} memory buffer. Similar to \citep{gopher}, the optimizer maintains a \texttt{fp32} version of those weights/gradients with which to calculate updates. When using the Adam optimizer in this setting, as we also need to keep track of exponential averages, we require 20 bytes per model parameters:

\begin{equation}
\underbrace{2_{\text{model param}} + 2_{\text{model grad}}}_{\texttt{bfloat16}} + \underbrace{4_{\text{opt param}} + 4_{\text{opt grad}} + 4_{\text{exp avgs}} + 4_{\text{exp avg sqs}}}_{\texttt{float32}} = 20\; \texttt{bytes}/\texttt{param}
\label{mem_usage}
\end{equation}

A full state of Falcon-7/40/180B will occupy 140GB, 800GB, and 3,600GB of memory per replica, requiring at least 4, 20, or 90 A100 40GB per replica just in weights+gradients+states of memory.

\begin{figure}
    \centering
    \begin{tikzpicture}[
dot/.style = {circle, fill, minimum size=#1,
                inner sep=0pt, outer sep=0pt},
               scale=0.6, every node/.style={scale=0.6}
]]
\draw[fill=architecture,draw=none] (6.0, 9.6) rectangle (10.0, 10.0);
\draw[fill=objective,draw=none] (6.0, 9.2) rectangle (10.0, 9.6);
\draw[fill=fine-tuning-darker,draw=none] (6.0, 8.8) rectangle (10.0, 9.2);
\draw[fill=fine-tuning-darker,draw=none] (6.0, 8.4) rectangle (10.0, 8.8);
\draw[fill=fine-tuning,draw=none] (6.0, 8.0) rectangle (10.0, 8.4);
\draw[fill=fine-tuning,draw=none] (6.0, 7.6) rectangle (10.0, 8.0);
\draw[fill=fine-tuning-darker,draw=none] (6.0, 7.2) rectangle (10.0, 7.6);
\draw[fill=fine-tuning-darker,draw=none] (6.0, 6.8) rectangle (10.0, 7.2);
\draw[fill=fine-tuning,draw=none] (6.0, 6.4) rectangle (10.0, 6.8);
\draw[fill=fine-tuning,draw=none] (6.0, 6.0) rectangle (10.0, 6.4);
\node[left] at (6.0, 9.8) {parameters \texttt{bf16}};
\node[left] at (6.0, 9.399999999999999) {gradients \texttt{bf16}};
\node[left] at (6.0, 7.2) {exp avg gradients \texttt{fp32}};
\node[left] at (6.0, 6.4) {exp avg gradients$^2$ \texttt{fp32}};
\node[left] at (6.0, 8.8) {parameters \texttt{fp32}};
\node[left] at (6.0, 8.0) {gradients \texttt{fp32}};
\draw[fill=architecture,draw=none] (0.0, 3.6) rectangle (4.0, 4.0);
\draw[fill=objective,draw=none] (0.0, 3.2) rectangle (4.0, 3.6);
\draw[fill=fine-tuning,draw=none] (0.0, 2.8000000000000003) rectangle (0.4, 3.2);
\draw[fill=fine-tuning,draw=none] (0.0, 2.4000000000000004) rectangle (0.4, 2.8000000000000003);
\draw[fill=fine-tuning,draw=none] (0.0, 2.0) rectangle (0.4, 2.4000000000000004);
\draw[fill=fine-tuning,draw=none] (0.0, 1.6) rectangle (0.4, 2.0);
\draw[fill=fine-tuning,draw=none] (0.0, 1.2000000000000002) rectangle (0.4, 1.6);
\draw[fill=fine-tuning,draw=none] (0.0, 0.8) rectangle (0.4, 1.2000000000000002);
\draw[fill=fine-tuning,draw=none] (0.0, 0.4) rectangle (0.4, 0.8);
\draw[fill=fine-tuning,draw=none] (0.0, 0.0) rectangle (0.4, 0.4);
\draw[fill=architecture,draw=none] (6.0, 3.6) rectangle (10.0, 4.0);
\draw[fill=objective,draw=none] (6.0, 3.2) rectangle (10.0, 3.6);
\draw[fill=fine-tuning,draw=none] (7.6, 2.8000000000000003) rectangle (8.0, 3.2);
\draw[fill=fine-tuning,draw=none] (7.6, 2.4000000000000004) rectangle (8.0, 2.8000000000000003);
\draw[fill=fine-tuning,draw=none] (7.6, 2.0) rectangle (8.0, 2.4000000000000004);
\draw[fill=fine-tuning,draw=none] (7.6, 1.6) rectangle (8.0, 2.0);
\draw[fill=fine-tuning,draw=none] (7.6, 1.2000000000000002) rectangle (8.0, 1.6);
\draw[fill=fine-tuning,draw=none] (7.6, 0.8) rectangle (8.0, 1.2000000000000002);
\draw[fill=fine-tuning,draw=none] (7.6, 0.4) rectangle (8.0, 0.8);
\draw[fill=fine-tuning,draw=none] (7.6, 0.0) rectangle (8.0, 0.4);
\draw[fill=architecture,draw=none] (12.0, 3.6) rectangle (16.0, 4.0);
\draw[fill=objective,draw=none] (12.0, 3.2) rectangle (16.0, 3.6);
\draw[fill=fine-tuning,draw=none] (15.6, 2.8000000000000003) rectangle (16.0, 3.2);
\draw[fill=fine-tuning,draw=none] (15.6, 2.4000000000000004) rectangle (16.0, 2.8000000000000003);
\draw[fill=fine-tuning,draw=none] (15.6, 2.0) rectangle (16.0, 2.4000000000000004);
\draw[fill=fine-tuning,draw=none] (15.6, 1.6) rectangle (16.0, 2.0);
\draw[fill=fine-tuning,draw=none] (15.6, 1.2000000000000002) rectangle (16.0, 1.6);
\draw[fill=fine-tuning,draw=none] (15.6, 0.8) rectangle (16.0, 1.2000000000000002);
\draw[fill=fine-tuning,draw=none] (15.6, 0.4) rectangle (16.0, 0.8);
\draw[fill=fine-tuning,draw=none] (15.6, 0.0) rectangle (16.0, 0.4);
\node[] at (5.0, 2.0) {\huge ...};
\node[] at (11.0, 2.0) {\huge ...};
\node[above=0.3] at (2.0, 4.0) {\large $\textbf{GPU}_{0}$};
\node[above=0.3] at (8.0, 4.0) {\large $\textbf{GPU}_{i}$};
\node[above=0.3] at (14.0, 4.0) {\large $\textbf{GPU}_{n-1}$};
\node[above=0.3] at (8.0, 10.0) {\large $\textbf{GPU}_{i}$};
\node[left=0.5] at (0.0, 2.0) {\large Sharded Optim State};
\node[left=0.5] at (0.0, 8.0) {\large Full Optim State};

\end{tikzpicture}
    \caption{\textbf{\hardware{Optimizer sharding splits the large optimizer state across data parallel degrees, reducing memory footprint and improving scalability.}} The freed memory can be traded for an increased microbatch size, improving throughput. Figure inspired from \citet{zero-msft}.}
    \label{fig:zero}
\end{figure}
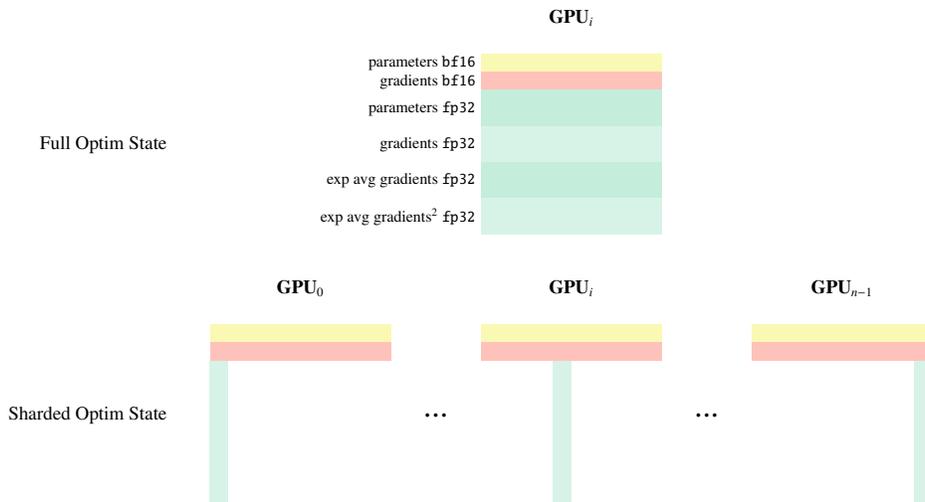

The optimizer state alone accounts for 80\% of the memory footprint, and eventually bottlenecks our memory use, preventing us from, for instance, increasing the batch size to better saturate GPUs instead. Accordingly, we choose to shard the optimizer state across data parallel degrees--a practice commonly referred to as ZeRO-1 in the literature \citep{zero-msft}, and illustrated in \cref{fig:zero}. Rather than storing a full redundant copy of the optimizer on each data parallel degree, the optimizer is independently sharded DP times. With optimizer state sharding the number of bytes per parameter is further reduced with increased data parallelism, improving scalability:

\begin{equation}
\text{Bytes per parameter} = 4 + \frac{16}{\texttt{DP}}
\label{zero_mem_usage}
\end{equation}

For Falcon-7/40/180B, we now only require 30, 215, and 765GB of memory, or 1, 6, 20 A100 40GB to hold a model replica. Note that these do not account for activation memory. Now that the optimizer is split, we however have to face increased communication burden during the backward. First, a \texttt{reduce\_scatter} is applied on the \texttt{bfloat16} model gradients: for each optimizer shard, this derives the relevant gradient shard reduced across all DP degrees. The resulting synchronized gradient chunk is then copied to the \texttt{fp32} optimizer gradient buffer for use by the optimizer. The optimizer step is performed, resulting in an updated copy of the relevant \texttt{fp32} sharded optimizer weights. Finally, an \texttt{all\_gather} is used to consolidate the \texttt{bfloat16} model weights for all workers from the aforementioned \texttt{fp32} sharded optimizer weights. Pseudo-code for this \texttt{ZeRO-1} workflow is provided in \cref{zero_pseudocode}. Interestingly, in terms of total communication volume, this workflow is \textbf{equivalent} to the \texttt{all\_reduce} usually used in the data parallel setting \cite{zero-msft}.

We found that across all scales (from 1 billion parameters to 180 billion parameters), optimizer sharding had no performance overhead compared to traditional data parallelism. In fact, since optimizer sharding frees-up memory, it enables us to use a larger microbatch size, hence resulting in better ressource saturation and higher throughput systematically. 

\textbf{No sequence parallelism.} \cite{seq_parallel} proposed a novel strategy to reduce the activation memory during training: it's possible to shard the activations on the residual stream across the tensor parallel workers where these activations are otherwise replicated. This sharding is relatively cheap: instead of using an \texttt{all\_reduce} after the MLP one can replace it with a \texttt{reduce\_scatter} followed by an \texttt{all\_gather} right before the next decoder block. However, after implementing the measures to save memory discussed in \ref{selective_recomputation}, we observe that sequence parallelism is not necessary. We also saw a slight decrease in throughput which made us decide against employing it during the final training.

\subsubsection{State-of-the-art throughput with dedicated Triton kernels}
\label{other_kernels}

\textbf{Flash Attention.} Memory efficient attention alternatives have long garnered the attention of the community--however, these came with significant changes to the attention scheme, often resulting in degraded downstream performance, especially when scaling-up. It's only (relatively) recently that exact algorithms have been developed to compute attention without requiring materialization of the full attention matrix in memory \citep{flash_attn_prev}. \citet{flash_attn} subsequently showed that unified kernels leveraging the same techniques can significantly speed-up model training. We use a custom implementation in Triton \citep{triton}, which incorporates some of the improvements concurrently developed for \citet{flash_attn_2}. We note that the use of FlashAttention is the main driver of improved throughput during training. Notably, the memory savings allow us to not rely on activation checkpointing during training, instead focusing our FLOPS on contributing to model training directly. Dedicated FlashAttention kernels are also faster in absolute terms than their naive counterparts. Interestingly, we note that FlashAttention is significantly more numerically accurate than traditional attention when both are performed in \texttt{bfloat16} and compared to \texttt{fp32} as the ground truth.

\textbf{Other Triton kernels.} We also employ specialised kernels for the rotary positional embeddings as well as the layer norms in Triton. For rotary embeddings we see a particularly large improvement compared to their unfused PyTorch counterpart. We note that since the transformation is the same for all heads at the same position it's possible to reuse the expensive trigonometric functions, additionally simplifying its use as nothing needs to be cached in RAM, as is common in other implementations.

\subsubsection{Efficient memory use via selective recomputation implemented as a monolayer}
\label{selective_recomputation}

As we target A100 40GB for cost-efficiency, memory footprint is a significant concern for us. Both ZeRO and FlashAttention already significantly improve memory availability for us, but we free up additionnal memory via selective recomputation. \citet{seq_parallel}, on top of sequence parallelism, proposed to recompute some activations rather than storing their output in memory for backpropagation--as activations are typically cheap to evaluate. We take this idea a step further, and recompute not only all the activation functions but also the layer norms. We save to memory the statistics of the layer norm only, which makes recomputation trivial. Overall, this implementation of selective recomputation reduces the memory consumption of the decoder block by a factor 2x, while causing no degradation in throughput from the additional computations.  

Due to limitations in the \texttt{autograd} engine of PyTorch, it's rather difficult to only recompute e.g. the layer norms without also recomputing the subsequent linear layer. In order to achieve this, we create a single custom \texttt{autograd} function for the entire decoder block--we dub this idea the \textit{monolayer}.

\subsubsection{Numerical precision: all you need is \texttt{bfloat16}?}

\citet{brown2020gpt3, zhang2022opt, bloom} all reported stability issues training models in the hundred billion parameters range when relying on \texttt{fp16}; we instead adopt \texttt{bfloat16}, which results in stable training more or less out-of-the-box. \citet{gopher} reported significant improvements from using stochastic rounding when quantizing the \texttt{float32} parameters to \texttt{bfloat16} after the optimization step. In our case however, where the entire optimizer state is stored in \texttt{float32}, we do not observe a similar improvement. Instead, we observe that stochastic rounding helps during the initial steps, before the optimizer state has equilibriated to the training dynamics, but that training without stochastic rounding approaches the same training trajectory shortly thereafter.

\subsubsection{Quality-of-life features for improved flexibility and reliability}

\textbf{Topology-agnostic checkpoints.} Surprisingly, when we developed the Falcon series, most distributed training libraries strongly tied their checkpoint format and distribution topology: a change to the topology would require manually converting the checkpoint to a new format. Since we were planning to grow our clusters during training, we ensure that the model and optimizer checkpoints are readable/writable between any topology configurations--similar to \texttt{t5x} \citep{t5x}. 

\textbf{Low-discrepancy data loading.} When aggregating different sources into a single mixed dataset it is common to randomly sample each source according to a target probability/weighing per dataset. However, this only guarantees sampling from the sources with the target probability in expectation. Instead, it is desirable that the effective weight during each subset of the training is constant. We use a predefined sampling pattern between different data-sources with a relative short length of $10,000$ sequences which is guaranteed to contain the exact weights for each data source. 

\textbf{Topology discovery and optimization.}  In order to optimize throughput, it's important to place ranks with high communication volume on instances close to one another. At variance with traditional HPC platforms, AWS does not expose topology information for the instances allocated to a training. Accordingly, to optimize placement, we first discover the topology by measuring the bandwidth between all pairs of nodes, and we subsequently optimize rank placement against the measured topology. Regarding the topology discovery phase, a naive solution would be for the first node to first check the bandwidth to all other nodes, then the second node and so on. That would require $\mathcal{O}(n^2)$ sequential measurements. Instead, it is possible to achieve this in only $\mathcal{O}(n)$ steps, by doing many comparisons in parallel--see pseudocode provided in \ref{all_to_all_code}. Once the topology has been (efficiently) discovered, we leverage Gromov-Wasserstein optimal transport \citep{gromov-wasserstein} for placement. We measure the source distance matrix based on the measured bandwidths, and define the target distance matrix from the distances in the pipeline/data parallel grid.  We then use the Python Optimal Transport Toolbox \citep{POT} to solve the resulting problem efficiently. 

\subsection{Run management: keeping large-scale infrastructure running}
\label{spikes}

\textbf{Hardware failures.} When scaling-up to hundreds of nodes, hardware failures become increasingly common--for Falcon-180B, every day on 4,096 A100 is equivalent to over 11 years of cumulative use! Furthermore, in our cloud setup, whenever the run is restarted, we sample a new anonymised selection of nodes--we are unable to keep a history of which nodes from previous allocations were suspect. Accordingly, it is critical to be able to rapidly identify and exclude faulty nodes. We found a large majority of hardware failures to be linked to faulty A100s, specifically to corrupted memory rows. These failures do not always manifest with an \texttt{Xid} code, and require manual tests to catch them--they typically result in computations returning \texttt{NaN}s. On start-up, we run a series of large matrix multiplications to catch these failures; during training, we also keep track of \texttt{NaN}s in communications to rapidly identify culprit nodes. Additionally, we run some simple communication tests on start-up, to ensure that the communication primitives work as expected.

\textbf{Monitoring.} We find that many web-based monitoring tools sample metrics, even when no smoothing is applied: this can hide critical events like spikes. Accordingly, we deploy our own local viewers.

\textbf{Spikes.} We encountered few spikes during training. Similar to \citet{palm}, when a spike occurred, we resumed from the latest pre-spike checkpoint and skip 1 billion tokens. Nine spikes were encountered during the training of both the 40B model and the 180B model.

\section{Results}
\label{sec:evaluation}

\textbf{Background.} The natural language processing community has developed a plethora of benchmarks to assess the capabilities of models: from tasks inspired by reading comprehension tests (e.g., RACE~\citet{lai2017large}), to world-knowledge evaluations (e.g., OpenBookQA \citet{mihaylov2press2021train018openbookqa}), or so-called commonsense tasks (e.g., HellaSwag \citep{zellers2019hellaswag}). Historical benchmarks like GLUE and SuperGLUE also aggregate many linguistically-motivated tasks, measuring abilities such as word disambiguation or recognizing entailment \citep{wang2018glue, wang2019superglue}. However, as large language models have gained in capabilities and broadened in their applications, the landscape of evaluations has shifted away from this last genre of tasks; instead, recent benchmarks such as MMLU~\citep{hendrycks2020measuring} or BigBench \citep{srivastava2023beyond} attempt to capture generic knowledge and abilities, rather than linguistic behavior--perhaps in an attempt to stay closer to common use cases of large language models. Code evaluations have also grown increasingly common \citep{chen2021evaluating}, along with mathematics tasks \citep{cobbe2021training}. Recently, technical reports on latest models have also included numerous academic and professionnal exams \citep{gpt-4, palm-2}.

Along with this shift in subjects, practices around \textit{how} models are evaluated have also changed. For zero/few-shot evaluations, the canonical setup was popularized by \citet{brown2020gpt3}: as most classic NLP tasks have a limited number of choices, these can be evaluated by calculating the log-probabilities of the choices given the question, instead of generating freeform answers. This puts strict guardrails on the model, simplifying implementation and reproduction of evaluations as they do not depend on autoregressive inference with sampling in this framing. However, this inflexibility is not always desired: for instance, chain-of-thought prompting \citep{wei2022chain} let the model write down intermediate steps of its reasoning before giving a final answer. This setup may also not best illustrate how models are used downstream; common freeform tasks, such as summarization, require autoregressive generation. However, for these tasks, a separate challenge of evaluating the model's answer rapidly arise. For instance, \citet{stiennon2020learning} found that ROUGE scores for model-generated summaries are not necessarily aligned with human preferences.  

\begin{figure}[b]
    \centering
    \vspace{-0.2in}
    \includegraphics[width=0.5\textwidth]{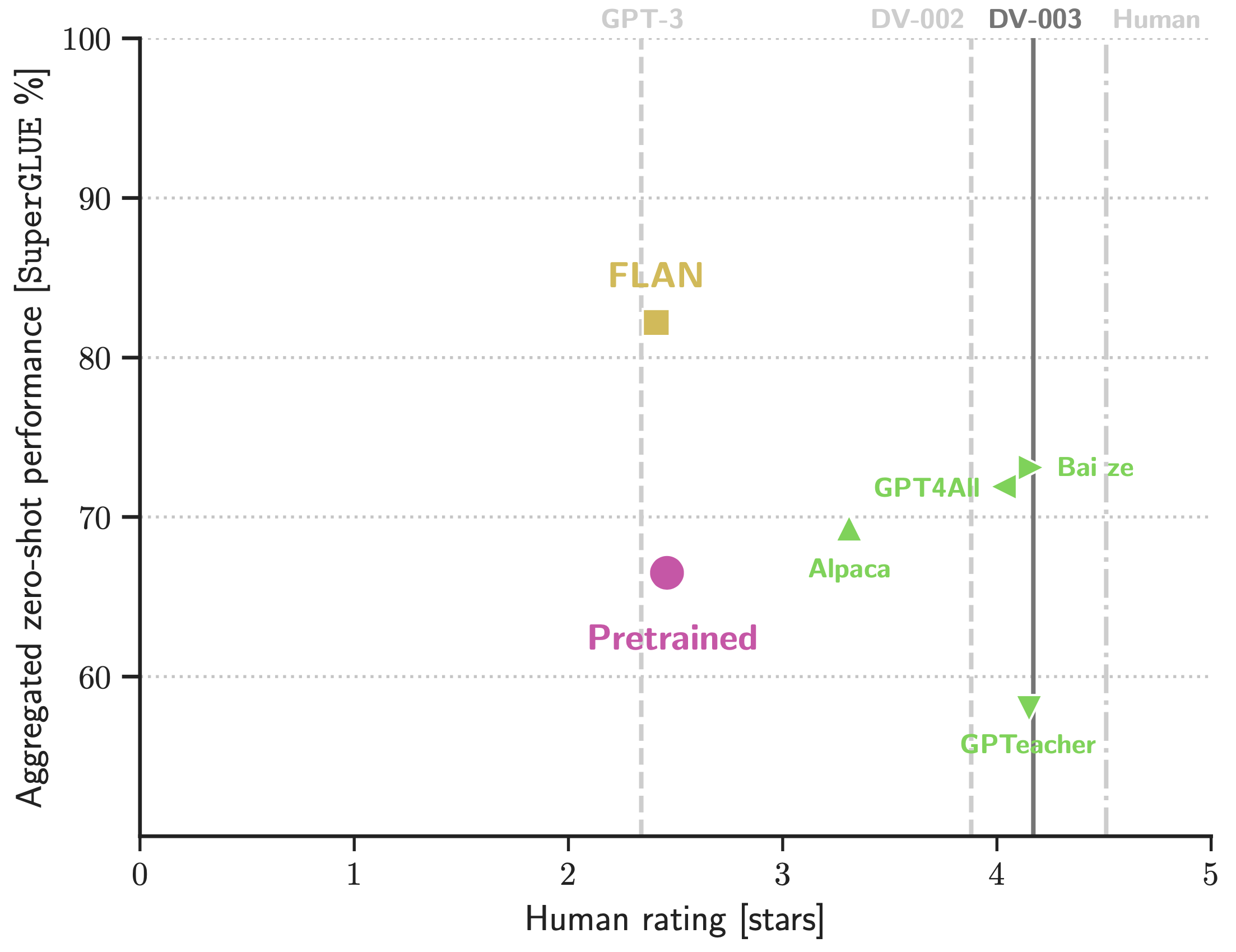}
    \caption{\textbf{\performance{Human ratings can be at odds with NLP task performance.}} Variants of Falcon-40B finetuned on different datasets, horizontal lines for baselines from the OpenAI API and from human annotators. Star ratings collected blind, from a pool of 15 annotators across 250 unique prompts.}
    \label{fig:human_pref}
\end{figure}

Typical large language models will be deployed as a chatbot/virtual assistant. This is an extremely challenging use case to quantify: it requires appropriate style and chattiness, wide world knowledge, and often reasoning/code abilities. Such deployments should also ideally be harmless and honest on top of being helpful \citep{bai2022training}, further complexifying evaluation and tradeoffs. Notably, classical NLP recipes translate poorly into better chatbots: we illustrate this in \cref{fig:human_pref}, where we evaluate variants of Falcon-40B on the popular SuperGLUE benchmark \citep{wang2019superglue}, and on a set of 250 prompts with completions rated by human annotators. The popular FLAN recipe \citep{longpre2023flan} results in the model with the best zero-shot performance on SuperGLUE; however, such a model is also the one least preferred by human annotators. Instead, models trained on so-called self-instruct~\citep{wang2022self, alpaca} datasets improve SuperGLUE performance in a smaller way, but results in significant improvements to annotator preference--unfortunately, human ratings are more expensive to collect than evaluating on SuperGLUE!

As a middle ground between expensive human preference data and cheap NLP evaluations, it has been recently proposed to use a strong external model (e.g., GPT-4) as a judge \citep{chiang2023can, vicuna2023, zheng2023judging}--broadly inspired by RLAIF, which seeks to substitute human annotators in RLHF with models themselves \citep{bai2022constitutional, dubois2023alpacafarm}. However, it is unclear yet how reliable these practices are \citep{wang2023large}--they are also predominantly focused on evaluating models finetuned for downstream usecases, not pretrained models.

Finally, fair comparisons across models are difficult. First, task selections diverge widely: the PaLM papers \citep{palm, palm-2} typically reproduce the setup of \citet{brown2020gpt3}; recent technical reports \citep{gpt-4, inflection-1} report arbitrary selection of tasks in varying settings; and state-of-the-art models like Gopher \citep{gopher}, Chinchilla \citep{hoffmann2022training}, or LLaMA \citep{llama, llama2} have all elected to report varying selections of NLP tasks. Notably, there is a lack of standardization across these setups; while standardized benchmarks exist, such as the Eleuther AI Evaluation Harness \citep{eval-harness}, or HELM \citep{liang2022holistic}, they have only seen limited adoption among open-access models--and most papers do not report details of their evaluation setup, including prompts used. Differing practices on data formatting or tokenization may result in widely different task scores \citep{whatshappening}; in a way, one-size-fits-all evaluation does not quite exist yet. We further discuss this issue in \cref{sec:prompt} with concrete examples.

\textbf{Our evaluation setup.} Since this paper focuses on the pretrained models in the Falcon series, we choose to center our evaluations on the more classic logprobs-based setup of \citet{brown2020gpt3}. This simplifies comparisons, although fairness concerns remain across models. To address these, we split our evaluations across four comparisons: (1) in 1-shot against the PaLM models \citep{palm, palm-2}, with the tasks of \citep{brown2020gpt3}; (2) on a small set of few-shot tasks reported by the GPT-4 paper \citep{gpt-4}; (3) against state-of-the-art models across common sense, question answering, and code tasks; (4) against models which also report results from the EAI Harness \citep{eval-harness}, for which we are able to compare with identical prompts and metrics. We note that this evaluation setup is only intended to provide a broad, first look at the performance of the model; adequate in-depth domain specific evaluations after dedicated specialization of the model would be required to inform about the use of the Falcon series in downstream use cases.

\vspace{-0.1in}
\subsection{To prompt or not to prompt: comparing evaluations across codebases}
\label{sec:prompt}

\begin{figure}[b]
    \centering
    \vspace{-0.1in}
    \includegraphics[width=\textwidth]{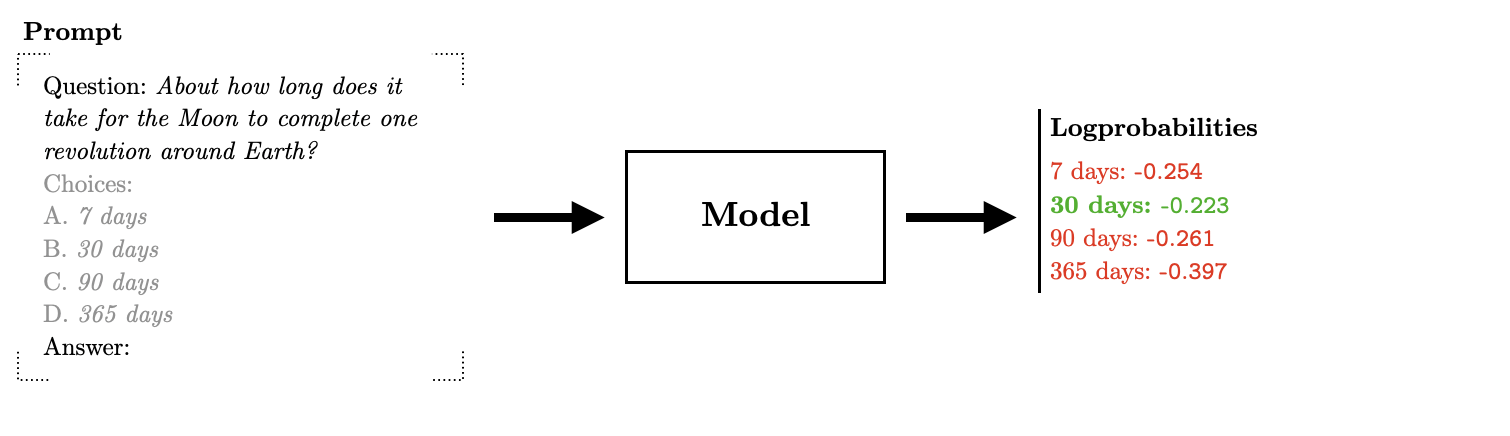}
    \caption{\textbf{\performance{In all our evaluations, we rank possible answers by their logprobabilities according to the model given a prompt.}} In {\color{Gray}grey} in the prompt, we highlight the alternative possibility of outlining each candidate explicitely in the prompt (\texttt{CH})--in line with \citet{gopher}, we find this beneficial for larger and more capable models, but detrimental for smaller ones.}
    \label{fig:prompt}
\end{figure}

When evaluating models in few-shot, two main discrepancies across setups can arise: (1) variations in the style and writing of the prompt; (2) differences in the metrics used to measure accuracy. 

\textbf{Prompting.} Specifically regarding (1), not only can small wording tweaks drive changes in performance (use of plurals, yes/no instead of true/false, typos, etc.), but tasks can also be framed in different ways. For instance \citet{gopher} found that for multiple-choice question answering tasks, writing down the options in the prompt was beneficial for larger models, but detrimental for smaller models on RACE \citep{lai2017large}--regardless of whether the model was asked to predict the answer itself or a letter key for it. In the case of RACE, this can account for an absolute change of 10-20\%. As most commonly used tasks were created before zero/few-shot evaluations became popular, there is rarely such a thing as a canonical prompt, or even framing, for a task. Furthermore, the vast majority of papers do not report the prompts they used, which hinders reproducibility. 

\textbf{Metrics.} On most tasks, accuracy is typically reported; however, practices on the calculation of the logprobabilities used to evaluate candidate answers differ. For tasks with multiple choices, $P(\text{candidate}|\text{context})$ is evaluated with each choice used as a candidate successively, and the choice with the higehst probability is taken as the model's answer. For tasks with long answer (i.e., which don't fit in a single token) the logprobabilities of each token in the candidate are summed to estimate the logprobabilities of the overall completion. When comparing choices/completions with widely differing lengths, this can be unfair to the longer ones: they end-up being more unlikely. To compensate for this, the logprobabilities are sometimes normalized by the length of the candidate answer--a first potential divergence. \citet{brown2020gpt3} also found that some answers were heavily skewed by being more likely to follow the final sequence "Answer:" in the prompt. Accordingly, for ARC, OpenBookQA, and RACE they choose to normalize the probabilities by the so-called unconditional probability of each completion: $P(\text{candidate}|\text{context}) / P(\text{candidate}|\text{"Answer:"})$. \citet{llama} also adopts this practice for OpenBookQA and BoolQ, but it's often unclear whether other papers have adopted it as well, as details on the calculations of logprobabilities are rarely provided.

\textbf{Experiments.} We explore the impact of outlining multiple-choice options in the prompt (see \cref{fig:prompt} for an illustration), and of normalizing by the unconditional probability, in zero-shot on ARC and OpenBookQA, for Falcon-7B/40B/180B in \cref{tab:evals_mess}. We find widely different effects for the two practices across both model size and tasks. On ARC-Easy, unconditional normalization (\texttt{UN}) always degrade performance, while it improves it for all models on ARC-Challenge, and improves performance only for Falcon-180B on OpenBookQA. Outlining the choices in the prompt (\texttt{CH}) always degrade performance for Falcon-7B, but always improves it for Falcon-180B; on Falcon-40B, it slightly degrades performance on ARC-Easy, and improves it significantly on ARC-Challenge and OpenBookQA. These findings are in line with the observations of \citet{gopher} on RACE, and we find that the effect is indeed very significant (+10-20\% absolute gains on Falcon-180B). Combining both practices never improves performance over using one individually. 

\begin{table}[t]
\caption{\performance{\textbf{At different model sizes, different prompt formulations and metrics calculations deliver the best zero-shot performance.}} Notably, we finding that outlining multiple-choice questions options in the prompt explicitly (\texttt{CH}) nearly always improve performance for larger models, but degrades it for smaller, less capable models. Unconditional normalization (\texttt{UN}) has a smaller effect, with improvements on some tasks but not others. Using both tricks together (\texttt{BT}) never improves performance over using the best one individually. {\color{Bittersweet} Degradation}, {{\color{ForestGreen}improvement}} over baseline: \citet{eval-harness} default prompt and length-normalized logprob calculation for \texttt{CH} and \texttt{UN}, best of \texttt{CH}/\texttt{UN} for \texttt{BT}.}
\vspace{0.1in}
\centering
\label{tab:evals_mess}
\begin{small}
\begin{tabular}{lllllllllllll}
\toprule
                     & \multicolumn{4}{l}{\textbf{ARC-Easy}}                      & \multicolumn{4}{l}{\textbf{ARC-Challenge}}       & \multicolumn{4}{l}{\textbf{OpenBookQA}}          \\
                     &                   & \texttt{UN}   & \texttt{CH}      & \texttt{BT}         &                        & \texttt{UN}        & \texttt{CH}      & \texttt{BT}    &                     & \texttt{UN}        & \texttt{CH}      & \texttt{BT}   \\ \midrule
\textbf{Falcon-7B}   & \textbf{73,7}    & \color{Bittersweet} 68,1     & \color{Bittersweet} 33,3          & \color{Bittersweet} 24,7          & \multicolumn{1}{|l}{44,5}                  & \color{ForestGreen} \textbf{49,5} & \color{Bittersweet} 27,4          & \color{Bittersweet} 24,8     & \multicolumn{1}{|l}{\textbf{44,6}}       & \color{Bittersweet} 39,2      & \color{Bittersweet} 24,4          & \color{Bittersweet} 24,0      \\
\textbf{Falcon-40B}  & \textbf{81,2}    & \color{Bittersweet} 76,6     & \color{Bittersweet} 80,5          & \color{Bittersweet} 78,8          & \multicolumn{1}{|l}{56,7}                  & \color{ForestGreen} 59,0          & \color{ForestGreen} \textbf{62,1} & \color{Bittersweet} 61,3     & \multicolumn{1}{|l}{48,0}                & \color{Bittersweet} 43,2      & \color{ForestGreen} \textbf{61,2} & \color{Bittersweet} 56,4      \\
\textbf{Falcon-180B} & 84,7             & \color{Bittersweet} 79,5     & \color{ForestGreen} \textbf{94,4} & \color{Bittersweet} 84,6  & \multicolumn{1}{|l}{63,7}                  & \color{ForestGreen} 63,9          & \color{ForestGreen} \textbf{83,5} & \color{Bittersweet} 81,6     & \multicolumn{1}{|l}{47,6}               & \color{ForestGreen} 48,8      & \color{ForestGreen} \textbf{76,4} & \color{Bittersweet} 71,8     \\
\bottomrule
\end{tabular}
\end{small}
\vspace{-0.1in}
\end{table}
\vspace{-0.05in}
For fairness, we seek to match the evaluation setting of other papers and models when comparing to their results--so that using one of the trick above doesn't put us at an unfair advantage for example.
\vspace{-0.1in}

\begin{itemize}
    \item \textbf{Comparisons with PaLM.} Like PaLM \citep{palm, palm-2} we reproduce the evaluation setup of \citet{brown2020gpt3}: we do not outline candidates in the prompt for multiple-choice questions (e.g., ARC, RACE), and use unconditionnal normalization for OpenBookQA. When candidates are longer than one token, we normalize logprobabilities by length. We do not significantly tweak prompts, staying close to the ones detailed by \citet{brown2020gpt3}. These results are presented in detail in \cref{sec:palm}.
    \item \textbf{Comparisons with GPT-3.5/GPT-4.} For MMLU, we use the format proposed by the authors originally \citep{hendrycks2020measuring}; for ARC, we explicitly outline choices in the prompt, as reported in the GPT-4 paper \citep{gpt-4}. For HellaSwag and Winogrande, we do not change the prompt used by \citet{eval-harness}. Results are reported in \cref{sec:gpt}.
    \item \textbf{State-of-the-art comparisons.} Since these are the widest comparisons, they are likely the ones with the largest variation in practices. We use unconditional normalization on OpenBookQA, and outline answers explicitely on ARC, OpenBookQA, and RACE. We lightly tweak the wording of prompts, but stay close to the ones of \citet{brown2020gpt3}, as implemented in \citet{eval-harness}. These results are reported in \cref{sec:common_sense}.
    \item \textbf{Comparisons with the EleutherAI Evaluation Harness.} We strictly use \citet{eval-harness}, and do not make any change to the prompt formulation or logprobability calculations for these. We only compare against other papers that adopt the same practice. This is our fairest set-up, for which direct one-to-one comparisons are possible and for which we didn't had to guess the set-up used by other models. These results are reported in \cref{sec:eai}.
\end{itemize}

\vspace{-0.1in}

Finally, for all set-ups, we provide the prompts used in Appendix \ref{sec:ohmyprompt} for reproducibility.

\vspace{-0.1in}

\subsection{Comparisons with PaLM on a natural language tasks aggregate}
\label{sec:palm}

\textbf{Set-up.} We compare with PaLM \citep{palm} and PaLM-2 \citep{palm-2} in the 1-shot NLP task benchmark reported in the PaLM-2 paper. This benchmark broadly reproduces the set of tasks used for GPT-3 \citep{brown2020gpt3}. We note that we are a missing a few of these tasks (e.g., StoryCloze) as we did not implement and validate them in our evaluation framework.  

\textbf{Results.} We report detailed results in \cref{tab:palm-2}. We find that when averaging performance across tasks, Falcon-180B recovers 99.5\% of the performance of PaLM-2 Large; significantly above the 94.8\% and 94.4\% of PaLM-2 Medium and PaLM. Notably, Falcon-180B even outperforms PaLM-2 on some classic benchmarks such as HellaSwag, Winogrande, and PIQA. However, Falcon-180B lags behind on two particular benchmarks: RACE and ANLI. We found RACE to be very sensitive to the formatting of the prompt and samples, which may explain some of the difference; furthermore, we observed that ANLI typically exhibits staggered progression throughout training, rather than a smooth increase (e.g., like HellaSwag)--it's possible Falcon-180B is just under the next phase transition that would allow it to catch-up to the range of performance of PaLM-2 Large.

\begin{table}[h]
\vspace{-0.1in}
\caption{\performance{\textbf{Falcon-180B matches the performance of PaLM-2 Large on most tasks, with the notable exception of ANLI and RACE for which it performs closer to PaLM-2 Medium.}} We found these tasks to be very sensitive to the prompt format used, in particular RACE, and suspect that further improvements could be unlocked with additional tweaking--however, for fairness, we stay close to the prompts proposed by \citet{brown2020gpt3}. Overall, we find that Falcon-180B recovers 99.5\% of the performance of PaLM-2 Large, significantly above the 94.8\% of PaLM-2 Medium.}
\label{tab:palm-2}
\centering
\vspace{0.1in}
\begin{tabular}{llccccc}
\toprule
\textbf{Task}                     & \textbf{Subtask} & \textbf{PaLM} & \multicolumn{3}{c}{\textbf{PaLM-2}}           & \textbf{Falcon}      \\
                     & \textbf{} & \textbf{} & \textbf{S} & \textbf{M} & \textbf{L} & \textbf{180B} \\ \midrule
WebQuestions (EM)    &           & 22,6          & 21,8              & 26,9              & 28,2              & \textbf{31,9}        \\ \midrule
HellaSwag            &           & 83,6          & 82,0              & 84,0              & 86,8              & \textbf{87,5}        \\
LAMBADA              &           & 81,8          & 80,7              & 83,7              & \textbf{86,9}     & 84,4                 \\ \midrule
WSC                  &           & 86,3          & 84,6              & \textbf{88,1}     & 86,9              & 87,5                 \\
Winogrande           &           & 83,7          & 77,9              & 79,2              & 83,0              & \textbf{85,1}        \\ \midrule
RACE                 & Hard      & 52,1          & 53,3              & 57,2        & \textbf{62,3}     & 56,7                 \\ \midrule
PIQA                 &           & 83,9          & 82,2              & 83,2              & 85,0              & \textbf{86,1}        \\
ARC                  & Challenge & 60,1          & 59,6              & 64,9              & \textbf{69,2}     & 67,8                 \\
                     & Easy      & 85,0          & 85,6              & 88,0              & \textbf{89,7}     & 88,8                 \\
                     & Overall   & 72,6          & 72,6              & 76,5              & \textbf{79,5}     & 78,3                 \\
OpenBookQA           &           & 53,6          & 57,4              & 56,2              & 58,5              & \textbf{64,2}        \\ \midrule
BoolQ                &           & 88,7          & 88,1              & 88,6              & \textbf{90,9}     & 89,0                 \\
CB                   &           & 83,9          & 82,1              & 80,4              & 87,5              & \textbf{89,3}        \\
COPA                 &           & 91,0          & 89,0              & 90,0              & \textbf{96,0}     & \textbf{96,0}        \\
RTE                  &           & 78,7          & 78,7              & \textbf{81,8}     & 79,3              & 80,1                 \\
WiC                  &           & 63,2          & 50,6              & 52,0              & \textbf{66,8}     & 66,1                 \\
ReCORD               &           & 92,8          & 92,1              & 92,4              & \textbf{93,8}     & 93,2                 \\ \midrule
ANLI                 & R1        & 52,6          & 53,1              & 58,1              & \textbf{73,1}     & 60,5                 \\
                     & R2        & 48,7          & 48,8              & 49,5              & \textbf{63,4}     & 55,5                 \\
                     & R3        & 52,3          & 53,2              & 54,5              & \textbf{67,1}     & 56,8                 \\ \midrule \midrule
\multicolumn{2}{l}{Task average}                   & 73,1          & 71,6              & 73,4              & \textbf{77,5}     & 77,1                 \\
\multicolumn{2}{l}{Fraction of PaLM-2 L}           & 94,4          & 92,4              & 94,8              &              & \textbf{99,5}       \\ \bottomrule         
\end{tabular}
\end{table}

\FloatBarrier
\newpage

Overall, these are strong scores, exhibiting the robustness of the Falcon recipe. Beside dataset composition and architectural tweaks, we note two differences between PaLM-2 Large and Falcon-180B: (1) PaLM-2 L was supposedly trained with a larger compute budget, with twice the parameters and a similar amount of tokens as Falcon-180B (see \cref{sec:unofficial}); (2) PaLM-2 models used a mixture of objectives \citep{tay2022ul2}, which has been reported to enhance downstream task performance. Further a posteriori adaptation of Falcon-180B (e.g., with the PaLM-U recipe \citep{tay2022transcending}) could help recover more of the performance of PaLM-2 L with Falcon-180B as a base model.

\subsection{Comparisons with GPT-3.5 and GPT-4 on a limited set of tasks}
\label{sec:gpt}

\textbf{Set-up.} We compare with GPT-3.5 and the GPT-4 as reported in GPT-4 paper \citep{gpt-4}, focusing on natural language tasks: MMLU, HellaSwag, Winogrande, and ARC. We use the same number of shots as proposed, except for ARC, for which we report 2-shot instead of 25-shot accuracy. 

\textbf{Results.} Results are in \cref{tab:gpt4}. We find that Falcon-180B systematically performs above GPT-3.5, but below GPT-4 on all tasks. Notably, Falcon-180B achieves strong performance on commonsense tasks: on HellaSwag, it is close to midway between GPT-3.5 and GPT-4, while on Winogrande, it nearly matches the performance of GPT-4. We find the performance of Falcon-180B on multiple choice question answering tasks to be closer to GPT-3.5, albeit always higher. We note that GPT-4 was allegedly trained with 4-5x more pretraining compute than Falcon-180B (see \cref{sec:unofficial}), which likely contributes to most of the difference in performance between the two models.

\begin{table}[t]
\caption{\textbf{\performance{Falcon-180B delivers downstream performance between GPT-3.5 and GPT-4.}} Falcon-180B performs well on commonsense tasks (HellaSwag and Winogrande), where it is well ahead of GPT-3.5. For multiple choice question answering (ARC and MMLU), Falcon-180B performs above GPT-3.5 but not as significantly so.$^*$ we report 2-shot, not 25-shot performance on ARC Challenge.}
\label{tab:gpt4}
\vspace{0.1in}
\centering
\begin{tabular}{lccc}
\toprule
                                 & \textbf{GPT-3.5} & \textbf{GPT-4} & \textbf{Falcon-180B} \\ \midrule
\textbf{HellaSwag} (10-shot)     & 85.5             & \textbf{95.3}           & 89.0                 \\
\textbf{Winogrande} (5-shot)     & 81.6             & \textbf{87.5}           & 87.1                 \\
\textbf{ARC Challenge} (25-shot) & 85.2             & \textbf{96.3}           & 87.8$^*$             \\
\textbf{MMLU} (5-shot)           & 70.0             & \textbf{86.5}           & 70.6    \\ \bottomrule            
\end{tabular}
\end{table}

\subsection{State-of-the-art comparisons on common sense, question answering, and code tasks}
\label{sec:common_sense}

\textbf{Set-up.} In this section, we compare the Falcon series with other models on commonsense, question answering, and code tasks. We compare with models that have successively defined the state-of-the-art "before" PaLM-2 Large and GPT-4: GPT-3 \citep{brown2020gpt3}, Gopher \citep{gopher}, Chinchilla \citep{hoffmann2022training}, MT-NLG \citep{smith2022using}, PaLM \citep{palm}, LLaMA-2 \citep{llama2}, and Inflection-1 \citep{inflection-1}. For commonsense tasks, we include PIQA \citep{bisk2020piqa}, HellaSwag \citep{zellers2019hellaswag}, Winogrande \citep{sakaguchi2019winogrande}, BoolQ \citep{clark2019boolq}, and LAMBADA \citep{paperno2016lambada}--only for BoolQ does our prompt differ slightly from the default one of the EleutherAI Evaluation Harness \citep{eval-harness}. For question answering datasets, we include ARC \citep{clark2018arc}, OpenBookQA \citep{mihaylov2press2021train018openbookqa}, and MMLU \citep{hendrycks2020measuring}. To enable fair comparisons, we evaluate ARC without outlining choices (see \cref{sec:prompt}). For OpenBookQA, we outline choices, as we find performance to be flat otherwise; finally, for MMLU, we use the canonical setup from~\citet{hendrycks2020measuring}. For all these tasks we take results for other models from the papers outlined above. For code, we reproduce the set-up of the BigCode Models Leaderboard for HumanEval in Python, and include further code-specialized models such as Codex \citep{chen2021evaluating}, StarCoder \citep{li2023starcoder}, and Code LLaMA \citep{roziere2023code}. As we focus on comparisons with the Falcon models immediately after pretraining, we do not consider instruct variants of the models above.

\textbf{Commonsense.} We report results in \cref{tab:broad_tasks}. With the exception of BoolQ, Falcon-180B improves significantly over state-of-the-art models across all tasks. Broadly speaking, we find that Falcon-40B is slightly under LLaMA-2 34B, which we attribute to smaller pretraining compute: 2,800PF-days for Falcon-40B against 4,700PF-days for LLaMA-2 34B (nearly 70\% more). Falcon-7B also slightly underperforms LLaMA-2 7B; this time the difference in compute is smaller (730PF-days against 970PF-days, 30\% more), but we suspect that multiquery with a single head of dimension 64 is a very aggressive configuration for Falcon-7B. We find that LLaMA-2 and Inflection-1 achieve relatively similar performance, with Inflection-1 perhaps ahead thanks to exceptional performance on BoolQ.

\textbf{Question answering.} In \cref{tab:qa}, we find again that Falcon-180B strongly outperforms other models from the state-of-art on question answering tasks. We do note that the Falcon series seems to underperform slightly (comparatively to other tasks) on MMLU: we believe this may be attributed to the large prevalence of web data in our pretraining dataset, compared to more technical sources which may be more immediately relevant to the style and content of questions found in MMLU. 

\begin{table}[t]
\caption{\textbf{\performance{Outside of PaLM-2 Large and GPT-4, Falcon-180B significantly improves other state-of-the-art models such as LLaMA-2 or Inflection-1 on commonsense tasks}}. Falcon-40B performs slightly under LLaMA-2 34B, because of a significantly smaller compute budget (2,800PF-days against 4,700PF-days, 70\% more for LLaMA-2). We note the exceptional performance of Inflection-1 on BoolQ; conversely, we note that, despite our best prompt engineering efforts, we were unable to reproduce the performance reported by the LLaMA-2 paper on BoolQ for the Falcon series: the results we report can likely be improved. \textbf{Bold} for best, \underline{underline} for second-best.}
\label{tab:broad_tasks}
\vspace{0.1in}
\centering
\begin{tabular}{llccccccc}
\toprule
                      &      & \textbf{PIQA} & \multicolumn{2}{l}{\textbf{HellaSwag}} & \multicolumn{2}{l}{\textbf{Winogrande}} & \textbf{BoolQ} & \textbf{LAMBADA} \\
                      &      &               &                 & \textit{(10-shot)}   &                  & \textit{(5-shot)}    &                &                  \\ \midrule
\textbf{GPT-3}        &      & 81,0          & 78,9            &                      & 70,2             &                      & 60,5           & 76,2             \\
\textbf{Gopher}       &      & 81,8          & 79,2            &                      & 70,1             &                      & 79,4           & 74,5             \\
\textbf{Chinchilla}   &      & 81,8          & 80,8            &                      & 74,9             &                      & 83,7           & 77,4             \\
\textbf{MT-NLG}       &      & 82,0          & 80,2            &                      & 73               &                      & 78,2           & 76,6             \\
\textbf{PaLM}         &      & 82,3          & 83,4            &                      & 81,1             &                      & 88,0           & 77,9             \\
\textbf{LLaMA-2}      & 7B   & 78,8          & 77,2            & 78,6                 & 69,2             &                      & 77,4           &                  \\
\textbf{}             & 13B  & 80,5          & 80,7            & 82,1                 & 72,8             &                      & 81,7           &                  \\
\textbf{}             & 34B  & 81,9          & 83,3            &                      & 76,7             &                      & 83,7           &                  \\
\textbf{}             & 70B  & 82,8          & \underline{85,3}            & \underline{87,3}                 & \underline{80,2}    &                      & 85,0           &                  \\
\textbf{Inflection-1} &      & \underline{84,2}          & 84,3            & 85,8                 &                  & \underline{83,3}                 & \textbf{89,7}  & \underline{78,5}             \\ \midrule
\textbf{Falcon}       & 7B   & 80,3          & 76,3            & 78,1                 & 67,2             & 72,6                 & 73,8           & 74,9             \\
\textbf{}             & 40B  & 83,0          & 82,7            & 85,3                 & 76,0             & 81,8                 & 81,9           & 77,3             \\
\textbf{}             & 180B & \textbf{84,9} & \textbf{85,9}   & \textbf{89,0}        & \textbf{80,3}    & \textbf{87,1}        & \underline{87,8}           & \textbf{79,8}   \\ \bottomrule
\end{tabular}
\end{table}

\begin{table}[b]
\vspace{-0.2in}
\caption{\textbf{\performance{Falcon-180B improves significantly over GPT-3, PaLM, and LLaMA-2 on question answering datasets, while Falcon-40B performs in-line with LLaMA-2 34B.}} $^*$: for Falcon-7B on OpenBookQA, we report accuracy without outlining candidates, as performance is otherwise close to random (see \cref{tab:evals_mess} for details). \textbf{Bold} for best, \underline{underline} for second-best.}
\label{tab:qa}
\centering
\vspace{0.1in}
\begin{tabular}{llcccc}
\toprule
                 &      & \textbf{ARC-Challenge} & \textbf{ARC-Easy} & \textbf{OpenBookQA} & \textbf{MMLU} \\ \midrule
\textbf{GPT-3}   &      & 51,4                   & 68,8              & 57,6                &               \\
\textbf{PaLM}    &      & 53,0                   & 76,6              & 53,4                & 69,3          \\
\textbf{LLaMA-2} & 7B   & 45,9                   & 75,2              & 58,6                & 45,3          \\
\textbf{}        & 13B  & 49,4                   & 77,3              & 57,0                & 54,8          \\
\textbf{}        & 34B  & 54,5                   & 79,4              & 58,2                & 62,6          \\
\textbf{}        & 70B  & \underline{57,4}             & 80,2              & 60,2                & \underline{68,9}    \\ \midrule
\textbf{Falcon}  & 7B   & 44,5                   & 73,6              & 44,6$^*$            & 28,0          \\
\textbf{}        & 40B  & 56,7                   & \underline{81,2}        & \underline{61,2}          & 57,0          \\
\textbf{}        & 180B & \textbf{63,7}          & \textbf{84,7}     & \textbf{76,4}       & \textbf{70,6} \\ \bottomrule
\end{tabular}
\end{table}

\textbf{Code.} We report results on HumanEval in \cref{tab:code}. We find that Falcon-180B performs best amongst models focusing on natural language, with performance only matched by Inflection-1. In fact, despite being trained on only 3\% code, Falcon-180B nearly matches the performance of both PaLM-Coder and PaLM-2 S$^*$, two models which have undergone dedicated code specialization following their pretraining. This is an encouraging result for the development of a Falcon-Coder specialization.

\begin{table}[t]
\caption{\textbf{\performance{Falcon-180B outperforms all other predominantly English language models on HumanEval. Morever, despite being trained on only 3\% code, it nearly matches the performance of PaLM-Coder and PaLM-2 S$^*$, both having undergone code specialization.}} $^\dagger$: the GPT-3.5 and GPT-4 models are somewhat unique in being pretrained on a large fraction of code on top of natural language \citep{openai_modelindex}. \textbf{Bold} for best, \underline{underline} for second-best (per specialization), pass@1.}
\label{tab:code}
\centering
\vspace{0.1in}
\begin{tabular}{llcc}
\toprule
                           &             & \textbf{Specialized for code?} & \textbf{HumanEval} \\ \midrule
\textbf{PaLM}              &             &                                & 26,2               \\ 
\textbf{LLaMA-2}           & 7B          &                                & 12,2               \\
\textbf{}                  & 13B         &                                & 20,1               \\
\textbf{}                  & 34B         &                                & 22,6               \\
\textbf{}                  & 70B         &                                & \underline{30,5}               \\
\textbf{Inflection-1}            &        &                                & \textbf{35,4}               \\
\textbf{Falcon}            & 180B        &                                & \textbf{35,4}               \\ \midrule

\textbf{Codex}             & \texttt{cushman-001} & $\checkmark$                   & 33,5               \\
                           & \texttt{davinci-002} & $\checkmark$                   & 45,9               \\
\textbf{StarCoder}         &             & $\checkmark$                   & 30,4               \\
\textbf{PaLM-Coder}        &             & $\checkmark$                   & 35,9               \\
\textbf{PaLM-2 S$^*$}      &             & $\checkmark$                   & 37,6               \\
\textbf{GPT-3.5} &             & $\checkmark^\dagger$                 & 48,1               \\
\textbf{GPT-4} &             & $\checkmark^\dagger$                 & \textbf{67.0}               \\
\textbf{Code LLaMA}         & 7B          & $\checkmark$                   & 33,5               \\
                           & 13B         & $\checkmark$                   & 36,0               \\
                           & 34B         & $\checkmark$                   & \underline{48,8}               \\ \bottomrule

\end{tabular}
\end{table}

\subsection{Comparison with other models using the EleutherAI Evaluation Harness}
\label{sec:eai}

\textbf{Set-up.} For this final comparison, we only consider models which have had reports of evaluations performed with the EleutherAI Evaluation Harness \citep{eval-harness}. These results are the most directly comparable, as the preprocessing of samples, the prompt, and the calculations of the metrics are identical. We report results for GPT-3 evaluated through the API by the BigScience group, FairSeq~\citep{artetxe2021efficient}, GPT-Neo-1.3B \citep{gpt-neo}, GPT-J \citep{gpt-j}, GPT-NeoX-20B \citep{black2022gpt}, OPT \citep{zhang2022opt}, Pythia \citep{biderman2023pythia}, CerebrasGPT \citep{dey2023cerebras}, Aleph Alpha \citep{alephalpha}, and BLOOM \citep{bloom}. We average performance on HellaSwag \citep{zellers2019hellaswag}, LAMBADA \citep{paperno2016lambada}, Winogrande \citep{sakaguchi2019winogrande}, PIQA \citep{bisk2020piqa}, ARC \citep{clark2018arc}, and OpenBookQA \citep{mihaylov2press2021train018openbookqa}--this was the widest set of tasks we could assemble based on different reporting practices across papers. We note that these models mostly span a smaller compute range (up to a few thousands PF-days at most), and with performance starkly below other state-of-the-art models on with which we have compared before. We report results both for the Falcon series presented in this paper, and for the smaller-scale Falcon-RefinedWeb models trained in \citet{refinedweb}, which used only the RefinedWeb dataset for performance validation.

\begin{figure}[t]
    \centering
    \includegraphics[width=\textwidth]{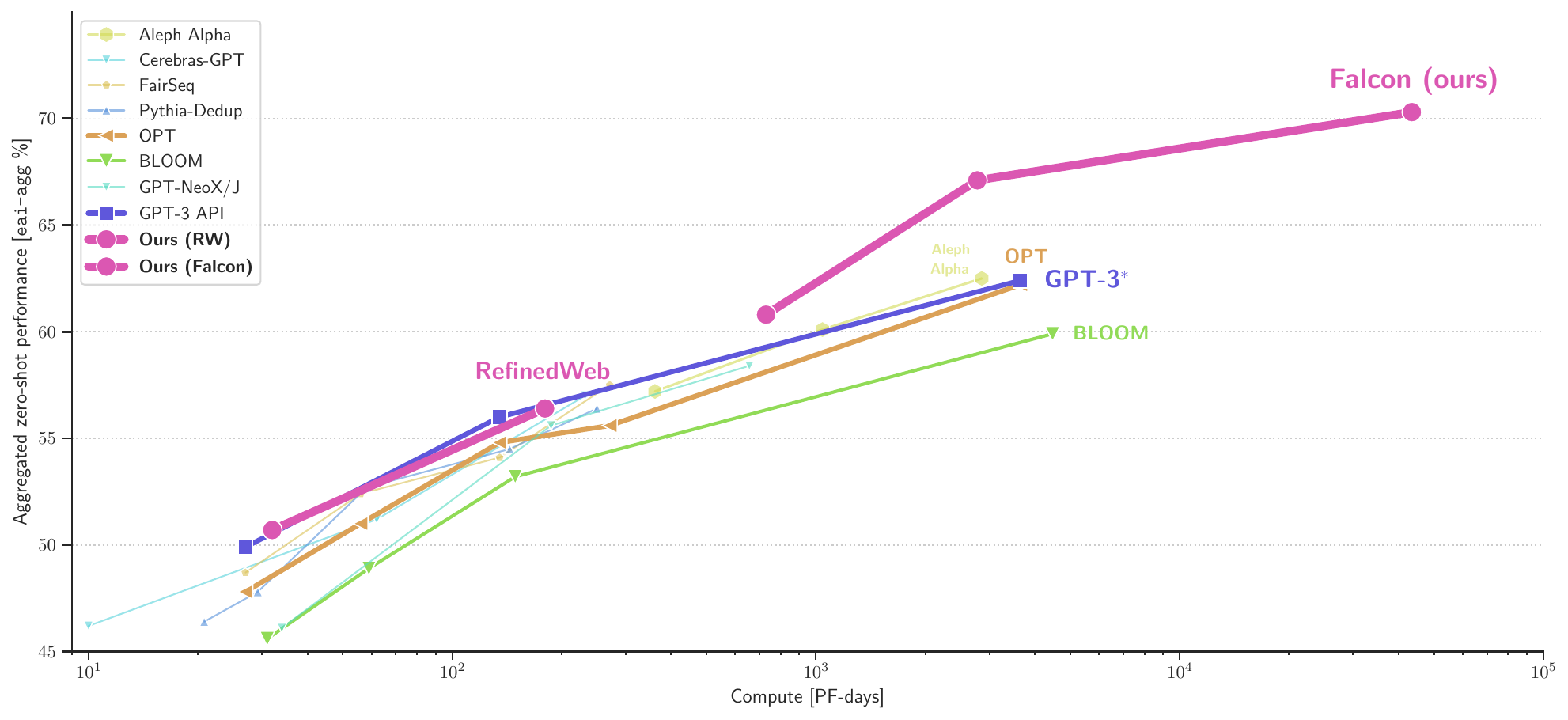}
    \caption{\textbf{\performance{The Falcon series strongly improves at all scales against other models which have reported results with the EleutherAI Evaluation Harness.}} Models trained on RefinedWeb alone also reproduce the performance of the GPT-3 series, despite not using curated data. Aggregated zero-shot accuracy on HellaSwag, LAMBADA, Winogrande, PIQA, ARC, and OpenBookQA.}
    \label{fig:eai_results}
\end{figure}

\textbf{Results.} We present results in \cref{fig:eai_results}. We find that, across scales, the Falcon series significantly improves against other models in this set of comparisons. Notably, Falcon-40B outperforms GPT-3 175B, despite being trained with a smaller compute budget. In fact, even Falcon-7B approaches the performance of GPT-3 175B: we believe that with longer training and a less aggressive multiquery setup, it should be possible to match the performance of the original GPT-3 model with 7B parameters (or less). Finally, we also note that the smaller validation models trained on RefinedWeb alone also compare favorably, reproducing the performance of the GPT-3 models of the same size. 

Taking a step back, it's interesting to note some broader trends in this plot. Most older series of models achieve very similar performance, with the GPT-3 series as a roofline. The OPT models, despite underperforming at the smaller sizes, eventually match the performance of GPT-3 175B. Two outsiders stand out: GPT-Neo-1.3B, the first open-source large language model, likely because of issues in its pretraining, and the BLOOM series, likely because of its heavily multingual pretraining data and conservative use of an additionnal layer norm after the embeddings \citep{what_lang_model}.

\section{Limitations}
\label{sec:limitations}

We highlight in this section limitations of our findings and of the Falcon series of models itself, both to nuance potential applications of our models, and to highlight further research directions. Broadly speaking, these limitations stem from two factors: (1) most of the research on the Falcon series was conducted prior to December 2022, when model training started--in the meanwhile significant developments have occurred in the world of large language models; (2) constraints in compute resources preventing a more exhaustive exploration of some directions.

\subsection{Limitations of our findings and ablations}

\textbf{Scale.} Our ablations are conducted with 1B and 3B models trained to optimality; although we validated the final architecture and datasets with 1B and 7B models trained for 350B tokens, this remains order of magnitude below the compute budget of the actual models in the Falcon series. On the model side, as scale increases, other works have noted the emergence of outlier features, which may drive important dynamics which are not captured by our set-up~\citep{dettmers2022gpt3}. On the data side, increasingly large models are known to become increasingly sensitive to memorization~\citep{carlini2022quantifying}, potentially leading to catastrophic quality degradation \citep{hernandez2022scaling}. 

\textbf{Benchmarks.} In our ablations, we focused on measuring zero-shot performance on a limited set of tasks, using only logprobs-based evaluations. Not only can downstream task performance be at odds with actual human preference (see \cref{fig:human_pref} for our own experience of that issue), but our set of tasks does not capture code or multilingual performance. Evaluating general purpose language models is difficult, and single aggregated metrics often poorly translate the nuances brought forth by modeling or data interventions. Broadly speaking, on our set of tasks, we never experienced a consistent trend with one intervention systematically improving a specific metrics but not the others (e.g., adding technical data did not consistently uplift performance on SciQ or PubMedQA); however, this is likely to be a limitation of our small-scale setup, which end-up predominantly measuring general gains. It is likely the performance on popular benchmarks such as MMLU \citep{hendrycks2020measuring} or AGIEval \citep{zhong2023agieval} could be improved with dedicated in-domain data. We notably believe in the value of larger-scale ablations on optimal pretraining data composition, and on better understanding quantitative versus qualitative aspects of deduplication (i.e., deduplication on web data may be unreasonably effective because it happens to filter out some of the worst samples). 

Overall, we view \cref{sec:ablations} as a set of cursory experiments that helped us ground and broadly validate some decisions for the Falcon series. Future practitioners will likely want to revisit these experiments and results at increased scale, enabling broader benchmarking to better capture performance.

\subsection{Limitations of the Falcon models}

\textbf{Decoupling training and inference compute.} In the past year, there has been a dramatic explosion in the adoption of large language models, leading to significantly increased downstream use. Under that paradigm, inference costs can become predominant, shadowing pretraining costs; furthermore, smaller open-source models are easier for the community to build upon, enabling local deployments and even finetuning. When scaling-up pretraining, additional compute may be either spent towards a larger model, or towards longer training. While increasing parameter count caries over to inference, increasing costs over the entire lifecycle of the model, this is not the case for longer pretraining, offering an axis for decoupling training and inference compute. Recent open-source models, such as the LLaMA series \citep{llama, llama2}, have adopted this practice by eventually training their 7-70B parameters models for a fixed 2,000B tokens. While our 7B follows a similar idea, our 40B and 180B models are trained closer to the pretraining optimality suggested by \citet{hoffmann2022training}. This makes deployment of models in the Falcon series more challenging. At the time, this decision was partially motivated by constraints over data availability: we elected early on to strictly stick to a single epoch of training, and the total stock of RefinedWeb tokens was unknown to us when we started training (data processing jobs only concluded in early January). Recent works have however suggested that up to 4 epochs may lead to minimal degradation in performance, at least for models in the 10B parameters range \citep{token_crisis, data_constrained}. Based on the final size of our available data (around 5,000B tokens of English and 1,000B tokens of code) we would recommend training future models on at least 4,000-5,000B tokens, and skewing towards 10,000-15,000B tokens for the larger ones. Note that this does pose some challenges during pretraining, as larger models are easier to distribute efficiently. We also see promise in enabling training and inference compute decoupling through architecture interventions such as layerwise mixture of experts (MoE), also known as routed language models \citep{clark2022unified, fedus2022switch}. In MoE models, parameters are only sparsely activated, allowing for another 8-16 reduction in inference costs. 

\textbf{Code performance and pretraining data.} As we did not allow upsampling for pretraining subsets, we predominantly trained Falcon on web data from RefinedWeb. We also were conservative with the fraction of code in the pretraining, whereas available data could have allowed us to reach 10-30\% of code data in pretraining. For future models, we would recommend making code data significantly more prevalent, and potentially upsampling sources illustrative of common usecases or data domains of interest for downstream use cases. Code data is particularly promising, because it is widely available from a single source (i.e., GitHub) and can undergo large-scale processing similar to web data--deduplication is also extremely effective for code data \citep{kocetkov2022stack, allal2023santacoder, li2023starcoder}. Furthermore, the GPT-3.5 and GPT-4 models have been seemingly trained on a mix of both natural language and code data \citep{openai_modelindex, gpt-4}. Such hybrid language/code models are also likely to be easier to adapt for downstream use cases as chatbots which may make extensive use of code to answer user questions, or to interface with a variety of tools. We also see potential promise in the use of synthetic data \citep{gunasekar2023textbooks, li2023textbooks}; however, many of these methods are currently closer to distillation than true pretraining, and it is somewhat unclear how they may be used to bootstrap a new class of larger models beyond downstream adaptation.

\textbf{Longer sequence lengths.} All Falcon models are pretrained with a 2,048 sequence length, which can be limiting for multi-turn chat or code use cases. Thanks to the use of rotary embeddings \citep{su2021roformer}, a posteriori adaptation to longer sequences is possible: \citet{rope_extension, chen2023extending} concurrently found that extending the context length with rotary positionnal embeddings was possible by interpolation and lightweight finetuning. In our own experiments on long context finetuning, we found long context data to be plentiful in RefinedWeb. Upward of and more than 13\% of the tokens come from documents of over 8k tokens, with 1.5\% of the tokens in documents of over 32k tokens. This approach was further refined for LLaMA-2 Long \citep{xiong2023effective}, resulting in a significant performance boost on tasks with many shots (e.g., MMLU and ARC in \cref{sec:gpt}).

\textbf{Pretrained only.} Although we make lightly tuned versions of the Falcon models available to demonstrate possibilities as instruct or chat models, our work predominantly concerns the pretrained versions. Prior to deployment of the Falcon models, we strongly recommend for adequate guardrails to be deployed and for bias/harm evaluations specific to the target use case to be systematically undertaken. In particular, we note that reinforcement learning from human feedback may be relevant for not only making the models more helpful, but also more robust to adversarial queries \citep{ganguli2022red}. We also see promise in code adaptation of the Falcon models, similar to PaLM-Coder \citep{palm} and PaLM-2 S$^*$ \citep{palm-2} or Code LLaMA \citep{roziere2023code}.

\section{Conclusion}

In this paper, we have introduced and extensively described the Falcon series of pretrained models, with Falcon-7/40/180B, trained on up to 3,500B tokens. 

First, we described some of the ablations and experiments we performed to prepare the training dataset and architecture of the Falcon models. We found adequately filtered and deduplicated web data to be a surprisingly strong baseline; concerns around memorization for larger models \citep{carlini2022quantifying, hernandez2022scaling} led us to elect not to upsample any sources, and to predominantly train on this web data. On the architecture-side, we found most interventions to have a limited effect, and adopted multigroup attention (an extension of multiquery \citep{multiquery}) as a way to improve inference scalability by significantly reducing the size of the K,V-cache. For future generations of the Falcon series, we see promise in significantly increasing the fraction of code in the pretraining data, and in training with longer sequence lengths in a staged process (e.g., half of the training up to 8k, and second half up to 16k, with downstream adaptation to 32-256k).

Then, we described our implementation of our final strategy for the pretraining of the Falcon series. We report extensively on our data pipeline in \citet{refinedweb}. We described our approach to conversational masking, and our distributed training strategy for running on cost-efficient cloud infrastructure, relying notably on 3D parallelism combined with ZeRO. We also discussed some of our interventions for fast memory efficient training, such as dedicated FlashAttention \citep{flash_attn} kernels in Triton \citep{triton} and our monolayer strategy. We also discussed some of the details around setting hyper-parameters and managing runs over multiple thousand GPUs.

Finally, we outlined some of the results obtained by the Falcon series on key benchmarks. We found Falcon-180B to near the performance of PaLM-2 Large \citep{palm-2}, and to end-up in-between GPT-3.5 and GPT-4 \citep{gpt-4}. Falcon-180B is the best open-source model currently available, and likely one of the best models overall. We note our evaluation predominantly focuses on classic natural language tasks, and that further work will be required for evaluating human preferences on downstream versions of Falcon having undergone dedicated finetuning or reinforcement learning.

To foster open research on large language models, and accelerate technological development in this space, we make the following artefacts public available under permissive open-source licenses:

\begin{itemize}
    \item \textbf{Falcon-7/40/180B.} We make all models in the Falcon series available, with Falcon-7/40B under an Apache 2.0 license and Falcon-180B under a dedicated responsible AI use license. At time of release, Falcon-180B is the most powerful open large language model available.
    \item \textbf{A 600B tokens extract of RefinedWeb.} We make a 600B tokens extract of our web dataset available, for use by researchers to study large-scale filtered and deduplicated web data, and for other practioners to adopt as a standard for high-quality web data. We also open-source 1/7B models trained on 350B tokens from this extract.
    \item \textbf{Detailed research.} With this paper and the RefinedWeb paper \citep{refinedweb}, we have detailed numerous of our decisions and experiments concerning the Falcon series.
\end{itemize}

We believe large language models to be a foundational technology for the future of our civilization, and in turn we believe they should be  shared responsibly. Widespread exchange of ideas is a staple of accelerated technological and economical progress in our history; in turn, this acceleration uplifts all. By open-sourcing artificial intelligence research and models, we can foster a broader and more diverse community, and benefit from vibrant collaborative efforts to improve the safety and reliability of large language models. We hope the Falcon series can be a small step towards this vision.

\newpage

\bibliography{refs}
\bibliographystyle{apalike}

\appendix

\newpage

\section{Contributions}
\label{sec:contributions}

\begin{itemize}
    \item \textbf{\textsc{Engineering \& Tooling.}} 
    \begin{itemize}
        \item \textbf{Distributed training codebase.} Baptiste Pannier, Daniel Hesslow. 
        \item \textbf{Web data.} Guilherme Penedo, Ruxandra Cojocaru \emph{(multilingual data)}, Quentin Malartic \emph{(data quality)}, Alessandro Cappelli, Hamza Alobeidli. 
        \item \textbf{Curated data.} Alessandro Cappelli, Etienne Goffinet \emph{(code data)}, Quentin Malartic, Ruxandra Cojocaru, Abdulaziz Alshamsi \emph{(books data)}.
        \item \textbf{Inference.} Daniel Hesslow, Baptiste Pannier, Guilherme Penedo \emph{(deployment)}.
        \item \textbf{Infrastructure.} Daniele Mazzotta, Etienne Goffinet, Guilherme Penedo.
        \item \textbf{Hardware correctness.} Baptiste Pannier, Julien Launay, Daniel Hesslow.
    \end{itemize}
    \item  \textbf{\textsc{Pretraining.}}  Baptiste Pannier, Julien Launay, Daniel Hesslow.
    \item \textbf{\textsc{Research.}} 
    \begin{itemize}
    \item \textbf{Data ablations.} Julien Launay, Ruxandra Cojocaru \emph{(curriculum learning)}, Alessandro Cappelli, Quentin Malartic \emph{(fine-grained filters)}, Daniel Hesslow.
    \item \textbf{Architecture ablations.} Julien Launay, Daniel Hesslow, Etienne Goffinet, Quentin Malartic \emph{(training objectives)}, Baptiste Pannier, Badreddine Noune \emph{(optimizers)}.
     \item \textbf{Model finetuning.} Quentin Malartic, Alessandro Capelli, Etienne Goffinet \emph{(code specialization)}, Guilherme Penedo \emph{(human evaluation)}, Baptiste Pannier \emph{(long-context)}, Julien Launay.
     \item \textbf{Evaluation.} Julien Launay, Daniel Hesslow, Quentin Malartic.
     \item \textbf{Paper writing.} Julien Launay, Daniel Hesslow, Alessandro Cappelli, Baptiste Pannier, Ebtesam Almazrouei.
    \end{itemize}
     \item \textbf{\textsc{Leadership.}}  Julien Launay, Ebtesam Almazrouei, Mérouane Debbah.
\end{itemize}

\section{Acknowledgements}

We would like to thank the AWS team, in particular Olivier Cruchant, for their support throughout the project, enabling us to eventually scale to up to 4,096 A100s the training of Falcon-180B. We would also like to thank Axel Marmet, Tri Dao, Dan Fu, Colin Raffel, Katherine Lee, Thomas Wolf, Iz Beltagy, and Dirk Groeneveld for insightful discussions throughout the project.

\newpage

\vspace{-0.2in}
\section{Model card}
\vspace{-0.2in}

\begin{longtable}{p{4cm}|p{9cm}}
    \toprule
    \multicolumn{2}{c}{\textsc{\textbf{Model details}}} \\
    \midrule
    \textbf{Organization} & The models were created by the Technology Innovation Institute. \\ \midrule
    \textbf{Model date} & Training of the Falcon models started in December and completed in the first half of 2023. \\ \midrule
    \textbf{Model type and information about training} & Falcon are autoregressive Transformer models trained with a causal language modeling objective. Architecture based on PaLM~\cite{palm}, with an extension of multiquery attention for tensor parallelism (multigroup) and minor tweaks (no SwiGLU, etc.). See \cref{sec:ablations} and \cref{sec:implementation} for details. \\ \midrule
    \textbf{Licence} & Falcon-7B and Falcon-40B are made available under the Apache 2.0 license; Falcon-180B is made available under the Falcon-180B TII license, with restrictions related to responsible use. \\ \midrule
    \textbf{Point of contact} & falconllm@tii.ae \\ \midrule
    \multicolumn{2}{c}{\textsc{\textbf{Intended use}}} \\ \midrule
    \textbf{Primary intended uses} & Research on large language models; as a foundation for further specialization for specific use cases (e.g., chatbot, etc.) \\ \midrule
    \textbf{Primary intended users} & NLP researchers and engineers. \\ \midrule
    \textbf{Out-of-scope use cases} & Production use without adequate assessment of risks and mitigation; use cases which may be considered irresponsible or harmful. \\ \midrule
    \multicolumn{2}{c}{\textsc{\textbf{Factors}}} \\ \midrule
    \textbf{Relevant factors} & The Falcon models are predominantly trained on English data from a large-scale web corpora representative of the web. Accordingly, they will carry the stereotypes and biases commonly encountered online, and are unlikely to generalize appropriately beyond English or European latin languages. \\ \midrule
    \textbf{Evaluation factors} & We evaluated the toxicity of the underlying pretraining dataset and found it to be in line with common curated pretraining datasets such as The Pile, see \citet{refinedweb}. Note that this only accounts for toxicity under the definition of Perspective API: "content that is rude or disrespectful". Notably, this fails to include concerns about social biases or harmfulness. \\ \midrule
    \multicolumn{2}{c}{\textsc{\textbf{Metrics}}} \\ \midrule
    \textbf{Model performance measures} & We focus our evaluation on the zero-shot generalization capabilities of our models across a wide range of tasks, leveraging the Eleuther AI language model evaluation harness \cite{eval-harness}.  \\ \midrule
    \textbf{Variation approaches} & Due to the costs associated with training Falcon we cannot train the models multiple times and measure variability across runs. \\ \midrule
    \multicolumn{2}{c}{\textsc{\textbf{Evaluation data}}} \\ \midrule
    \textbf{Datasets} & We evaluate zero-shot accuracy on 18 natural language tasks and one Python programming task, detailed in \cref{sec:evaluation}. \\ \midrule
    \textbf{Motivation} & We selected and aggregated tasks to build comparisons with other models in the literature (see \cref{sec:prompt}). \\ \midrule
    \textbf{Preprocessing} & We mostly use the default setup of \cite{eval-harness}, see \cref{sec:ohmyprompt} for the custom prompts we used for some evaluations \\ \midrule
    \multicolumn{2}{c}{\textsc{\textbf{Training data}}} \\ \midrule
    \multicolumn{2}{c}{\textbf{See the dedicated datasheet in \citet{refinedweb}.}} \\
    \bottomrule
    \caption{\textbf{Model card for Falcon}, following the framework introduced by \cite{mitchell2019model}.}
    \label{tab:model_card}
\end{longtable}

\section{Datasheet}

See the dedicated RefinedWeb paper \citep{refinedweb}.

\section{Comparisons with undocumented models}
\label{sec:unofficial}

A number of recent models have elected to release scant promotional technical reports instead of adequately documented research papers; they notably only provide sparse evaluations and modeling details are often absent. This makes comparisons challenging, even more so as many of the details end-up being leaked or known through back-channels--a format which lends itself poorly to citation and attribution. For \citet{palm-2}, parameter count and pretraining length were reported by CNBC\footnote{\url{https://www.cnbc.com/2023/05/16/googles-palm-2-uses-nearly-five-times-more-text-data-than-predecessor.html}} for the largest of the three models. For \citet{gpt-4}, a number of leaks have occurred, most of which have been summarized in a piece by SemiAnalysis\footnote{\url{https://www.semianalysis.com/p/gpt-4-architecture-infrastructure}}.  

\section{Pseudocode samples}

\subsection{Measurement plan to measure all to all bandwidths/latencies efficiently}
\label{all_to_all_code}
\begin{python}
def get_all_comps(n:int):
    # n: power of two
    def op(l, d=4, r=1):
        l = l.reshape(-1, d)
        l[1::2] = np.roll(l[1::2], r, axis=1)
        return l.T.reshape(-1)

    x = np.array(list(range(n)))

    comps = []
    d = 1
    while d < n:
        for r in range(d):
            comps.append(op(x, d=d, r=r).copy())
        d *= 2
    ret = np.stack(comps)
    return ret.reshape(ret.shape[0], -1, 2)
\end{python}

\subsection{Converting tree token depth into an attention mask:}
\label{sec:attn_mask_code}
\begin{python}
    def attn_mask_pos_(x, attention_mask):
    for i in range(len(x)):
        attn = False
        for j in range(0, len(x)):
            attn |= (i + 1) == j
            attn &= x[i] < x[j]
            attention_mask[j, i] = ~attn & (i != j)
\end{python}

\subsection{\texttt{Zero-1} pseudo-code}
\label{zero_pseudocode}
\begin{python}
def reduce_scatter_grads(buffer):
    chunk_size = buffer.model_grads.numel() // dp_world_size
    t = empty_tensor(chunk_size, dtype=bfloat16)
    reduce_scatter_tensor(t, buffer.model_grad, group=data_parallel_group)
    buffer.optimizer_grads.copy(t)

def all_gather_opt_params_to_model_params(buffer):
    bfloat_sharded_params = buffer.optimizer_params.to(bfloat16)
    all_gather_into_tensor(buffer.model_params, bfloat_sharded_params, group=data_parallel_group)

...
reduce_scatter_grads(buffer)
optimizer.step()
all_gather_opt_params_to_model_params(buffer)
\end{python}

\section{Prompts}
\label{sec:ohmyprompt}

For \cref{sec:ablations} and \cref{sec:eai}, we use the default prompts of the Eleuther AI Harness \citep{eval-harness}--these prompts aim to reproduce the setup of the GPT-3 evaluations \citep{brown2020gpt3}. For results in \cref{sec:palm}, \cref{sec:gpt}, \cref{sec:common_sense} we use the prompts outlined thereafter; if no prompt is specified, then the default one from the Eleuther AI Harness is used. Specifically for \cref{sec:palm} we never outline candidates in multiple-choice questions (e.g., ARC, RACE, etc.) to allow for fair comparisons with PaLM-2: this left-out content is colored in \textcolor{gray}{gray} in the following prompts. 

The \textbf{Context} is appended once at the beginning of the prompt; \textbf{Sample(s)} are repeated for each shot, with the answer provided for all but the last sample; \textbf{Candidates} have their logprobability evaluated given the previous context and sample(s). Content in \emph{italics} is fetched from the task data (from the train set for illustrative purpose thereafter, but the test or validation sets are used whenever available for actual evaluation). Weay write down the candidates in a comma-separated list {[}candidate1, candidate2, ...{]}. Candidates are always preceded by a space for tokenization. 

\begin{table}[h]
\caption{\textbf{ANLI.}}
\begin{tabular}{lp{11.5cm}}
\toprule
\textbf{Context}   & Determine whether the statement made about the extract is true, false, or unsure.                                                                                                            \\
\textbf{Sample(s)} & Extract: \emph{Ernest Jones is a British jeweller and watchmaker. Established in 1949, its first store was opened in Oxford Street, London. Ernest Jones specialises in diamonds and watches, stocking brands such as Gucci and Emporio Armani. Ernest Jones is part of the Signet Jewelers group.} \\
& Statement: \emph{The first Ernest Jones store was opened on the continent of Europe.} \\
& Question: true, false, or unsure? \\
& Answer:\\
\textbf{Candidates}   & {[}true, false, unsure{]}       \\                                                 \bottomrule                                                                                                         
\end{tabular}
\end{table}

\begin{table}[h]
\caption{\textbf{ARC.} When options are not provided in the prompt, the candidates are directly the possible answers (e.g., dry palms, wet palms, etc.) instead of the letter keys.}
\begin{tabular}{lp{11.5cm}}
\toprule
\textbf{Context}   & Answer the following multiple-choice questions.\\
\textbf{Sample(s)} & Question: \emph{George wants to warm his hands quickly by rubbing them. Which skin surface will produce the most heat?} \\
& \textcolor{gray}{A. \emph{dry palms}} \\
& \textcolor{gray}{B. \emph{wet palms}} \\
& \textcolor{gray}{C. \emph{palms covered with oil}} \\
& \textcolor{gray}{D. \emph{palms covered with lotion}} \\
& Answer:\\
\textbf{Candidates}   & {[}A., B., C., D.{]}         \\                                                 \bottomrule                                                                                                         
\end{tabular}
\end{table}

\begin{table}[h]
\caption{\textbf{RTE.}}
\begin{tabular}{lp{11.5cm}}
\toprule
\textbf{Context}   & Determine whether the statement made about an extract is True, or False.\\
\textbf{Sample(s)} & Extract: \emph{Herceptin was already approved to treat the sickest breast cancer patients, and the company said, Monday, it will discuss with federal regulators the possibility of prescribing the drug for more breast cancer patients.} \\
& Statement: \emph{Herceptin can be used to treat breast cancer.} True or False? \\
& Answer:\\
\textbf{Candidates}   & {[}True, False{]}         \\ \bottomrule                                                                                            
\end{tabular}
\end{table}

\begin{table}[h]
\caption{\textbf{LAMBADA.} We evaluate whether the correct answer is the most likely one according to the model overall, without constraint to a set of predetermined candidates.}
\begin{tabular}{lp{11.5cm}}
\toprule
\textbf{Context}   & Complete the \_\_\_\_ in the following extracts.\\
\textbf{Sample(s)} & Extract: \emph{My wife refused to allow me to come to Hong Kong when the plague was at its height and –" "Your wife, Johanne? You are married at last ?" Johanne grinned. "Well, when a man gets to my age, he starts to need a few home comforts. After my dear mother passed away ten years ago now, I became \_\_\_\_} \\
& Completion:\\         \bottomrule                                                                        
\end{tabular}
\end{table}

\begin{table}[h]
\caption{\textbf{OpenBookQA.} When options are not provided in the prompt, the candidates are directly the possible answers (e.g., puppies learning new tricks, etc.) instead of the letter keys.}
\begin{tabular}{lp{11.5cm}}
\toprule
\textbf{Context}   & Answer the following multiple-choice questions.\\
\textbf{Sample(s)} & Question: \emph{The sun is responsible for} \\
& \textcolor{gray}{A. \emph{puppies learning new tricks}} \\
& \textcolor{gray}{B. \emph{children growing up and getting old}} \\
& \textcolor{gray}{C. \emph{flowers wilting in a vase}} \\
& \textcolor{gray}{D. \emph{plants sprouting, blooming and wilting}} \\
& Answer:\\
\textbf{Candidates}   & {[}A., B., C., D.{]}         \\                                                 \bottomrule                                                                                                         
\end{tabular}
\end{table}

\begin{table}[h]
\caption{\textbf{RACE.} Note that the way \citet{brown2020gpt3} and \citet{eval-harness} implement RACE means that the number of shots is at the article level; 3 questions regarding each article are always provided, even in 0-shot.}
\begin{tabular}{lp{11.5cm}}
\toprule
\textbf{Sample(s)} & Article: \emph{Last week I talked with some of my students about what they wanted to do after they graduated, and what kind of job prospects they thought they had. Given that I teach students who are training to be doctors, I was surprised do find that most thought that they would not be able to get the jobs they wanted without "outside help". "What kind of help is that?" I asked, expecting them to tell me that they would need a or family friend to help them out. "Surgery ," one replied. I was pretty alarmed by that response. It seems that the graduates of today are increasingly willing to go under the knife to get ahead of others when it comes to getting a job . One girl told me that she was considering surgery to increase her height. "They break your legs, put in special extending screws, and slowly expand the gap between the two ends of the bone as it re-grows, you can get at least 5 cm taller!" At that point, I was shocked. I am short, I can't deny that, but I don't think I would put myself through months of agony just to be a few centimetres taller. I don't even bother to wear shoes with thick soles, as I'm not trying to hide the fact that I am just not tall! It seems to me that there is a trend towards wanting "perfection" , and that is an ideal that just does not exist in reality. No one is born perfect, yet magazines, TV shows and movies present images of thin, tall, beautiful people as being the norm. Advertisements for slimming aids, beauty treatments and cosmetic surgery clinics fill the pages of newspapers, further creating an idea that "perfection" is a requirement, and that it must be purchased, no matter what the cost. In my opinion, skills, rather than appearance, should determine how successful a person is in his/her chosen career} \\
& Question: \emph{We can know from the passage that the author works as a\_.} \\
& \textcolor{gray}{A. \emph{doctor}} \\
& \textcolor{gray}{B. \emph{model}} \\
& \textcolor{gray}{C. \emph{teacher}} \\
& \textcolor{gray}{D. \emph{reporter}} \\
& Answer: C. teacher\\
& Question: \emph{Many graduates today turn to cosmetic surgery to\_.} \\
& \textcolor{gray}{A. \emph{marry a better man/woman}} \\
& \textcolor{gray}{B. \emph{become a model}} \\
& \textcolor{gray}{C. \emph{get an advantage over others in job-hunting}} \\
& \textcolor{gray}{D. \emph{attract more admirers}} \\
& Answer: C. get an advantage over others in job-hunting\\
& Question: \emph{According to the passage, the author believes that\_.} \\
& \textcolor{gray}{A. \emph{everyone should purchase perfection, whatever the cost}} \\
& \textcolor{gray}{B. \emph{it's right for graduates to ask for others to help them out in hunting for jobs}} \\
& \textcolor{gray}{C. \emph{it is one's appearance instead of skills that really matters in one's career}} \\
& \textcolor{gray}{D. \emph{media are to blame for misleading young people in their seeking for surgery}} \\
& Answer: D. media are to blame for misleading young people in their seeking for surgery\\
& Question: \emph{Which' s the best title for the passage?} \\
& \textcolor{gray}{A. \emph{Young Graduates Have Higher Expectations}} \\
& \textcolor{gray}{B. \emph{Young Graduates Look to Surgery for Better Jobs}} \\
& \textcolor{gray}{C. \emph{Young Graduates' Opinion About Cosmetic Surgery}} \\
& \textcolor{gray}{D. \emph{Young Graduates Face a Different Situation in Job-hunting}} \\
& Answer:\\
\textbf{Candidates}   & {[}A., B., C., D.{]}         \\                                                 \bottomrule                                                                                                         
\end{tabular}
\end{table}

\begin{table}[h]
\caption{\textbf{BoolQ.}}
\begin{tabular}{lp{11.5cm}}
\toprule
\textbf{Context}   & Answer the following questions about an extract by yes, or no.\\
\textbf{Sample(s)} & Extract: \emph{Powdered sugar, also called confectioners' sugar, icing sugar, and icing cake, is a finely ground sugar produced by milling granulated sugar into a powdered state. It usually contains a small amount of anti-caking agent to prevent clumping and improve flow. Although most often produced in a factory, powdered sugar can also be made by processing ordinary granulated sugar in a coffee grinder, or by crushing it by hand in a mortar and pestle.} \\
& Question: \emph{is confectionary sugar the same as powdered suga}, yes or no? \\
& Answer:\\
\textbf{Candidates}   & {[}yes, no{]}         \\ \bottomrule                                                 
\end{tabular}
\end{table}

\begin{table}[h]
\caption{\textbf{CB.}}
\begin{tabular}{lp{11.5cm}}
\toprule
\textbf{Context}   & Determine whether the statement made about by the extract is True, False, or Unsure.\\
\textbf{Sample(s)} & Extract: \emph{It was a complex language. Not written down but handed down. One might say it was peeled down.} \\
& Statement: \emph{the language was peeled down}. \\
& Question: True, False, or Unsure? \\
& Answer:\\
\textbf{Candidates}   & {[}True, False, Unsure{]}         \\ \bottomrule                                                 
\end{tabular}
\end{table}

\begin{table}[h]
\caption{\textbf{COPA.} We keep the formatting of \citet{eval-harness} unchanged, but add an instruction.}
\begin{tabular}{lp{11.5cm}}
\toprule
\textbf{Context}   & Complete the following sentences.\\ \bottomrule                                                 
\end{tabular}
\end{table}

\begin{table}[h]
\caption{\textbf{WiC.}}
\begin{tabular}{lp{11.5cm}}
\toprule
\textbf{Sample(s)} & Extract: \emph{place} has a similar meaning in the following two sentences. Yes or no? \\
& Sentence 1: \emph{Do you want to come over to my place later?} \\
& Sentence 2: \emph{A political system with no place for the less prominent groups.} \\
& Answer:\\
\textbf{Candidates}   & {[}yes, no{]}         \\ \bottomrule                                                    
\end{tabular}
\end{table}

\begin{table}[h]
\caption{\textbf{Winograd.} We keep the formatting of \citet{eval-harness} unchanged, but switch yes/no in the candidates for true/false.}
\begin{tabular}{lp{11.5cm}}
\toprule
\textbf{Candidates}   & {[}true, false{]}         \\ \bottomrule                                                    
\end{tabular}
\end{table}

\end{document}